\documentclass[preprint,12pt,authoryear]{elsarticle}

\usepackage{amssymb}

\usepackage{graphicx}
\usepackage{subcaption}
\usepackage{xcolor}
\usepackage{listings}
\usepackage{caption}
\usepackage{algorithm}
\usepackage{algpseudocode}
\usepackage{hyperref}

\journal{Artificial Intelligence Journal}

\begin{document}

\begin{frontmatter}



\title{Learning neuro-symbolic convergent term rewriting systems}



\author[inst1]{Flavio Petruzzellis}
\author[inst1,inst2]{Alberto Testolin}
\author[inst1,inst3]{Alessandro Sperduti}

\affiliation[inst1]{organization={Department of Mathematics, University of Padova},
            addressline={Via Trieste 63}, 
            city={Padova},
            postcode={35121}, 
            country={Italy}}
            
\affiliation[inst2]{organization={Department of General Psychology, University of Padova},
            addressline={Via Venezia 12}, 
            city={Padova},
            postcode={35131}, 
            country={Italy}}

\affiliation[inst3]{organization={Augmented Intelligence Center, Fondazione Bruno Kessler},
            addressline={Via Sommarive},
            city={Povo},
            postcode={38123},
            country={Italy}}

\begin{abstract}
Building neural systems that can learn to execute symbolic algorithms is a challenging open problem in artificial intelligence, especially when aiming for strong generalization and out-of-distribution performance. In this work, we introduce a general framework for learning convergent term rewriting systems using a neuro-symbolic architecture inspired by the rewriting algorithm itself. We present two modular implementations of such architecture: the Neural Rewriting System (NRS) and the Fast Neural Rewriting System (FastNRS). As a result of algorithmic-inspired design and key architectural elements, both models can generalize to out-of-distribution instances, with FastNRS offering significant improvements in terms of memory efficiency, training speed, and inference time. We evaluate both architectures on four tasks involving the simplification of mathematical formulas and further demonstrate their versatility in a multi-domain learning scenario, where a single model is trained to solve multiple types of problems simultaneously. The proposed system significantly outperforms two strong neural baselines: the Neural Data Router, a recent transformer variant specifically designed to solve algorithmic problems, and GPT-4o, one of the most powerful general-purpose large-language models. Moreover, our system matches or outperforms the latest o1-preview model from OpenAI that excels in reasoning benchmarks.
\end{abstract}


\begin{highlights}
\item A neuro-symbolic system inspired by rewriting algorithms can learn convergent term rewriting systems from training data.
\item The system exhibits systematic generalization beyond the training distribution, outperforming relevant neural baselines.
\item The system can learn several convergent term rewriting systems at the same time when trained in a multi-domain setting.
\end{highlights}


\end{frontmatter}




\section{Introduction}
\label{sec:introduction}
Deep learning systems have spread dramatically in the last decade thanks to their ability to automatize tasks typically carried out by humans, from basic pattern recognition \citep{7780459}, to natural language processing \citep{DBLP:conf/nips/BrownMRSKDNSSAA20} and the generation of synthetic but stunningly realistic media content \citep{chang2023musetexttoimagegenerationmasked}.
However, despite these advances, neural network models struggle with tasks that require iterative and reflective reasoning, which in humans require conscious deliberation and understanding beyond pattern recognition
\citep{Kahneman2011-jy}. As a result, deep learning often fails in problems where the ability to reason systematically through compositional concepts is essential, such as learning to simulate symbolic algorithms or solve advanced mathematical formulas \citep{testolin2023can,davis2024mathematics}.

On the other hand, computer programs stemming from classical artificial intelligence techniques have traditionally been very successful in the domain of mathematics and formal reasoning \citep{newell1956logic}. In such a framework, computer scientists model real-world phenomena using two equally important and complementary formalization tools: algorithms and data structures. The latter can be viewed as a formal description of the entities involved in the modeled phenomena, while the former are descriptions of the processes that can bring the modeled entities to a desired state. Under the assumption that input data are stored in specific data structures that the algorithm was designed to process, the result of the algorithm's execution will be predictable for the end-user. The power of algorithms can therefore be reduced to the use of \textit{abstractions} that formally describe real-world entities and processes \citep{Cormen2022-px}.

Classical algorithms in computer science -- such as sorting or graph traversal algorithms -- can map inputs to outputs independently of the data distribution from which the input was drawn, with well-studied time and space scaling laws. On the other hand, from a statistical perspective, it is almost always assumed that the inputs to deep learning models belong to the same data distribution of training samples, making it challenging to design learning architectures that can handle out-of-distribution (OOD) test data  \citep{DBLP:conf/iccv/HendrycksBMKWDD21,DBLP:conf/nips/YeXCLLW21}.
At the same time, however, the design and implementation of classical algorithms requires a significant amount of human labor: the programmer needs to formalize each problem class using specific data structures and engineer explicit algorithms that can solve the problem at hand. 
Given the immense wealth of digital data that is now available to organizations and the value they can bring if processed by computer programs, new lines of research propose to reduce the need for humans in the automation loop exploiting machine learning. In this context, information is encoded using distributed representations and is processed by manipulating numerical vectors rather than with classical algorithms and data structures \citep{Rumelhart1986-uj}. For example, graph neural networks have recently been used to learn graph algorithms, combining the strong real-world data handling capabilities of deep learning with the theoretical guarantees of classical algorithms \citep{DBLP:conf/iclr/VelickovicYPHB20,DBLP:journals/patterns/VelickovicB21}.

In this work, we consider a class of problems from the tradition of artificial intelligence and computer science that can be formalized as convergent term rewriting systems \citep{DBLP:books/daglib/0092409}. Generally speaking, rewriting systems are composed of a set of elements and a set of rules that describe how to transform those elements. Elements can be several different mathematical objects, including strings, graphs, or terms in a formula. When combined with an appropriate algorithm, rewriting systems become programs that can execute the transformation of a sequence of objects into another one by the subsequent application of the given rules.
We consider here term rewriting systems, in which the elements are mathematical expressions represented as sequences of symbols, and the rules define their semantically equivalent forms.
Specifically, in a \textit{convergent} term rewriting system, rewrite rules applied sequentially always transform the input into the same final output, independently of the order of application, and sequences of rewrite rules never form loops.

We present two related neuro-symbolic architectures designed to learn convergent term rewriting systems: the Neural Rewriting System (NRS) and the Fast Neural Rewriting System (FastNRS). Both models are built upon a shared architectural blueprint inspired by rewriting algorithms. They exhibit strong generalization capabilities, approaching those of traditional term rewriting systems, but their processing dynamics emerges through learning from data rather than manual design.
Both models can be trained on a limited subset of formulas and effectively generalize to more complex ones, eliminating the need for exhaustive training on all possible formulas. Such capability for out-of-distribution generalization is enabled by a modular approach informed by the rewrite mechanism and by architectural modifications to the transformer block.

We show that these models can function in a multi-domain scenario, a setting where a single model is trained on multiple datasets simultaneously without task-specific architectural adjustments. This results in a ``multi-potent'' system capable of solving a variety of problem instances within the considered class. Unlike the more conventional multi-task setting in machine learning \citep{caruana1997multitask}, where a shared backbone architecture is typically combined with task-specific outputs, our architecture’s algorithmic-inspired design allows the same components to be effectively applied to learn multiple term rewriting systems, enabling robust generalization without task-specific modifications.

While the NRS has been described in recent work \citep{nrs}, the FastNRS is presented here for the first time, together with analysis of the performance of both systems in a multi-domain scenario. The FastNRS system improves efficiency in terms of memory usage, training, and inference time, without significantly sacrificing performance. In this article, we detail the architectural elements of both the NRS and the FastNRS and provide a comparative analysis of their performance and efficiency on the four different domains of logic, lists, integer arithmetic, and simple algebra, describing the benefits and trade-offs associated with each system.

As additional baselines that could exhibit a systematic behavior similar to the execution of convergent term rewriting systems, we consider three architectures from two independent but related streams of research: the Neural Data Router \citep{DBLP:conf/iclr/CsordasIS22}, a recently proposed variant of transformer designed to achieve strong systematic generalization capabilities, which can be considered as a representative of small-scale neural architectures specialized on a single domain; OpenAI's GPT-4o \citep{openai2023gpt4}, one of the best performing general-purpose LLMs currently available, with strong reasoning capabilities that can further improve through prompting methods like Chain-of-Thought \citep{DBLP:conf/nips/Wei0SBIXCLZ22}; and OpenAI's o1-preview \citep{o1report}, a recently proposed LLM based on GPT-4 that has been optimized to excel in systematic reasoning tasks thanks to the production of long reasoning chains.

The remainder of this paper is organized as follows. Section \ref{sec:background} provides background on systematic generalization and algorithmic learning with neural networks, establishing the theoretical foundation for our work. In Section \ref{sec:problem}, we define the specific class of problems addressed in this study — formula simplification problems. In Section \ref{sec:architecture} we present the architectural blueprint of our models designed to solve these problems, and we describe each model in more detail. Section \ref{sec:experiments} details the experimental setup, including datasets composition and baselines, and Section \ref{sec:results} contains a presentation and analysis of the experimental results. Finally, Section \ref{sec:conclusions} concludes the paper. Additional methodological details and supplementary results are provided in~\ref{app:train-details} and~\ref{sec:appendix-solver-conf}.

\section{Background and related works}
\label{sec:background}
Connectionist systems designed to process symbolic data have been proposed since the 1990s \citep{DBLP:journals/ai/Hinton90}.
In the deep learning era, this line of research has seen a significant increase in interest and development, thanks to the introduction of novel neural architectures that could effectively process symbolic sequences, such as those based on external memory like the Neural Turing Machine \citep{DBLP:journals/corr/GravesWD14} and its successor, the Differentiable Neural Computer \citep{DBLP:journals/nature/GravesWRHDGCGRA16}.
These models are designed to learn algorithmic tasks by leveraging mechanisms that mimic classical computational processes, with memory-based architectures incorporating external memory to handle complex data structures, and attention-based models focusing on selectively attending to parts of the input sequence to perform tasks like sorting or routing \citep{DBLP:conf/nips/VinyalsFJ15}. Similar goals motivated parallel research efforts on the possibility of simulating the execution of classical algorithms in computer science with Graph Neural Networks \citep{DBLP:conf/iclr/VelickovicYPHB20}.
These initial contributions laid the groundwork for the framework of Neural Algorithmic Reasoning \citep{DBLP:journals/patterns/VelickovicB21}, which focuses on training neural networks to perform classical algorithms on graph-based problems bridging the gap between deep learning and traditional algorithmic theory.

At the same time, inspired by the rich debate in cognitive science and linguistics about the role of rules in language acquisition \citep{Pinker1988-vw}, other researchers started to investigate the capability of popular sequence-to-sequence architectures to extrapolate simple compositional rules from the training distribution and apply them to out-of-distribution test samples \citep{DBLP:conf/icml/LakeB18}. 
Among these approaches, one model introduced an \textit{ad hoc} trainable component named the copy-decoder, specifically designed to assist in learning to copy parts of the input to the output \citep{DBLP:journals/corr/abs-2110-04169}; another one proposed the Neural Data Router (NDR), a variant of transformer encoder tailored to compositional generalization problems with sequential solution procedures, which we consider as a baseline in our work \citep{DBLP:conf/iclr/CsordasIS22}. Other recent work explored the effectiveness of several architectural elements on the compositional generalization capabilities of transformers, such as recursive decoding \citep{DBLP:journals/corr/abs-2201-11766}, positional encodings and early stopping \citep{DBLP:conf/emnlp/CsordasIS21}.
Both initial contributions and subsequent research efforts demonstrated that recurrent and transformer-based models can achieve varying degrees of success, yet they still struggle to learn the underlying rules systematically and reliably from training data \citep{DBLP:journals/jair/HupkesDMB20,DBLP:conf/emnlp/CsordasIS21}.

A research problem closely related to the principles of productivity, systematicity, and substitutivity as described in \citet{DBLP:journals/jair/HupkesDMB20} is solving mathematical problems with neural networks, where learning to apply these principles is crucial to achieve true compositionality.
In this area, significant progress has been made using transformers to tackle a range of mathematical tasks, including arithmetic \citep{cognolato2022transformers}, derivation and integration \citep{Lample2019DeepLF} and polynomial simplification \citep{DBLP:journals/corr/abs-2104-14095}. However, research has shown that while transformers can learn to solve generic mathematical problems, their ability to generalize and apply learned rules systematically remains limited \citep{DBLP:conf/iclr/SaxtonGHK19,testolin2023can,davis2024mathematics}.

As an alternative approach more aligned with the principles of compositionality identified by \citet{DBLP:journals/jair/HupkesDMB20}, some researchers have explored the implementation of neural architectures inspired by symbolic rewriting systems. In early work, network weights were used to represent tokens to be rewritten in unsupervised learning settings \citep{icnc09}. Other approaches included using custom feature engineering and feed-forward networks for algebraic problems \citep{Cai2018-xx}. Yet other developments have introduced reinforcement learning-based systems that learn a general rewriting mechanism, selecting regions of a formula to simplify and applying appropriate rewriting rules, thereby aiming to achieve a more systematic and compositional handling of symbolic expressions \citep{DBLP:conf/nips/ChenT19}.

Given the recent prominence of large language models (LLMs), there is also a growing interest in understanding the symbolic reasoning abilities of foundation models trained on huge corpora of text and/or code \citep{chen2021evaluating,petruzzellisAssessingICANN24,lrec-coling24}.
These models could be considered the pinnacle of scientific and engineering advancements in neural technology, and their proficiency in language manipulation makes them particularly relevant for investigating systematic generalization from a linguistic and cognitive science perspective. Reasoning is one of the key abilities that is believed to emerge in very large models \citep{DBLP:conf/nips/Wei0SBIXCLZ22}, yet it remains an area of active research, with ongoing efforts aimed at achieving further improvements.
Symbolic reasoning tasks, which require the model to follow structured logical rules to arrive at a conclusion, can serve as a testbed for these abilities.
These symbolic reasoning tasks are similar to the problems addressed in this work, as they involve synthetically generated instances that can be solved by applying simple, algorithmic rules.
Examples of such tasks include coin flip prediction, last letter concatenation \citep{DBLP:conf/nips/Wei0SBIXCLZ22}, and boolean variable assignment \citep{DBLP:conf/nips/AnilWALMRSGDN22}, all of which require a form of systematic rule application similar to compositional generalization benchmarks. Interestingly, research on prompt engineering techniques has shown that appropriate prompting can enhance the reasoning capabilities of LLMs also on symbolic reasoning tasks \citep{DBLP:conf/nips/Wei0SBIXCLZ22,DBLP:conf/iclr/0002WSLCNCZ23}. For instance, Chain-of-Thought prompting leverages the auto-regressive nature of these models to break down complex problems into multi-step reasoning chains, enabling more effective processing of contextual information.
Leveraging reasoning steps in context has also been adopted as the core strategy to improve reasoning capabilities in the new OpenAI o1 family of language models specialized for complex reasoning tasks \citep{o1report}, an exemplar of which we consider in this work.

\section{Formula simplification problems}
\label{sec:problem}
Convergent term rewriting systems are typically used to simplify mathematical formulas. Here, rules describe how expressions involving two or more operands can be rewritten into atomic elements that are semantically equivalent and represent their values. For example, in arithmetic formulas, the expression $(15+5)$ can be transformed into the equivalent atomic element $20$.
We call these problems ``formula simplification problems.''
We will now formally characterize these problems and then use this formalization to describe how convergent term rewriting systems for this class of problems can be formed, i.e., how rewriting rules can be written and what algorithm is implemented by such a rewriting system.

In formula simplification problems, we can see formulas $f \in F$ as entities that are composed of two semantically distinct elements: operators $o \in O$ and arguments $a \in A$. 
Arguments can be either atomic elements $e \in E$, such as integers, which are also the final values of any formula, or other formulas.
Furthermore, since these problems can be solved by convergent term rewriting systems and thus always have exactly one final value, the following equality holds for any formula: $f = o(a_1,...,a_n) = e$, where \mbox{$o\in O$, $e \in E$} and \mbox{$a_j \in F \cup E, ~\forall j \in [1, n]$.}
Finally, we can define leaf formulas \mbox{$F^L \subset F$} as the subset of formulas whose arguments are all atomic elements: \mbox{$f^L=o(a_1,...,a_n) \;\ \mathrm{s.t.}\;\  a_k \in E, ~\forall k \in [1,n]$}.

For any problem, we can define a set of rewriting rules $r \in R$ which map leaf formulas to their values: $r: F^L \rightarrow E, ~\forall r \in R$.
This set defines a convergent term rewriting system for any formula simplification problem.
Indeed, iteratively applying the rewriting rules described above in any order, the initial formula can be simplified to an atomic final element $e$.
The algorithm implemented by the rewriting system thus consists of the execution of four steps, which are repeated until the formula becomes an atomic value: \textbf{1.} pick any valid rewriting rule from the set;\footnote{In step \textbf{1.}, a rewriting rule that can be applied to the current formula should be chosen, i.e., one whose left-hand side appears in the formula.} \textbf{2.} find in the input formula all the elements that match the left-hand side of the rewriting rule; \textbf{3.} apply the rewriting rule to the elements found and compute the substitution; \textbf{4.} replace the elements with the computed values.
This algorithm serves as a blueprint for the design of our architectures, providing strong guarantees on the reliability of the models, regardless of how they are implemented.

\begin{figure}[t]
    \centering
    \begin{subfigure}[b]{0.47\textwidth}
        \centering
        \includegraphics[width=\textwidth]{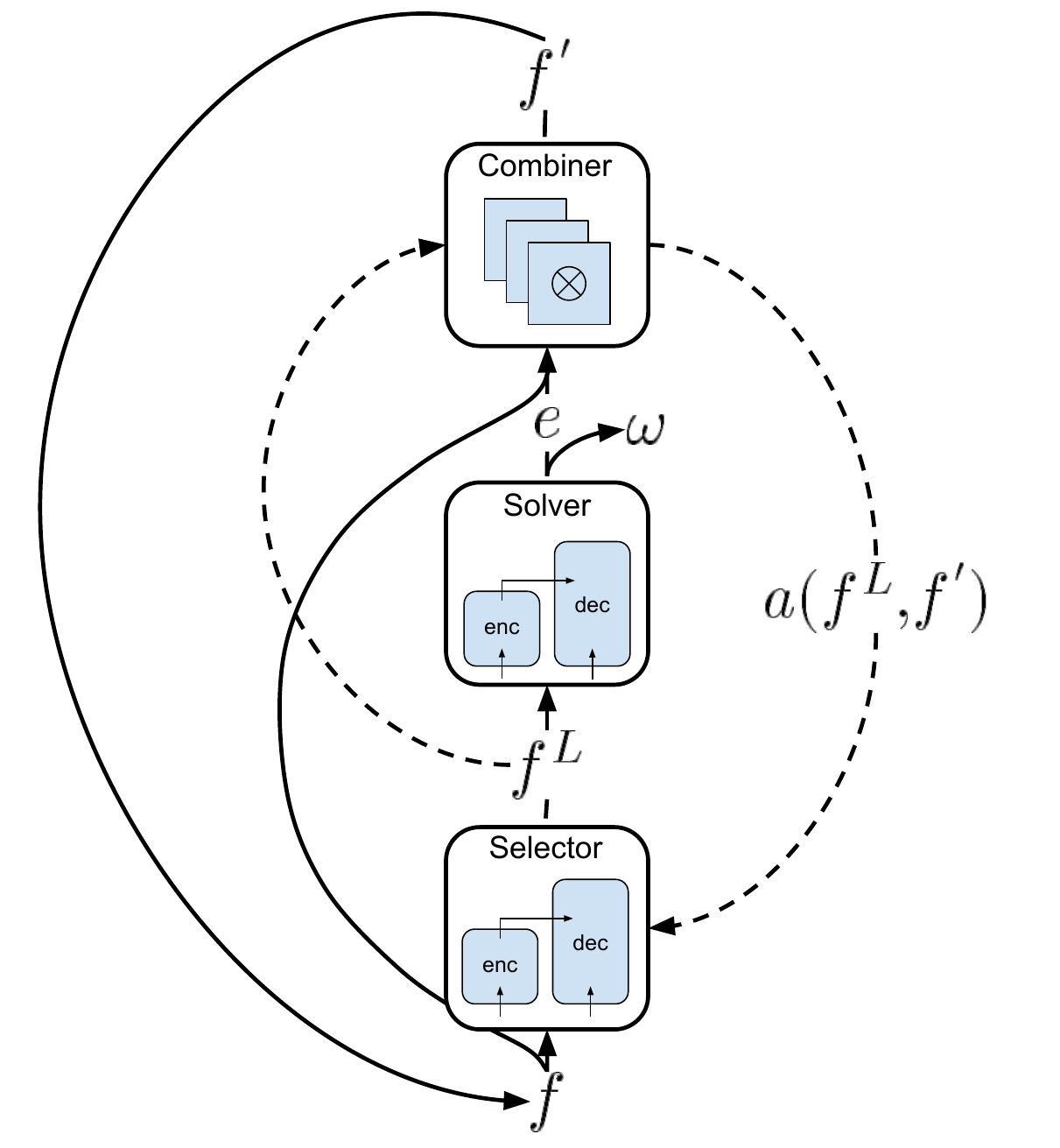}
        \caption{NRS architecture. The three modules in the NRS operate sequentially, but the Selector and the Combiner also interact via the agreement score $a(f^L, f')$ to produce the Selector output, as described in Paragraph \ref{subpar:nrs-sel}.}
        \label{fig:nrs}
    \end{subfigure}%
    \hfill
    \begin{subfigure}[b]{0.47\textwidth}
        \centering
        \includegraphics[width=\textwidth]{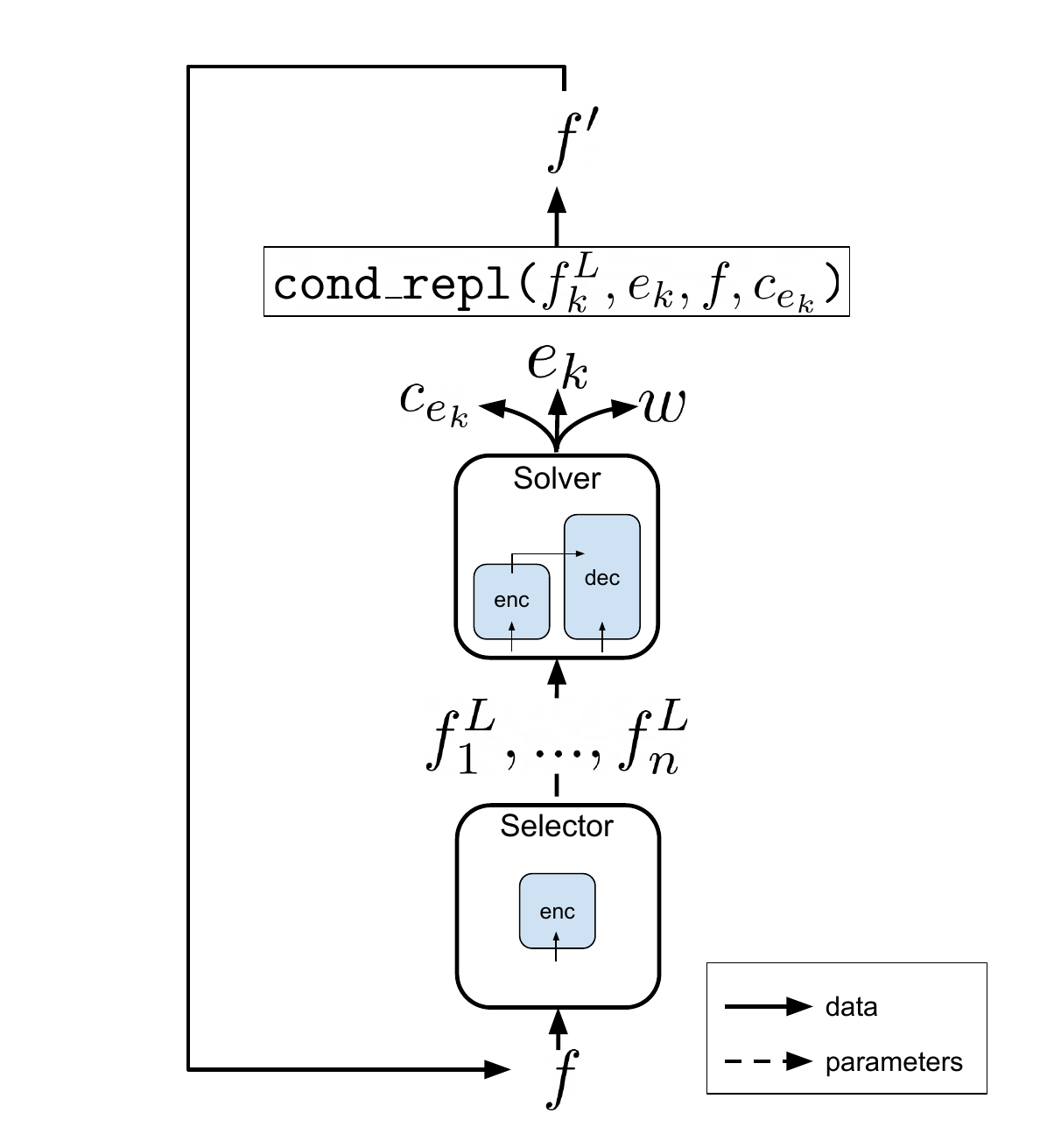}
        \caption{FastNRS architecture. The Selector module in the FastNRS operates by selecting $n$ leaf formulas in parallel with a text-segmentation mechanism, as described in Paragraph \ref{sec:selector}. Each leaf formula is then processed independently by the Solver.}
        \label{fig:fast-nrs}
    \end{subfigure}
    \caption{Schematic representations of the NRS and FastNRS architectures. Both architectures implement the three modules of the algorithmic blueprint described in Section \ref{sec:architecture}. The models process input formulas $f$, selecting one or more leaf formulas $f^L$. Solver modules simplify leaf formulas to atomic values $e$ and produce a special end-of-computation token $w$. Combiner modules produce simplified formulas $f'$.}
\end{figure}

\section{Neuro-Symbolic architectures to learn rewriting systems}
\label{sec:architecture}

Traditional term rewriting systems rely heavily on manually crafted rules and algorithms tailored to specific problems, which require significant human expertise and effort. In contrast, our approach aims to \textit{learn} both the rewriting rules and the contexts in which these rules should be applied. By mirroring the algorithm's steps in the architecture design itself, we adopt a structured approach similar to that of classical algorithms. At the same time, combining this with the flexibility of learning-based methods, we obtain a versatile system that can be applied to different problems within the class we consider. 
This adaptability is particularly advantageous in multi-domain scenarios, where the same architecture can be effectively employed for diverse problems, as we demonstrate with experimental results.

Compared to other neural systems, such as general-purpose Large Language Models (LLMs) and specialized neural architectures for compositional reasoning, our system balances generality and reliability. Indeed, while LLMs can tackle a wide range of tasks, they lack the specialized focus needed to provide strong guarantees in solving structured problems like those that can be addressed by term rewriting. On the other hand, small-scale transformer-based architectures designed for compositional reasoning tend to provide stronger generalization guarantees, but the spectrum of tasks they can handle is limited, if defined at all \citep{DBLP:conf/iclr/ZhouBLRSSBN24}. Our neuro-symbolic system is more specialized than LLMs but more versatile than narrowly specialized architectures, offering a structured and more reliable solution for solving formula simplification problems across different domains.

Differently from other neuro-symbolic architectures proposed in the literature, our system does not include any symbolic AI component (e.g., a symbolic solver).
We describe our system as neuro-symbolic since it is composed of neural modules whose interaction schema is informed by the rewriting algorithm: In the taxonomy of neuro-symbolic AI proposed by \citet{Kautz2022-kx}, our system could be grouped in the class of neural architectures whose structure is obtained using symbolic rules as a template, called Neuro\_\{Symbolic\} systems by the author.

The algorithm for resolving mathematical formulas we described in Section \ref{sec:problem} can be implemented in a neuro-symbolic architecture in various ways, depending on design choices and priorities. 
In this work, we present two neuro-symbolic implementations of this algorithmic blueprint: the Neural Rewriting System (NRS) and the Fast Neural Rewriting System (FastNRS).
Both these architectures are composed of three modules, the Selector, the Solver and the Combiner, which handle steps \textbf{1.} and \textbf{2.}, \textbf{3.} and \textbf{4.} of the algorithm, respectively.
Note that in a neural implementation, picking a valid rule can coincide with identifying a part of the input that can be simplified, as will be clear from our implementation of the Solver. 
These designs, represented schematically in Figures \ref{fig:nrs} and \ref{fig:fast-nrs}, differ mainly in how the Selector module is implemented, corresponding to steps \textbf{1.} and \textbf{2.} of the algorithm—the identification of matching elements for rule application. 
In the Neural Rewriting System, only one element matching the left-hand side of a rewriting rule is selected at a time. 
On the other hand, the Fast Neural Rewriting System selects and replaces multiple elements in parallel by framing the problem as a text segmentation task, allowing for a more efficient implementation at the expense of some accuracy.

\begin{algorithm}[ht]
\begin{algorithmic}[1]
\Function{NRS}{$f$}
\While{True}
    \State $f^L, c(f^L), a(f^L, f) \gets \Call{Selector}f$
    \If{$a(f^L, f) \neq 1$}
        \State \Return $\varnothing$
    \EndIf
    \State $e \gets \Call{Solver}{f^L}$
    \If{$e = \omega$}
        \State \Return $f$
    \EndIf
    \State $f \gets \Call{Combiner}{f, f^L, e}$
\EndWhile
\EndFunction
\end{algorithmic}
\captionof{algorithm}{Pseudo-code of the NRS  execution. The model executes the Selector, Solver and Combiner in a pipeline. Algorithm \ref{algo:nrs-sel} describes the execution of the Selector.}\label{algo:nrs}
\end{algorithm}

\subsection{The Neural Rewriting System (NRS)}
\label{sec:nrs}
The general functioning of the architecture can be described as the iterative execution of the Selector, Solver and Combiner modules in a pipeline.
A formal description in pseudo-code of the NRS execution is given in Algorithms \ref{algo:nrs} and \ref{algo:nrs-sel}.

\begin{algorithm}[ht]
\begin{algorithmic}[1]
\Require parameters $M, T$
\Function{Selector}{$f$}
\State $L \gets [\;]$
\For{$i = 1 \to M$}
    \If{$|f| < T$}
        \State $\hat{f}^{L}, c(\hat{f}^{L}) \gets \Call{Trfm}{f}$ 
        \State $a(\hat{f}^L, f) \gets \frac{\max ~ CNN_{\hat{f}^L}{(f)}}{|\hat{f}^L|}$
        \State Append $\langle\hat{f}^{L}, c(\hat{f}^{L}), a(\hat{f}^L, f)\rangle$ to $L$
    \Else
        \State $k \gets \mathrm{floor}(|f| \cdot \frac{i \bmod 20}{20})$ 
        \State $f_w \gets w({f, k})$ 
        \State $\hat{f}^{L}, c(\hat{f}^{L}) \gets \Call{Trfm}{f_w}$ 
        \State $a(\hat{f}^L, f_w) \gets \frac{\max ~ {CNN_{\hat{f}^L}}{(f_w})}{|\hat{f}^L|}$ 
        \State Append $\langle\hat{f}^{L}, c(\hat{f}^{L}), a(\hat{f}^L, f_w)\rangle $ to $L$
    \EndIf
\EndFor
\State Sort $L$ by $a(\hat{f}^L, f)$ and $c(\hat{f}^{L})$ 
\State \Return $L[0]$
\EndFunction
\end{algorithmic}
\captionof{algorithm}{Pseudo-code of the NRS Selector execution. The $\textsc{Trfm}$ function represents the transformer. The model conditionally applies Dynamic Windowing (lines 9-13) depending on the input length.}\label{algo:nrs-sel}
\end{algorithm}

\subsubsection{The Selector module}
\label{subpar:nrs-sel}
The Selector module in the Neural Rewriting System is responsible for identifying an element in the input formula that matches the left-hand side of a rewriting rule.
As mentioned earlier, it is designed to solve a sequence-to-sequence task.
Formally, it implements the $sel: F \rightarrow F^L$ function, i.e. it is trained to map a formula, which we assume to always be syntactically correct, to a leaf formula appearing therein. In analogy to what happens in humans when they deploy object-based attention to locate algebraic sub-expressions that can be simplified \citep{marghetis2016mastering}, the Selector is trained to identify the last leaf formula occurring in the input formula on which a rewriting rule can be applied.
For example, given the arithmetic expression \mbox{\tt (12+(3-(4+5)))} the Selector's task is to correctly identify the solvable leaf formula \mbox{\tt (4+5)}.

We use a variant of the transformer encoder-decoder \citep{DBLP:conf/nips/VaswaniSPUJGKP17} to implement the core of the NRS Selector. In order to achieve strong length generalization capabilities, we make two modifications to the vanilla transformer.
First, we follow recent evidence showing that length generalization in transformers can be influenced by the choice of positional encodings, especially when, at test time, these fall out of the range observed during training \citep{DBLP:conf/emnlp/CsordasIS21,DBLP:conf/acl/RuossDGGCBLV23,DBLP:conf/nips/KazemnejadPRDR23}. 
We thus use Label-based Positional Encodings \citep{DBLP:journals/corr/abs-2210-00400} to enable the Selector to identify leaf formulas in very long sequences.
The positional information of an input sequence of $L$ tokens is thus encoded in the following way: given a sequence of $N$ sinusoidal positional encodings, where $N$ is a large number that represents the maximum expected length of an input, $L$ integers are sampled in the interval $[0, N-1]$ and then sorted.
The encodings found in the positions corresponding to the sampled integers are then summed to the embeddings of the tokens in the input sequence before the forward pass.
Notice that the sampling and sorting mechanisms are applied internally in the transformer, similar to the pooling operations in convolutional networks. Furthermore, this type of positional encoding operates as a sort of data augmentation mechanism and thus is not learned and is not involved in the backpropagation of the errors.

As a second modification, we constrain the receptive field of the self-attention layer of the encoder.
This choice was motivated by both intuition and extensive experimentation.
Indeed, the Selector more likely learns a function with good length generalization properties in a smaller search space that contains functions with minimal dependencies on parts of the sequence that do not correspond to leaf formulas.
Furthermore, identifying leaf formulas is a local problem in any part of the input sequence, i.e., it can be solved without integrating information carried by tokens located in distant parts of the input sequence.
Therefore, we mask all entries in the self-attention matrix of the encoder but the ones around the main diagonal (i.e., we make them $-\mathrm{inf}$).
The active values in the self-attention matrix are thus located in a diagonal window that is $2k+1$ tokens wide, where $k$ is a hyperparameter. 
Preliminary experiments showed that models with vanilla self-attention achieved worse out-of-distribution generalization, and hyperparameter selection demonstrated that the strongest generalization can be achieved with a narrow diagonal window.

Other than these architectural modifications, the NRS Selector includes two specialized mechanisms -- the multi-output generation and the dynamic windowing -- that enhance accuracy and reliability, ensuring higher resilience to noise and errors in the neural network's outputs. 
The multi-output generation mechanism was introduced after we observed experimentally that it can be useful to repeat the auto-regressive generation of transformer outputs.
After sampling several output sequences from the probability distribution derived from the decoder's outputs, we choose the best one considering both a measure of confidence of the Selector and a measure of input-output agreement computed by the Combiner.

Given the specialized purpose of the Selector, we can see each output of the module as a candidate leaf formula $\hat f^L$.
We generate any token $\hat f_i^L$ in an output sequence $\hat f^L$ by sampling from the probability distribution obtained applying the $\mathrm{softmax}$ function to the logits produced by the final fully-connected layer of the decoder.
We do not use any temperature parameter when sampling the output tokens.
For any input formula $f$, we repeat the stochastic generation process $M$ times, thus generating a sequence of candidate leaf formulas \mbox{$\hat F^L = \langle \hat{f}^{L,1},...,\hat{f}^{L,M}\rangle$}.s
We define the confidence of the Selector on any $\hat f^{L,j}, 1 \leq j \leq M$ as the joint probability of sampling its tokens: $c(\hat f^{L,j})=\prod_{i=1}^{N} p_i^j$, where $N$ is the number of tokens in $\hat f^{L,j}$, and $p_i^j$ is the probability to sample token $\hat f_i^L$ in $\hat f^{L,j}$.
We also define an agreement score $a(\hat f^{L,j},f) \in [0,1]$ which gives information on the fraction of $\hat f^{L,j}$ that is exactly present in the input formula $f$.
This measure is computed by the Combiner and thus it is formally defined in Section \ref{subsec:com}.
We then select the final output $f^L$ of the Selector as the one with the highest confidence which has an agreement score equal to $1$ --- that is, it matches the input sequence exactly.
More formally, \mbox{$f^L = \hat f^{L,j} \in \hat F^L \;$} $\mathrm{s.t.}\;\ $ \mbox{$c(\hat f^{L,j}) \geq c(\hat f^{L,k}) ~\forall j,k \in [1, M] \land a(\hat f^{L,j}, f) = 1$.}

We also implement a dynamic windowing mechanism on longer input sequences that allows us to increase the model's generalization capacity on complex problem instances.
The core idea behind this mechanism is to repeat the process of selecting a leaf formula several times, changing each time the \textit{window} of the input formula that the Selector observes, and then relying on the confidence $c(\hat f^{L})$ to pick the best output.
We apply this mechanism on top of multi-output generation by modifying its behavior for sequences longer than a given threshold $T$.
Given an input formula $f$, if $|f|<T$ the computation is executed as described before.
Otherwise, we generate $M$ copies of the input $\langle f^{(1)},...,f^{(M)}\rangle$, whose lengths
will be reduced by applying a window function $w$.
Considering any input $f$ as a sequence $f_1,...,f_N$ of $N$ tokens, we define the window function $w(f,k)=f_{k+1},...,f_N$ which reduces the length of the input by giving as output its last $k$ tokens.
Since the Selector is trained to output the last leaf formula appearing in the input, the window function reduces the input length starting from the first tokens.
We divide the sequence of copies of the input $\langle f^{(1)},...,f^{(M)}\rangle$ into 20 groups $F^{(1)},...,F^{(20)}$ of equal size.
Intuitively, in each group the length of the input is reduced by a different percentage of tokens.
More formally, the window function will be parameterized by \mbox{$k=\mathrm{floor}(|f^{(i)}| \cdot \frac{j}{20}) ~\forall f^{(i)} \in F^{(j)}, ~\forall j \in [1, 20]$}.
We then pick the final leaf formula using the confidence and agreement scores, as described in the previous paragraph.
This ensures that the model can observe the whole input sequence and select a leaf expression in the part of the input where it can identify one with more confidence.

\subsubsection{The Solver module}
The Solver a central component in our system.
Indeed, as we described in Section \ref{sec:introduction}, classic rewriting systems are composed of a set of elements and a set of rules, which are then used in the algorithm to transform sequences.
In our neuro-symbolic architecture, both elements and rewriting rules are represented sub-symbolically in the Solver, which rewrites relevant parts of the input.

Given a leaf formula $f^L$ by the Selector, the Solver is trained to produce the equivalent reduction $e$ according to the corresponding rewriting rule.
Therefore, valid elements and rewriting rules are implicitly stored in the network weights through optimization. 
For example, given the leaf formula \mbox{\tt (4+5)}, the Selector produces its value, {\tt 9}.
The Solver also learns to recognize the termination state of the computation, signaling when such a state is reached. 
Given atomic elements representing the final value of a formula, such as the number {\tt 9} for an arithmetic formula, it is trained to map them to the special symbol $\omega$, indicating the end of computation.
During training, the Solver only observes well-formed leaf formulas and atomic values.

We frame the Solver task as a sequence-to-sequence problem. We implement it as a transformer encoder-decoder without any modification since it learns input-output mappings corresponding to the rules.

\subsubsection{The Combiner module}
\label{subsec:com}
The last module in the architecture is the Combiner, a neural implementation of the function $com: F \times F^L \times E \rightarrow F$. Its purpose is thus to produce a simplified version of the original formula, given the formula itself $f$, the leaf formula $f^L$ identified by the Selector, and its reduction $e$ computed by the Solver.

In order to carry out its task, the first operation that the Combiner must perform is finding the position in $f$ where the leaf formula $f^L$ appears. 
We notice that the convolution is a suitable operation to detect which portion of an input sequence has the highest match with another sequence used as a filter, so we implement this operation using a 2D Convolutional Neural Network (CNN) whose filters are set dynamically at execution time using the output of the Selector, rather than being learned with backpropagation.
For example, if we have the arithmetic expression \mbox{\tt (12+(3-(4+5)))} as input, using the leaf formula \mbox{\tt (4+5)} as the filter of a 2D CNN we can obtain a signal of correspondence between the leaf formula and the input expression, and therefore identify if the leaf formula is present in the input and where it is located.

More precisely, we represent both the input sequence $f$ and the leaf formula $f^L$ as sequences of one-hot vectors over the same vocabulary.
Since the leaf formulas found for different sequences in a batch can have different lengths, we pad each one with zeros to prevent the padding to match in the input.
Then, we set the filter of the 2D CNN to the 1-hot representation of $f^L$.
We refer to the CNN parameterized in this way as $\mathrm{CNN}_{f^L}$.
Doing so allows us to obtain from the output of the convolution both information on the location of the best match of $f^L$ in $f$ and on the number of tokens in $f^L$ that match $f$ exactly in some point.
Indeed, we can compute the location of the best match as $\mathrm{pos}(f^L, f)=\mathrm{argmax} ~\mathrm{CNN}_{f^L}(f)$.
Furthermore, we can calculate the agreement score $a(f^L,f)=\frac{\mathrm{max} ~\mathrm{CNN}_{f^L}(f)}{|f^L|}$, where $|f^L|$ is the number of tokens in $f^L$.
Dividing by $|f^L|$ makes the score normalized, which allows us to compare the agreement scores of leaf formulas with different lengths.
Indeed, as described in Section \ref{subpar:nrs-sel}, the Selector uses this score for multi-output generation to discard the outputs that do not have an exact match in the input formula.
Notice that in this case the CNN is parameterized using candidate leaf formulas $\hat f^L_j$ whose accuracy scores with $f$ are compared.
If there is no Selector output such that $a(\hat f^L_j, f)=1$, the computation on the input sequence $f$ is stopped, and this is considered a failure of the model.

After finding the position of the leaf expression in $f$, the Combiner replaces $f^L$ with $e$ in $f$, to compute the simplified version of the formula $f'$.
We implement this operation as a deterministic operator with input $f$, $f^L$, $e$, and $\mathrm{pos}(f^L, f)$.

\begin{algorithm}[ht]
\begin{algorithmic}[1]
  \Function{FastNRS}{$f$}
    \While{$\mathrm{True}$}
        \State $\mathrm{mask} \gets \Call{Selector}{f}$
        \State $\langle f_1^L,...,f_n^L \rangle \gets \Call{Extract}{\mathrm{mask}, f}$
        \State $\mathrm{replaced} \gets \mathrm{False}$
        \For{$f^L_k \enspace \mathrm{in} \enspace \langle f_1^L,...,f_n^L \rangle$}
            \State $e_k, c_{e_k} \gets \Call{Solver}{f_k^L}$
            \If{$e_k = \omega$}
                \State \Return $f$
            \EndIf
            \State $f, \mathrm{replaced} \gets \texttt{cond\_repl(}f_k^L, e_k, f, c_{e_k}\texttt{)}$
        \EndFor
        \If{$\mathrm{replaced} = \mathrm{False}$}
            \State $\textbf{return} \enspace \varnothing$
        \EndIf
    \EndWhile
  \EndFunction
\end{algorithmic}
\captionof{algorithm}{Pseudo-code of the FastNRS execution. The model iterates through the leaf formulas extracted by the Selector and conditionally replaces them in the main formula.}\label{algo:fast-nrs}
\end{algorithm}

\subsection{The Fast Neural Rewriting System (FastNRS)}
\label{sec:fast-nrs}
The general functioning of the architecture can be described as the iterative execution of the Selector, the Solver and a deterministic \texttt{cond\_repl} function in a pipeline.
A pseudo-code description of the FastNRS execution is shown in Algorithm \ref{algo:fast-nrs}.

\subsubsection{The Selector module}
\label{sec:selector}
Unlike the NRS, the FastNRS Selector is implemented using only a transformer encoder.
This module shares the same core architecture as the transformer encoder used in the Neural Rewriting System. Specifically, we use Label-based Positional Encodings to enable the Selector to identify leaf formulas within very long sequences, and we constrain the receptive field of the self-attention layer to obtain localized attention on the close neighbors of each token. 
In FastNRS, the Selector is designed to solve a text-segmentation task. Formally, it implements the function $multisel: F \rightarrow {F^L}^n$ where ${F^L}^n$ represents the $n$-ary cartesian product of the set of leaf formulas $F^L$. The function maps a formula to one or more leaf formulas within it, corresponding to the left-hand sides of applicable rewrite rules. 

In this implementation, given a sequence of tokens, the transformer performs a binary classification task on each token independently. A positive label indicates that a token is part of a leaf formula and will be selected for rewriting, while a negative label marks tokens that will not be selected.
Given an input formula $f$, the Selector produces a mask over the input in which all parts of the formula that cannot be rewritten are masked.
Using these masks, leaf formulas are extracted from the input, and a sequence of strings is obtained.
Each string should correspond to the left-hand sides of some rewriting rule, and thus, it is given as input to the Solver, which computes the substitution according to the corresponding rule.

\subsubsection{The Solver module}
The Solver module in FastNRS shares the same architecture as the Solver module in the NRS and is designed to solve the same problem: applying the appropriate rewriting rule to compute the required substitutions. Also in this case, if the Solver outputs the $\omega$ symbol signaling the end of computation, the algorithm stops.
Additionally, in FastNRS, we measure the confidence of the Solver's outputs, which plays a critical role in guiding the execution of the FastNRS Combiner module. This confidence measure helps ensure that only high-confidence outputs are used in the subsequent steps, enhancing the reliability of the overall system.
We define the confidence of the Solver on any output $e$ as the joint log-probability of sampling the output tokens from the distribution obtained by applying the $\mathrm{softmax}$ function to the logits produced by the final fully-connected layer of the decoder. Formally, $c_e=\sum_{i=1}^{N} \mathrm{log}(p_i)$, where $N$ is the number of tokens in $e$, and $p_i$ is the probability of sampling token $e_i$ in $e$.

\subsubsection{The Combiner module}
In contrast to the Neural Rewriting System (NRS), the Combiner module in FastNRS is not implemented using a neural architecture. Specifically, we do not use a convolutional neural network (CNN) to extract the position signal of the leaf formula identified by the 	Selector. Thanks to the text-segmentation implementation of the 	Selector, we can directly trace back the position of the identified leaf formula(s) within the input.

As a result, the Combiner module is implemented as a deterministic function, \texttt{cond\_repl}, which takes the original formula $f$, the identified leaf formula $f^L$, its replacement $sol(f^L)$ computed by the 	Solver, and the measure of the Solver’s confidence $c_{e_k}$ as inputs. This confidence measure handles cases where the Solver output may contain errors.
Indeed, despite the strong length generalization properties guaranteed by the modifications made to the Selector, minor defects in the segmentation of out-of-distribution sequences can still occur.
Such defects could cause the extraction of corrupt parts of the input formula that do not correspond to the left-hand side of any rewriting rule, thus leading to meaningless substitutions computed by the Solver.
Any $f^L$ will thus be replaced with $sol(f^L)$ in $f$ only if the corresponding measure of confidence of the Solver output $c_{e_k}$ is sufficiently high.
Intuitively, the measure $c_{e_k}$ reflects the distance of the input from the training distribution of true left-hand sides of rewriting rules, and thus allows the identification of parts of the input that are not valid left-hand sides with a sufficient degree of accuracy.
In our experiments, we set the confidence score threshold depending on the distribution of this quantity on the training samples, as detailed in Section \ref{sec:models}.

If the input mask is drastically corrupted, and no $e_k$ has a sufficiently high confidence score $c_{e_k}$, the computation is interrupted and this is considered an error of the model.

\section{Experimental Setup}
\label{sec:experiments}

\subsection{Datasets}
\begin{figure}
    \centering
    \includegraphics[width=1\linewidth]{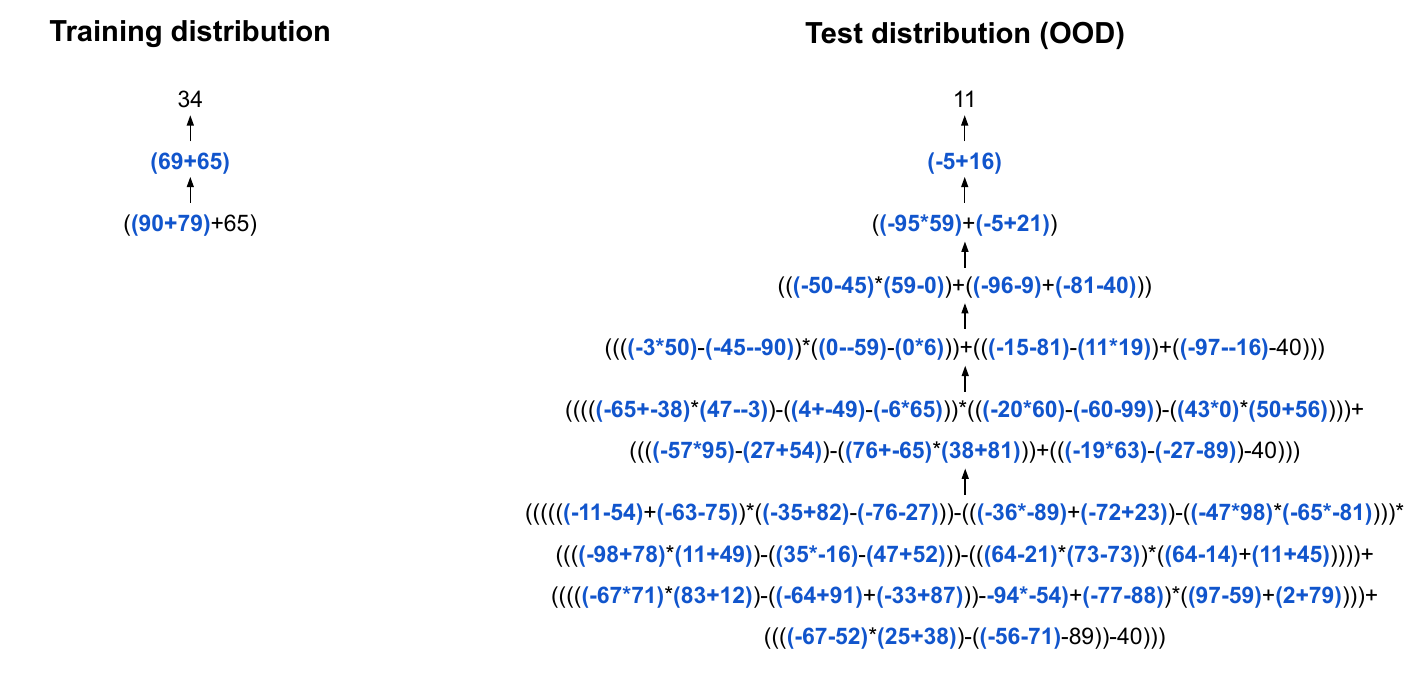}
    \caption{Visual representation of the simplification process of samples from the training set and the out-of-distribution (OOD) test set. The input parts that are simplified at each step are highlighted in blue.}
    \label{fig:dataset}
\end{figure}

We benchmark the proposed architecture on four formula simplification problems from different domains: logic, operations on lists, arithmetic, and algebra.
For all problems, formulas are generated automatically, and their difficulty is determined by specifying the desired nesting level of the formula.
Any formula is nested at each level in two points: exactly two arguments in the formulas on that level will be other formulas.
We now describe the formulas for each domain in more detail.

\subsubsection{Logic}
We build a dataset of nested logical formulas where the logical operators \texttt{AND}, \texttt{OR} and \texttt{NOT} are applied to non-grounded literal variables, represented by lowercase letters in $\{\texttt{a}, ..., \texttt{z}\}$ or grounded logical variables \texttt{True} and \texttt{False}.
Formulas are generated automatically specifying the desired number of nesting levels.
Unlike the other three datasets, logical formulas are nested up to 12 times, thus requiring more steps to be solved.
Each logical formula can be reduced either to a non-grounded literal variable or to a logical value between \texttt{True} and \texttt{False}.
For example, the logical formula \texttt{(((z OR (z OR (b AND False))) OR z) AND ((((j OR False) AND True) AND False) OR True))} is nested 5 times, contains the literal variables \texttt{b}, \texttt{j} and \texttt{z}, the logical values \texttt{True} and \texttt{False}, and evaluates to \texttt{z}.

\subsubsection{Listops}
The ListOps dataset \citep{DBLP:conf/naacl/NangiaB18} was designed to assess neural networks' ability to construct parse trees for nested formulas. Initially, the dataset featured formulas with operations on integer lists, such as minimum, maximum, median, and sum modulo $10$. We adapted the ListOps dataset to ensure that each nesting level had exactly two nesting points and allowed for specifying the number of arguments at any level. Focusing on the system's ability to generalize on deeply nested formulas rather than mastering specific operations, we limited the operations to minimum, maximum, and sum modulo $10$.

\subsubsection{Arithmetic}
We created formulas using sum, subtraction, and multiplication operations between two integers sampled from the interval $[-99,99]$.
Since this study does not explore the ability to generalize to numbers with more digits than those encountered during training, we applied modulo $100$ to the intermediate results obtained throughout the solution process.

\subsubsection{Algebra}
We focus on a subset of algebraic formulas that can always be deterministically simplified to a minimal form. These formulas consist of sums and subtractions between two monomials, with the final value always being a monomial. The numerical coefficients of the monomials were sampled from the interval $[-99,99]$, and each monomial could include up to four literal variables chosen from $\{a,b,x,y\}$. All monomials in a given formula shared the same literal variables. Similarly to the Arithmetic problem, all intermediate numerical values were computed modulo 100 when determining the formula's final value.

\subsection{Models}
\label{sec:models}

\subsubsection{Neural Rewriting System}
\label{sec:models-nrs}
We describe here how we built the training and validation sets for the Selector and Solver modules for both the NRS and the FastNRS. A visual representation of the solution process of samples from the training and test distributions is shown in Figure \ref{fig:dataset}.
Statistics on all development splits used to train the models are provided in~\ref{app:dataset-statistics}.
Further methodological details can be found in~\ref{sec:appendix-details-fastnrs}.

In the training set of the Selector module, we included formulas with nesting levels of 1, 2, and 3 for all problems, along with atomic elements representing the final value of the initial formula.
Simplifying formulas iteratively by reducing leaf formulas generates several intermediate simplifications of the original formula.
To demonstrate the full solution process to the Selector, we also included these intermediate formulas as steps in the training set.

We created a separate in-distribution validation set with samples mirroring the structural characteristics of those in the training set.
Unlike typical machine learning tasks, where models are tested on the same data distribution they were trained on, we aim for the Selector to demonstrate out-of-distribution (OOD) generalization abilities, identifying leaf formulas even in longer inputs than those encountered during training.
Therefore, we also developed a distinct OOD validation set featuring formulas with higher structural complexity, using this set for model selection.
For all problems, we included in this set samples with nesting levels of 4, 5, and 6.
To choose the most capable model throughout the iterative resolution process, we also added formulas representing examples of intermediate resolution steps.
To manage the structural complexity of the formulas, we balanced the OOD validation sets across the nesting levels of the leaf expressions.

The training and validation sets for the Solver contained two types of samples: leaf formulas, which were mapped to their equivalent atomic elements, and atomic elements, which were mapped to the end-of-computation special symbol $\omega$. To prevent bias toward solving leaf formulas, we generated training batches that included both types of samples with equal probability.

Both the NRS and the FastNRS behavior can be modulated by choosing the value of some hyperparameters at inference time.
The values of threshold \textit{T} that regulates the Dynamic Windowing mechanism in the NRS Selector was chosen by examining the average Selector confidence score for inputs of the same length.
We define these thresholds as 150 for ListOps and algebraic formulas, and 125 for arithmetic formulas, while we do not employ the mechanism on formulas in the Logic domain.
We provide a representation of the average confidence score values in~\ref{appendix-conf-scores}.
The values of the threshold on Solver confidence that regulates the \texttt{cond\_repl} function were chosen based on the distribution of these scores on training samples, as mentioned in Section \ref{sec:fast-nrs}.
The thresholds used were -6 for ListOps, -2 for Arithmetic, -3 for Algebra, and -0.005 for Logic.
The distributions of Solver confidence scores on training samples, are provided in~\ref{sec:appendix-solver-conf}.

\subsubsection{Neural Data Router}
\label{sec:exp-ndr}
The Neural Data Router (NDR) is a modified transformer encoder designed to tackle algorithmic problems with robust out-of-distribution generalization capabilities. Previously, this model has been tested on relatively simple algorithmic benchmarks, such as solving formulas in the original ListOps dataset and handling basic arithmetic formulas with single-digit integers, which closely resemble the problems we address. However, the key difference in our Arithmetic and Algebra problems is the increased complexity of the operands. Additionally, we employ significantly fewer and less complex samples during training, as the NDR was initially trained on arithmetic and ListOps formulas containing up to 5 nested operations.

We made a minor modification to the architecture to adapt the model to our specific problems. In the original work, the problems always resulted in a single-digit integer, which the model was trained to output as the first token in the sequence generated by the encoder. Since this is not generally applicable to our problems, we read the final answer from the first $k$ positions of the sequence produced by the encoder, where $k$ is the maximum length of a problem's targets.

We constructed all development sets for the NDR using the same top-level formulas included in the analogous sets for the Selector. Following the original experimental protocol, we ensured that the training set was balanced across nesting levels. Similar to the Selector module training, we created both in-distribution and out-of-distribution validation sets, using the latter to optimize hyperparameters through a Bayesian search using the Weights \& Biases platform in the same hyperparameters intervals described in the original work. The final hyperparameter values for each task are detailed in~\ref{sec:appendix-details-ndr}.

\subsubsection{OpenAI GPT-4}
In our experiments, we evaluate the performance of OpenAI's GPT-4 on the nested formulas in the four domains.
Using specialized prompts is currently considered the most effective method to improve the reasoning capabilities in large language models by researchers and practitioners.
Specifically, Chain-of-Thought (CoT) prompting has been found to enhance the performance of large language models on reasoning tasks by facilitating step-by-step solution procedures. We opted to prompt GPT-4 using a combination of self-consistency prompting \citep{DBLP:conf/iclr/0002WSLCNCZ23} and zero-shot Chain-of-Thought (CoT) \citep{DBLP:conf/nips/KojimaGRMI22}. Zero-shot CoT is a simpler alternative to traditional CoT prompting, achieving comparable performance on reasoning benchmarks without the need to engineer exemplars for few-shot reasoning. This is done by simply initiating the model's response with the sentence: ``Let's think step-by-step.'' Once the model generates a response, it is prompted again to produce a well-formatted output. Self-consistency prompting leverages the idea that reasoning problems can have multiple valid paths leading to the same conclusion. Thus, we generate 10 outputs for each input and select the most consistently produced one. This approach enhances confidence in the model's output and significantly improves accuracy, leading to a marked improvement in performance.

The zero-shot CoT prompt was designed by providing the model with a brief description of the problem and then asking it to solve it. For instance, the zero-shot CoT prompt for the ListOps input sample \texttt{[MIN[SM54][MIN39]]} is: ``\textit{\texttt{MIN}, \texttt{MAX}, and \texttt{SM} are operators on lists of single-digit integers, representing minimum, maximum, and sum modulo 10, respectively. Solve the following expression using these operators: \texttt{[MIN [SM 5 4] [MIN 3 9]]}.}'' Following this initial prompt, the model was subsequently prompted a second time to provide a well-formatted final answer.

\subsubsection{OpenAI o1-preview}
OpenAI has recently released a new family of highly capable models designed to excel in complex reasoning tasks. These models build on previous versions (like GPT-4) but focus on spending more time ``thinking'' before responding, making them particularly suitable for domains such as mathematics, coding, and logical reasoning.
The o1 series introduces several new features, including ``reasoning tokens'', which are assumed to represent the model's internal thought process. There are currently two versions available: o1-mini, which has been designed for efficiency, and o1-preview, which is larger, slower, and more expensive, but also more accurate. In this work, we therefore only focus on the latter, more powerful model.

\section{Results}
\label{sec:results}
In this section, we present the evaluation of the Neural Rewriting System and the Fast Neural Rewriting System, focusing on performance and efficiency, both in single-domain and multi-domain scenarios.
Since the final target of any formula corresponds to an atomic value, we measure the performance of the models on all tasks using Sequence Accuracy, i.e. the exact match between model output and target sequence.
The test sets on which all models are evaluated are composed of 100 formulas per nesting level. We observed non-significative variance across runs, which we therefore do not report.

\subsection{Learning domain-specific convergent term rewriting systems}
In this section, we evaluate the performance of both models across all four datasets — Logic, ListOps, Arithmetic, and Algebra — in a single-domain scenario. We compare the NRS and the FastNRS to the Neural Data Router (NDR), which constitutes a neural baseline that has also been separately trained on individual tasks.

\begin{figure}[t]
    \centering
    \begin{subfigure}[b]{0.355\textwidth}
     \centering
         \includegraphics[width=\textwidth]{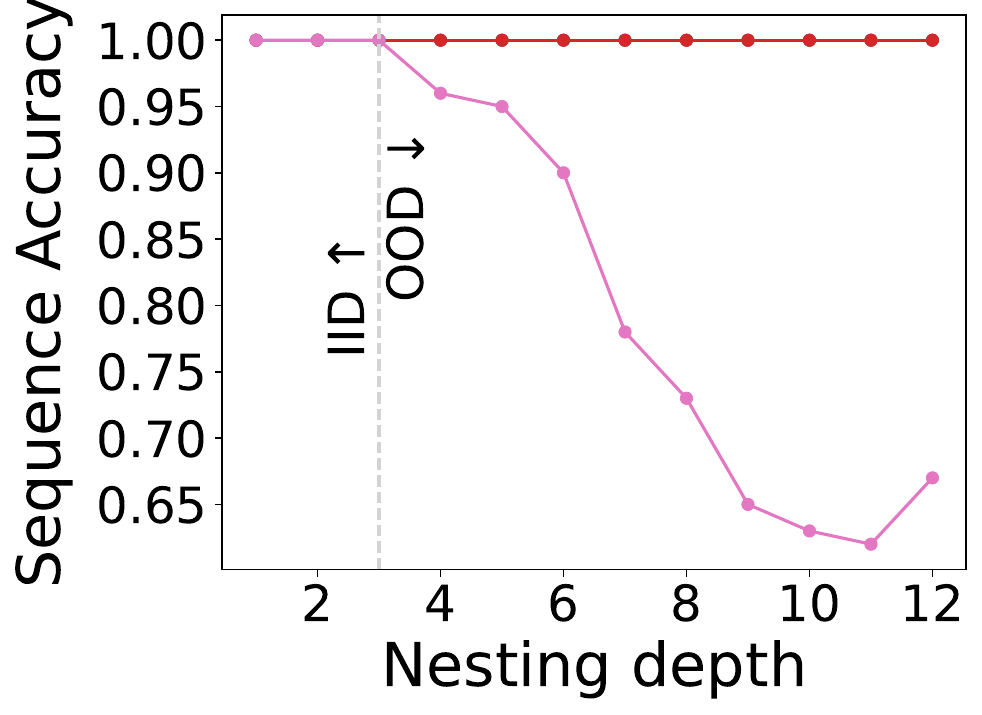}
         \caption{Logic}
         \label{fig:acc-table-logic}
    \end{subfigure}
    \begin{subfigure}[b]{0.315\textwidth}
     \centering
         \includegraphics[width=\textwidth]{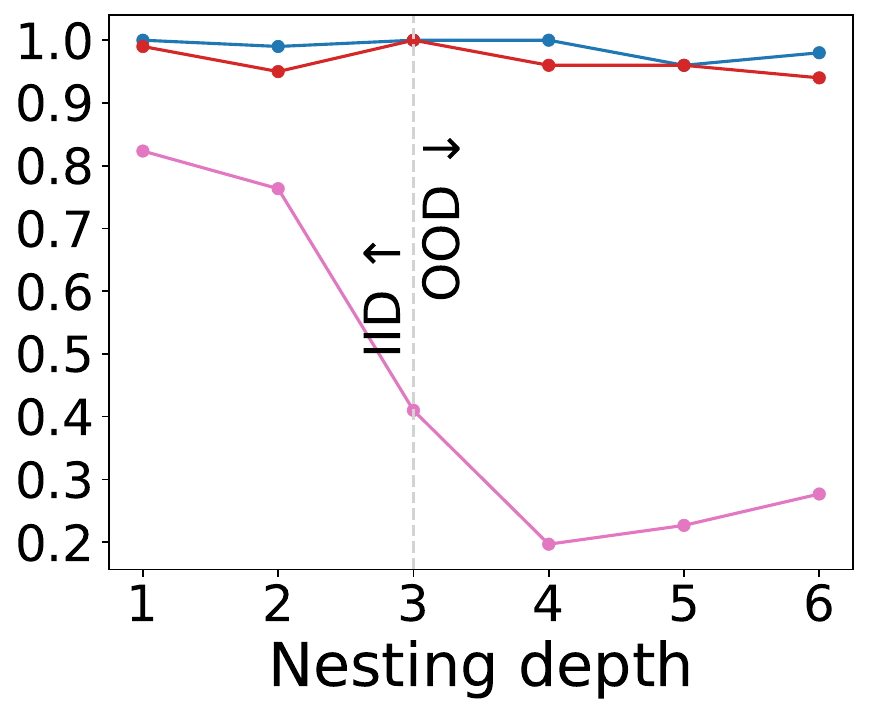}
         \caption{Listops}
         \label{fig:acc-table-listops}
    \end{subfigure}\\
    \begin{subfigure}[b]{0.355\textwidth}
     \centering
         \includegraphics[width=\textwidth]{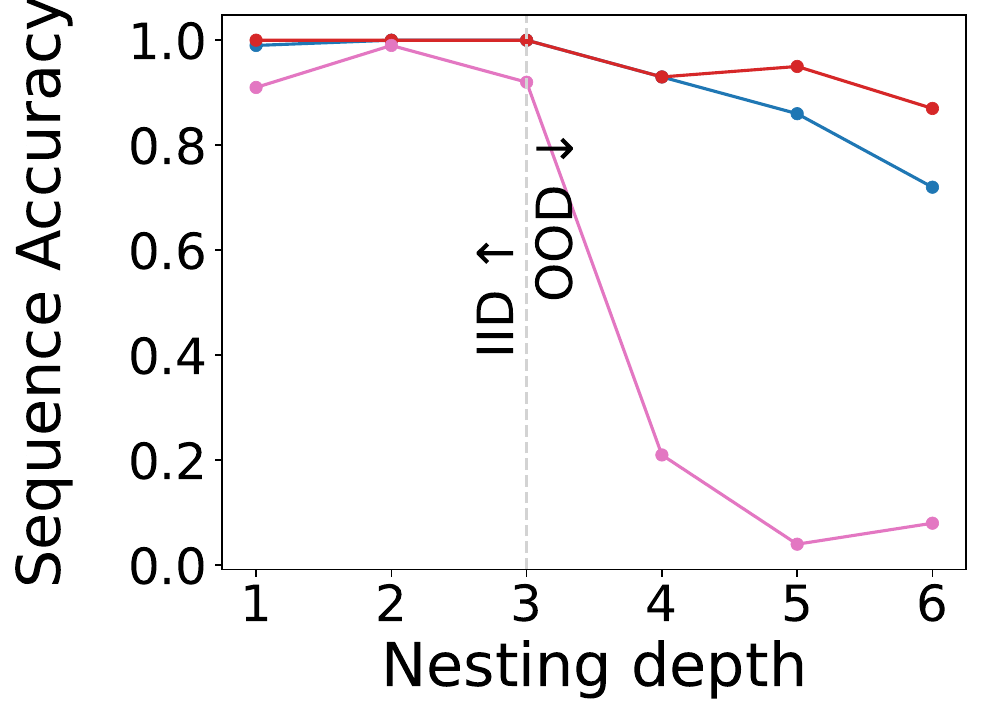}
         \caption{Arithmetic}
         \label{fig:acc-table-arithmetic}
    \end{subfigure}
    \begin{subfigure}[b]{0.315\textwidth}
     \centering
         \includegraphics[width=\textwidth]{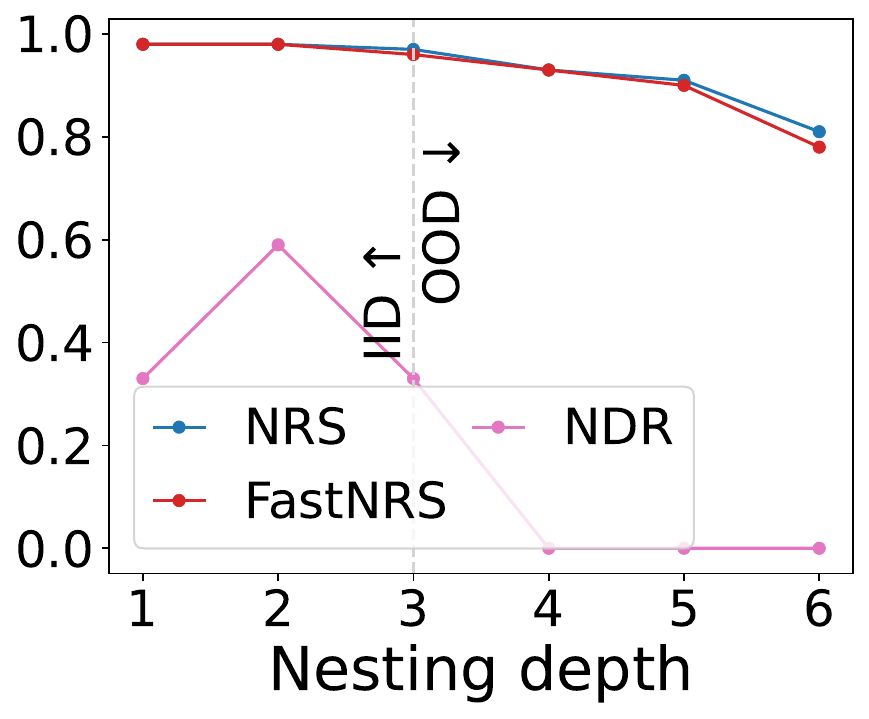}
         \caption{Algebra}
         \label{fig:acc-table-algebra}
    \end{subfigure}
    \caption{Performance of FastNRS, NRS, and NDR on each domain. Sequence accuracy is measured on data splits of 100 samples.}
    \label{fig:acc-tables}
\end{figure}

The performance of the models on all domains is represented in Figure 
\ref{fig:acc-tables}.
Across all datasets, the models show similar performance on in-distribution samples, with NDR generally performing the worst. 
However, on out-of-distribution samples, the baseline models exhibit a much sharper decline in performance compared to both the NRS and the FastNRS.

Interestingly, the FastNRS is more accurate than the NRS on deeply nested arithmetic formulas.
Therefore, in this case, the design choices in the FastNRS yield a significant improvement in terms of performance other than efficiency.
By examining the type of errors committed by both systems on arithmetic formulas in Section \ref{sec:err-analys}, we will clearly see how the superior performance of the FastNRS depends on the greater robustness in the identification of leaf formulas, guaranteed by the text segmentation-based implementation of the Selector module.

\subsection{Learning multi-domain convergent term rewriting systems}

\begin{figure}[t]
    \centering
    \begin{subfigure}[b]{0.355\textwidth}
     \centering
         \includegraphics[width=\textwidth]{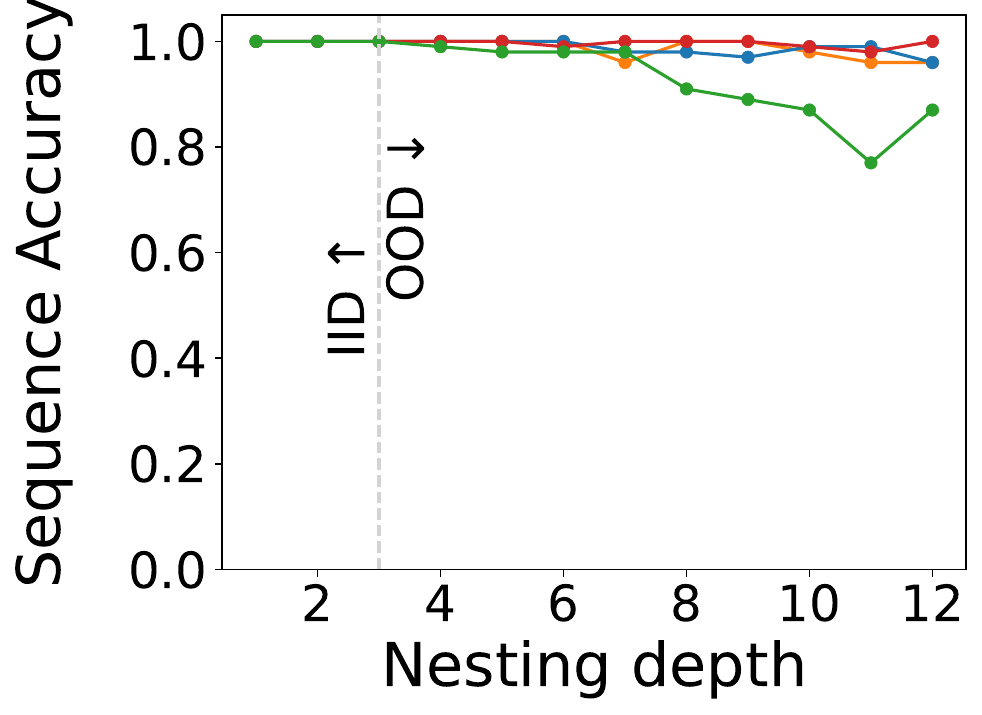}
         \caption{Logic}
         \label{fig:acc-table-logic-1}
    \end{subfigure}
    \begin{subfigure}[b]{0.315\textwidth}
     \centering
         \includegraphics[width=\textwidth]{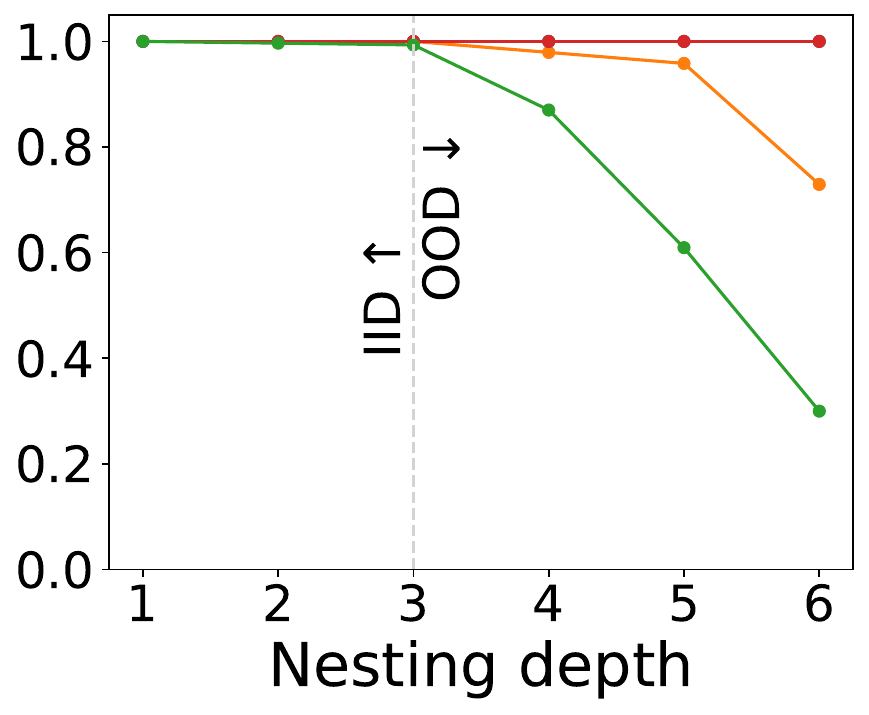}
         \caption{Listops}
         \label{fig:acc-table-listops-o1}
    \end{subfigure}\\
    \begin{subfigure}[b]{0.355\textwidth}
     \centering
         \includegraphics[width=\textwidth]{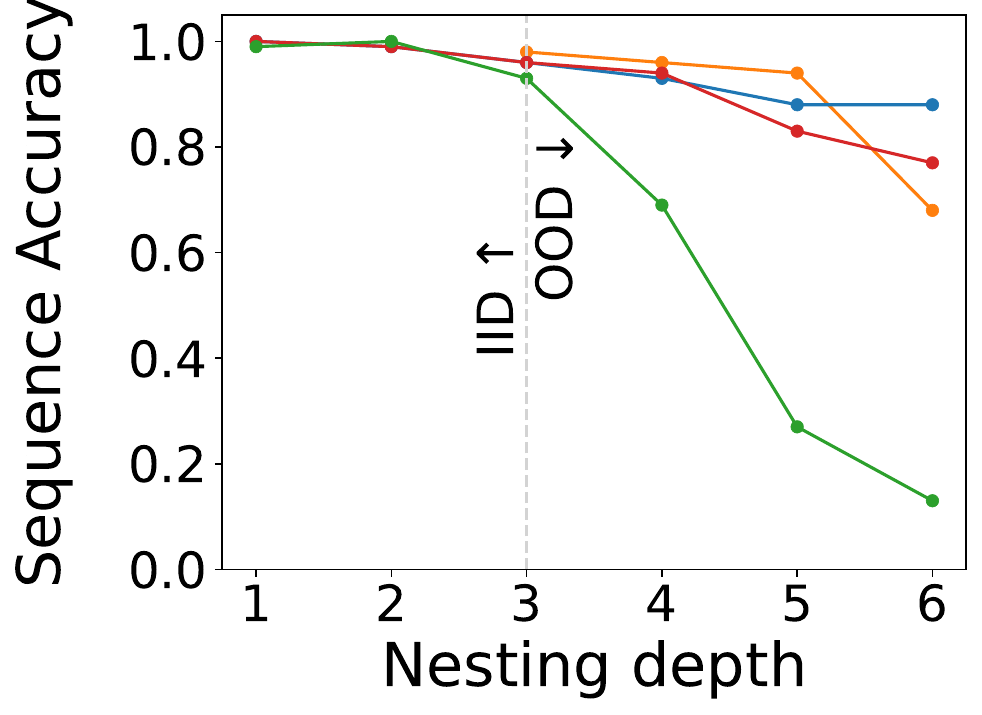}
         \caption{Arithmetic}
         \label{fig:acc-table-arithmetic-o1}
    \end{subfigure}
    \begin{subfigure}[b]{0.315\textwidth}
     \centering
         \includegraphics[width=\textwidth]{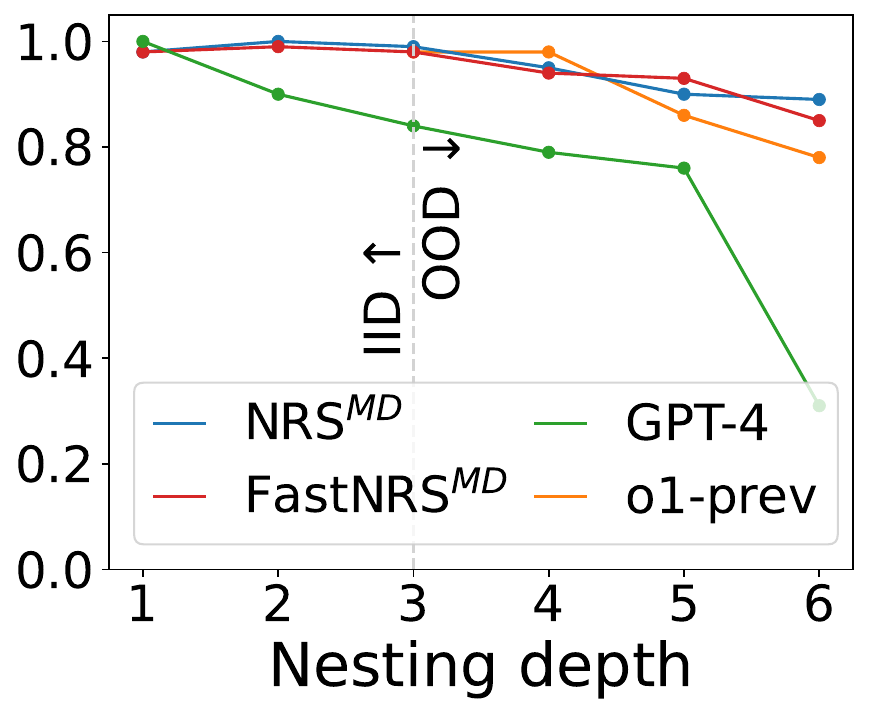}
         \caption{Algebra}
         \label{fig:acc-table-algebra-o1}
    \end{subfigure}
    \caption{Performance of multi-domain models: GPT-4, o1-preview, FastNRS and NRS on each domain. Sequence accuracy is measured on data splits of 100 samples.}
    \label{fig:acc-tables-o1}
\end{figure}

As detailed in Section 4, the architectures and execution dynamics of both the NRS and the FastNRS are specifically designed to support learning algorithms within the class of convergent term rewriting systems. This capability is primarily attributed to two key aspects: the algorithmic-inspired modular design of the architectures and the strong out-of-distribution generalization capability of the Selector module (see Section \ref{sec:res-selector}). As we mentioned, this generalization capability is particularly useful in a multi-domain scenario.

\begin{table}[ht]
    \centering
    \begin{tabular}{l|r|r}
     \textbf{Problem} & \textbf{\# Param.} & \textbf{Inf. time} \\
     \hline
     \hline
     Multi-domain & 18,651,574 & 51h 48m 7s\\
     \hline
     Logic & 3,047,365& 16m 57s\\
     ListOps & 3,842,209& 8h 24m 30s\\
     Arithmetic & 10,890,364& 7h 59m 30s\\
     Algebra & 9,904,856& 14h 16m 48s
    \end{tabular}
    \caption{Space and time efficiency statistics for the NRS.}
    \label{tab:nrs_table}
\end{table}

In this case, we choose to benchmark the models against OpenAI's GPT-4 and o1-preview, whose training regimen is multi-domain by definition. The o1-preview model was benchmarked only on out-of-distribution data splits, as simpler formulas can be considered trivial for this type of model. Performance metrics for the four models are illustrated in Figure \ref{fig:acc-tables-o1}.
GPT-4 is the weakest performer across the tasks, particularly struggling with the ListOps, Arithmetic, and Algebra benchmarks. While it still surpasses the previously proposed Neural Deductive Reasoner (NDR), it shows significant limitations when faced with deeply nested formulas with complex operands.
On the other hand, o1-preview demonstrates a substantial leap in performance over GPT-4. This improvement can be attributed to the more advanced CoT reasoning process implemented in the model, which enables it to process and simplify formulas more effectively, even within the complex tasks under consideration. These results could provide evidence of the importance of producing explicit reasoning steps for tasks that demand compositional reasoning.

In the multi-domain training scenario, both the Neural Rewriting Systems demonstrate their capacity to generalize to out-of-distribution samples.
Specifically, the NRS has a slightly better performance than the FastNRS, especially on out-of-distribution samples, demonstrating the effectiveness of the specialized architectural elements introduced for this purpose.
In the Logic and ListOps domains, the FastNRS maintains nearly the same accuracy as the NRS, with only a slight accuracy decrease of few percentage points in some cases. 
Notice that in the Logic domain, we evaluated the models on out-of-distribution test formulas with up to 12 nesting levels, where both models achieve consistently high accuracy.
In these domains, the models show superior or similar performance to o1-preview on both in-distribution and out-of-distribution data splits.
In the more complex Arithmetic and Algebra domains, there is a slightly larger drop in accuracy on certain out-of-distribution formulas.
On these two tasks, the o1-preview model shows superior performance on some of the out-of-distribution splits. 
However, the NRS still outperforms o1-preview on the most complex formulas, demonstrating its effectiveness in learning convergent term rewriting systems with significant generalization capabilities.

\begin{table}[ht]
    \centering
    \begin{tabular}{l|r|r}
         \textbf{Problem} & \textbf{\# Param.} & \textbf{Inf. time} \\
         \hline
         \hline
         Multi-domain & 15,061,616 & 3m 42s\\
         \hline
         Logic & 2,501,795& 38s\\
         ListOps & 4,095,633& 32s\\
         Arithmetic & 8,728,338& 50s\\
         Algebra & 8,752,920& 3m 02s
    \end{tabular}
    \caption{Space and time efficiency statistics for the FastNRS.}
    \label{tab:fastnrs_table}
\end{table}

\subsection{Analysis of efficiency}
The design of the FastNRS mainly leads to performance improvements with respect to the twin implementation of the framework we propose.
The number of parameters and inference time statistics for the NRS and the FastNRS in both training scenarios are reported in Table \ref{tab:nrs_table} and Table \ref{tab:fastnrs_table}\footnote{All runs were executed on a single NVIDIA A100 GPU. Statistics reported for the NRS refer to the simplest hyperparameters configuration that achieves the best performance in each test scenario.}. 
The FastNRS has, in most cases, fewer parameters than the NRS, and it achieves several orders of magnitude speed-up in inference time, both in the case of single- and multi-domain models (in the case of the multi-domain setting, cumulative inference time on all test samples in the four tasks is reported for both models).
This efficiency gain demonstrates the possibility of implementing an efficient learning mechanism for the solution process of symbolic formulas across different domains within a single, unified neural circuitry. 
The modifications introduced in the FastNRS thus not only improve computational efficiency but also preserve the model’s generalization capabilities, and, as we have seen, can occasionally improve performance.

For comparison, we report analogous statistics for the baseline models in \ref{app:baseline-stats}. The LLMs considered in this work have a much larger number of parameters compared to our neural architectures (several order of magnitudes, according to independent estimates), and their inference times are much higher compared to our efficient FastNRS model. The Neural Data Router, instead, has a slightly lower complexity both in terms of number of parameters and inference time.

\begin{figure}
    \centering
    \includegraphics[width=\linewidth]{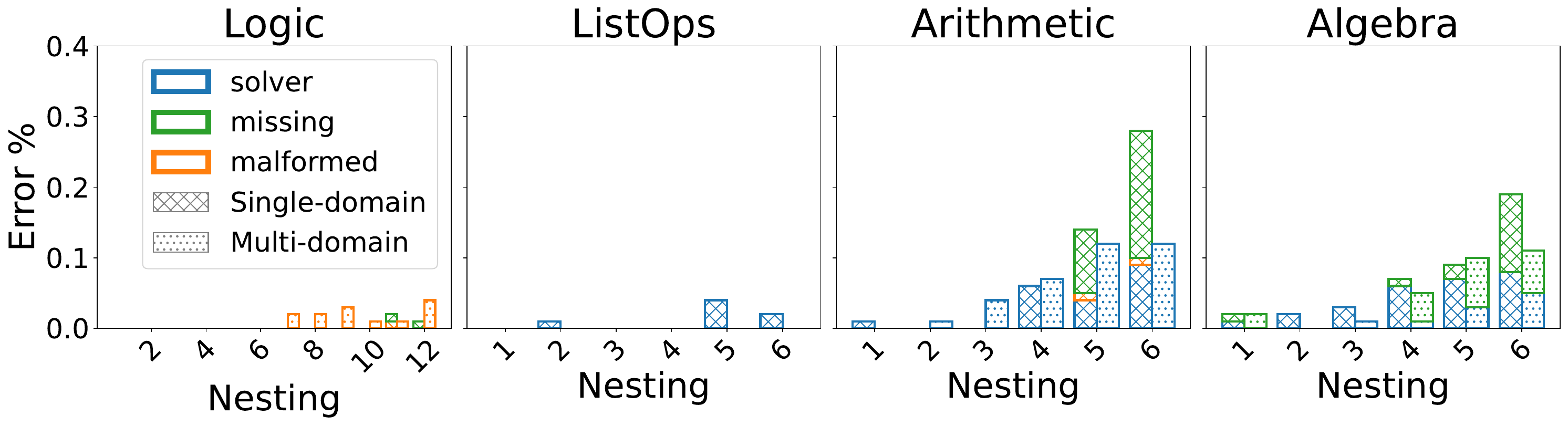}
    \caption{Breakdown of NRS errors by type in single- and  multi-domain settings.}
    \label{fig:errors_nrs}
\end{figure}

\begin{figure}
    \centering
    \includegraphics[width=\linewidth]{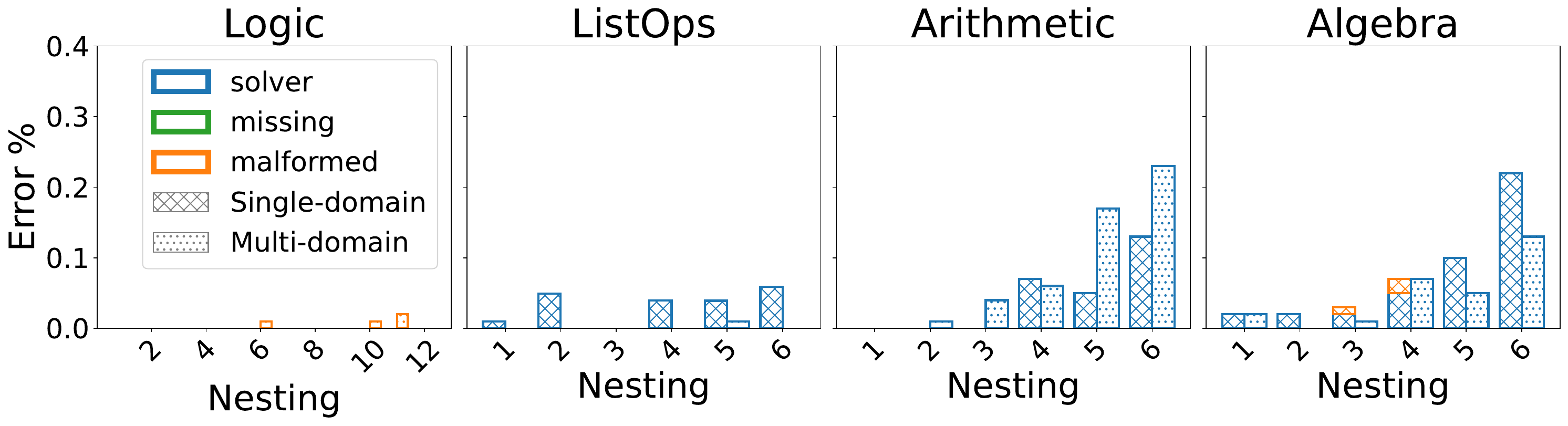}
        \caption{Breakdown of FastNRS errors by type in single- and  multi-domain settings.}
        \label{fig:errors_fastnrs}
\end{figure}

\subsection{Analysis of errors}
\label{sec:err-analys}
We analyze the errors committed by the Neural Rewriting Systems when simplifying mathematical formulas in the four domains.
We consider both the single-domain and multi-domain training settings. We breakdown errors for each domain by error type, and visualize the analysis in stacked barplots.
We consider three error cases: the one in which leaf formulas identified by the Selector are not present in the input formula (\texttt{missing}), the case in which they are present but they are not valid formulas (\texttt{malformed}), and the case in which they are solved incorrectly (\texttt{solver}).

We start by observing that there are no errors in the \texttt{missing} class in the case of FastNRS, which is expected given that we use a segmentation-based approach to the Selector task.
On the other hand, this type of error represents the majority of those committed by the Selector in the NRS, while errors in the \texttt{malformed} class are quite rare for both the NRS and the FastNRS.
Therefore, we can conclude that when Selector modules in both architectures identify leaf formulas that are present in the input, these tend to be well-formed.

The multi-domain setting reveals interesting positive effects, but also introduces challenges for both the NRS and FastNRS. 
In the case of NRS, multi-domain training seems to mostly have a neutral or positive effect across all tasks, significantly improving performance on algebraic and arithmetic formulas and reducing the amount of \texttt{missing} errors on the latter (see Figure \ref{fig:errors_nrs}). It seems that by training the Selector on multiple tasks, the NRS becomes more adept at identifying valid leaf formulas, reducing the number of this type of error.

The effects of multi-domain training on FastNRS are more heterogeneous and depend on the specific task. In the ListOps and algebra domains, the multi-domain model shows consistent improvements, with a marked reduction in Solver errors.
Interestingly, FastNRS performance worsens in the multi-domain setting on arithmetic formulas, where Solver errors increase.
This could reflect the fact that arithmetic formulas, involving operations between double-digit integers, are the hardest type of operation for the Solver.

Notably, in both single- and multi-domain scenarios, FastNRS never fails to find at least one valid leaf in any iteration, indicating robustness in the Selector module.

\subsection{Out-of-distribution generalization in the FastNRS Selector}
\label{sec:res-selector}
\begin{figure}
    \centering
    \begin{subfigure}[b]{0.355\textwidth}
     \centering
         \includegraphics[width=\textwidth]{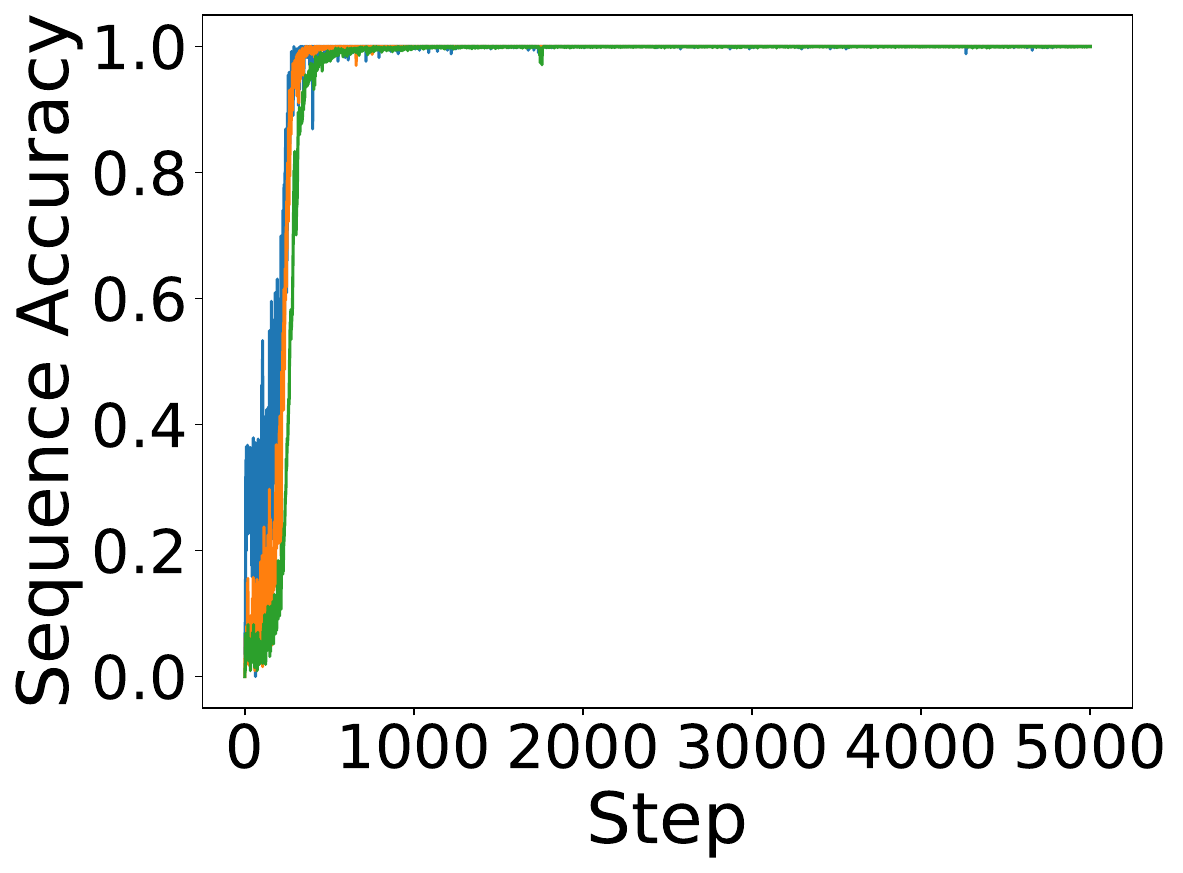}
         \caption{Logic}
         \label{fig:train-curve-logic}
    \end{subfigure}
    \begin{subfigure}[b]{0.30\textwidth}
     \centering
         \includegraphics[width=\textwidth]{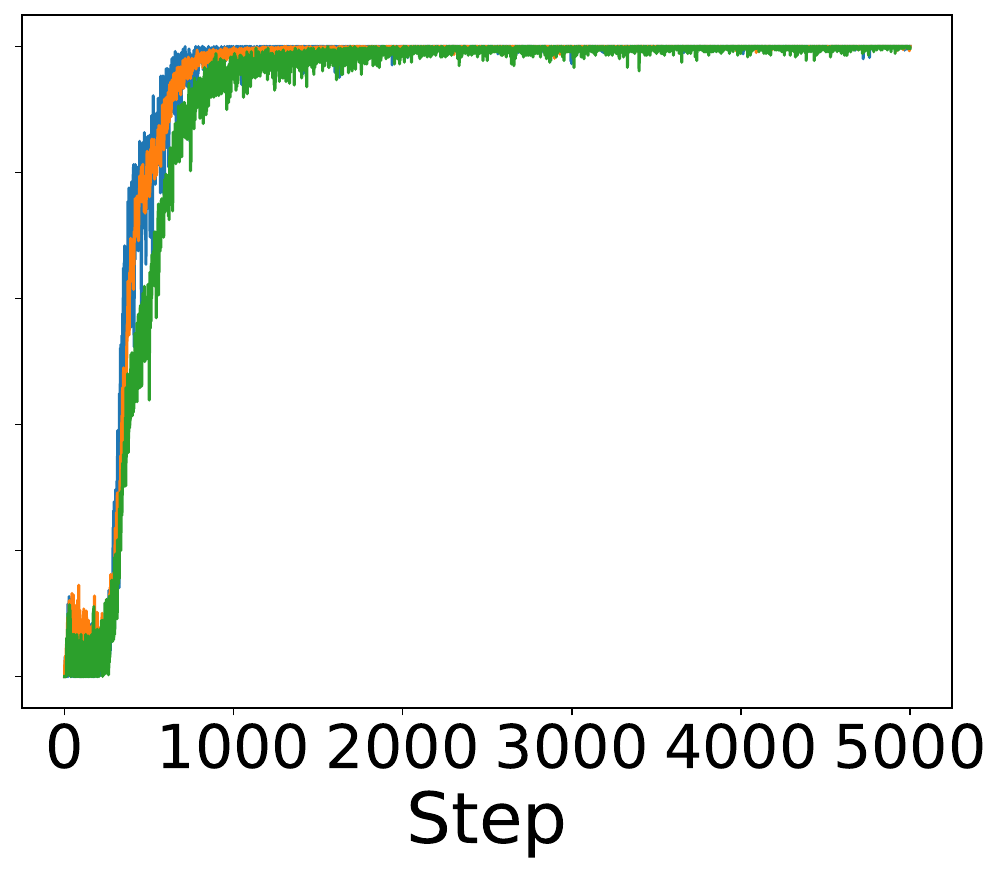}
         \caption{Listops}
         \label{fig:train-curve-listops}
    \end{subfigure}\\
    \begin{subfigure}[b]{0.355\textwidth}
     \centering
         \includegraphics[width=\textwidth]{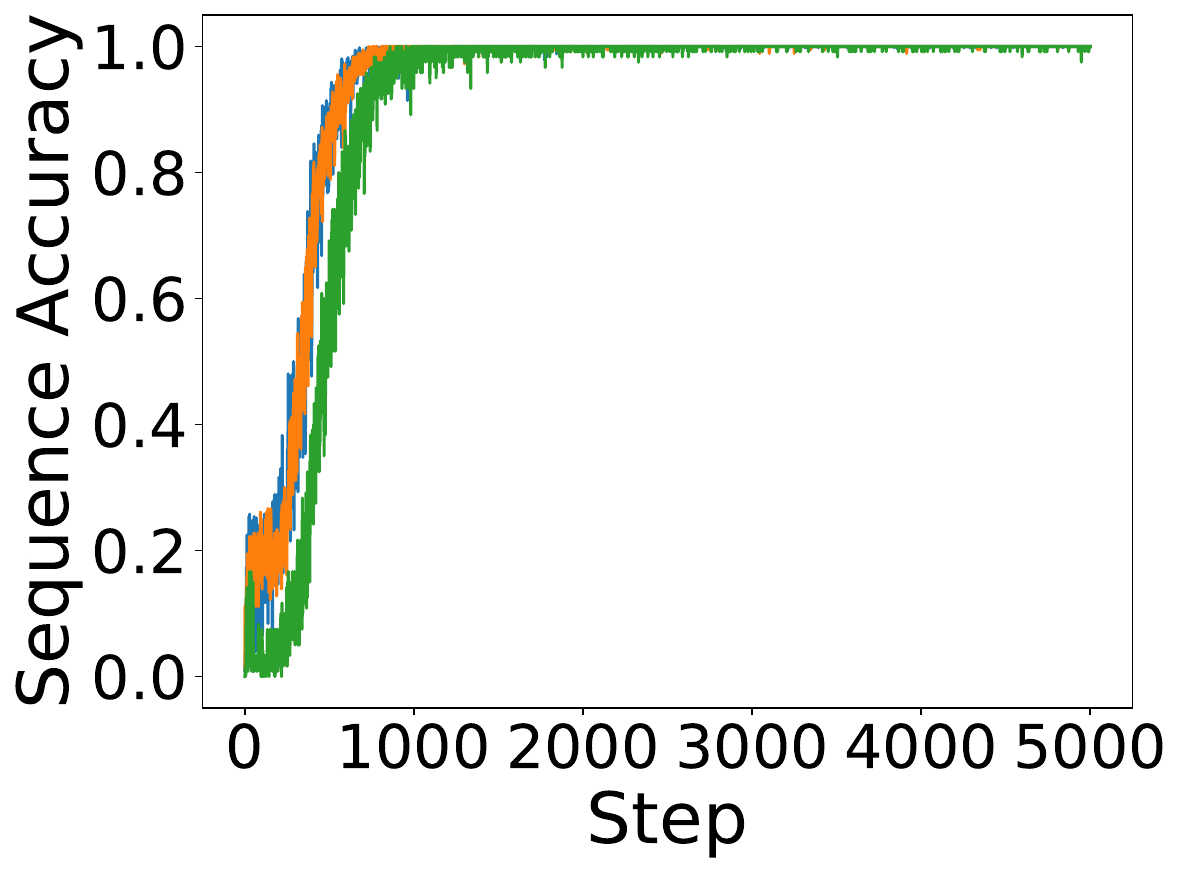}
         \caption{Arithmetic}
         \label{fig:train-curve-arithmetic}
    \end{subfigure}
    \begin{subfigure}[b]{0.30\textwidth}
     \centering
         \includegraphics[width=\textwidth]{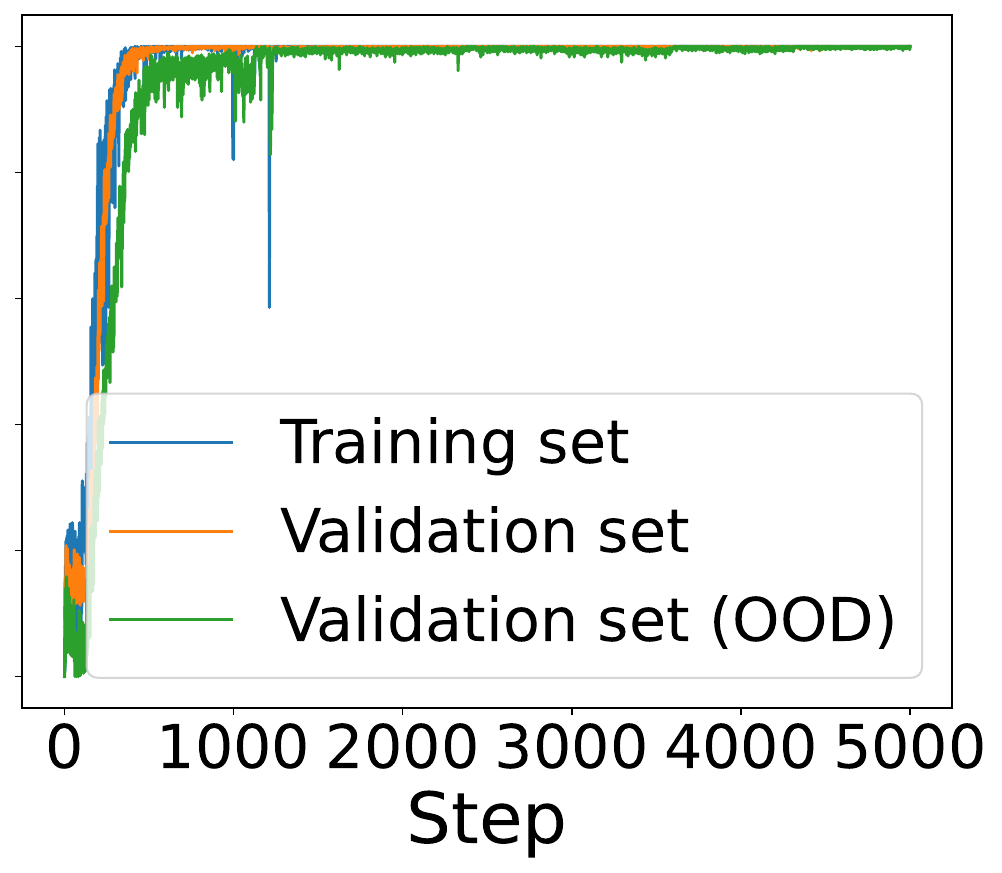}
         \caption{Algebra}
         \label{fig:train-curve-algebra}
    \end{subfigure}
    \caption{FastNRS Selector accuracy  during training on the text segmentation task.}
    \label{fig:train-curves}
\end{figure}

As we described in Section \ref{sec:selector}, the Selector module is a transformer encoder with two main architectural modifications: Label-based Positional Encodings and a strong limitation of the self-attention's receptive field.
Furthermore, we described how designing the Selector as a text segmentation module allowed us to simplify and improve the efficiency of the whole architecture.
The plots in Figure \ref{fig:train-curves} show how a Selector with the abovementioned architectural modifications, trained to segment input formulas, exhibits almost indistinguishable convergence trends on in- and out-of-distribution instances in all problems.
Carefully tuned Selector modules can thus segment an input formula several tens of tokens longer than formulas observed during training, with very limited or zero error rate.
Additional results about the impact of the width of the Selector's self-attention window and the number of layers in the model are reported in~\ref{sec:appendix-diagwidth}.

We also notice that the Selector is particularly sample efficient, as it requires only $5,000$ iterations to converge to almost perfect accuracy, while the best NRS Selector modules, chosen after hyperparameters tuning, could be trained for up to $30, 000$ iterations, depending on the task (see~\ref{sec:appendix-details-nrs}). We should also highlight that in all domains except ListOps, FastNRS Selector modules are several thousands of parameters smaller than their NRS counterparts in the same domain.

\section{Limitations}
Despite the significant generalization capabilities demonstrated by the framework we propose, its scope and applicability are constrained by several limitations.
First, the current implementation of NRS is restricted to tasks that can be framed as sequence-based rewriting problems. 
This assumption limits the range of tasks it can handle: for example, many real-world tasks involve hierarchical structures or visual reasoning, which cannot be addressed within the current sequence-only framework.
Second, rewriting rules must currently operate on local substrings of the input sequence, even in the FastNRS which identifies multiple substrings in parallel.
Finally, the NRS is built on an algorithmic structure where the steps of the rewriting process are predefined by the human designer. Although the system efficiently applies these predefined rules generalizing to more complex, unseen cases, it does not possess the capacity to learn or infer the rewriting algorithm from data.
Therefore, this design choice limits the possibility of applying the system to new problems beyond the algorithmic template initially provided.

\section{Conclusions}
\label{sec:conclusions}
In this work, we presented a general framework for learning convergent term rewriting systems using a neuro-symbolic architecture, inspired directly by the rewriting algorithm itself. Within this framework, we introduced two distinct implementations: the Neural Rewriting System (NRS) and the Fast Neural Rewriting System (FastNRS). Both architectures are designed to learn and generalize across the class of problems solvable with convergent term rewriting systems. The FastNRS, in particular, builds upon the NRS by incorporating key modifications that significantly improve memory efficiency, training time, and inference speed.

We evaluated both the NRS and FastNRS in both single-domain and multi-domain testing scenarios. In the multi-domain scenario, a single model is trained across multiple datasets or problem types simultaneously, resulting in a system that can solve various tasks within the same architecture. Using datasets such as Logic, ListOps, Arithmetic, and Algebra, we showed that both models consistently prove strong generalization capabilities across tasks, and that the FastNRS offers substantial reductions in computational costs.

We compared the Neural Rewriting Systems trained in a single-domain scenario with the Neural Data Router as a representative of small-scale neural architectures specialized to learn single reasoning tasks. Our systems clearly outperformed the baseline, especially on out-of-distribution samples.

We further compared both systems trained in a multi-domain scenario against two general-purpose Large Language Models: OpenAI's GPT-4 and the recently presented o1-preview model designed to excel in complex reasoning tasks.
Although the performance of our models on the most complex formulas consistently surpassed that of GPT-4 on the same problems, o1-preview showed surprising capabilities of solving even very complex formulas with a relatively high degree of accuracy.
While o1-preview outperformed the Neural Rewriting Systems on some out-of-distribution formulas of intermediate complexity in some tasks, the models we propose consistently achieved equal or higher accuracy on the hardest formulas in all tasks, demonstrating significantly higher systematic generalization capabilities.
The drop in performance of o1-preview on complex problems might suggest a fundamental lack of systematic reasoning capabilities and understanding of mathematical concepts still persistent in this new class of models, as also noted in recent work by \cite{DBLP:journals/corr/abs-2410-05229}.

The strengths of our architecture can be traced back to its modular design, which is informed by the rewrite algorithm, and to the architectural modifications to the transformer, which have proven effective in enabling strong out-of-distribution generalization.
However, these design choices also limit the scope of applicability of our system to sequence-based problems solvable by convergent term rewriting systems with local substitution rules.
Future work could be dedicated to expanding the system to handle rules that could act on patterns across different parts of the sequence. 
This would involve rethinking the Selector, where it would be necessary to design a neural circuit that is capable of generalizing on the selection of non-local patterns as consistently as the current circuit does with local ones.
Furthermore, the substitution mechanism should be designed to be capable of reliably replacing the global patterns, potentially maintaining a level of flexibility to noise and errors in the Selector output.
As we previously noticed, the algorithmic-informed design of our systems, which guarantees robustness, is defined \textit{a priori} rather than being learned from data, and thus limits their scope of applicability.
Designing an end-to-end learned system where the algorithmic blueprint of the problem at hand is inferred directly from data could prove to be a challenging and interesting venue for future research.
While defining the algorithmic blueprint for a specific class of problems in advance imposes on the system a strong inductive bias that is aligned to the class itself, a general-purpose framework for algorithmic learning would involve designing a learning bias that allows the system to dynamically align to the specific class of problems under consideration.

\section{Acknowledgements}
The authors wish to thank OpenAI for granting free research access to the GPT-4 and o1-preview APIs.
OpenAI had no involvement in the study design, collection, analysis and interpretation of data, writing of the report, or the decision to submit the article for publication.

\section{Funding}
This research did not receive any specific grant from funding agencies in the public, commercial, or not-for-profit sectors.

\newpage

\appendix
\section{Dataset statistics}
\label{app:dataset-statistics}
As described in Section \ref{sec:models-nrs}, the development sets for the Neural Rewriting System (NRS) across the four datasets—logic, listops, arithmetic, and algebra—were constructed to capture a diverse range of formula complexities. Each development set is composed of multiple subsets with varying nesting levels, alongside intermediate formulas generated during the resolution process of the main ones.
By design, the number of unique formulas available in splits with simpler formulas is smaller than in splits with more complex ones, due to the combinatorial nature of the problem.
Despite these differences in the number of unique formulas per split, during training, samples are drawn from each split with equal probability. This ensures balanced exposure across the splits, allowing the model to generalize across different formula complexities.
Refer to Tables \ref{tab:data_selector} and \ref{tab:data_solver} for the exact number of samples in each development split for the four tasks. We report in Table \ref{tab:data_test} the number of (unique) samples in the test sets of the four tasks used across all experiments.

\begin{table}[ht]
    \centering
    \begin{tabular}{l|r|r|r}
    Task & Training set & ID validation set & OOD validation set \\
    \hline
    Logic     & 238436& 710& 900\\
    ListOps     & 840209& 2332& 900\\
    Arithmetic     & 399218& 180& 60\\
    Algebra     & 323787& 900& 300\\
    \end{tabular}
    \caption{No. of unique samples in the NRS and FastNRS development sets. ID and OOD indicate in-distribution and out-of-distribution sets, respectively.}
    \label{tab:data_selector}
\end{table}

\begin{table}[ht]
    \centering
    \begin{tabular}{l|r|r}
    Task & Training set & ID validation set\\
    \hline
    Logic     & 155& 45\\
    ListOps     & 22158& 5541\\
    Arithmetic     & 30510& 7628\\
    Algebra     & 18992& 4749\\
    \end{tabular}
    \caption{Number of unique samples in the NRS and FastNRS Solver development sets. ID indicates the in-distribution set.}
    \label{tab:data_solver}
\end{table}

\begin{table}[ht]
    \centering
    \begin{tabular}{l|r}
    Task & Num. samples\\
    \hline
    Logic     & 1200\\
    ListOps     & 600\\
    Arithmetic     & 600\\
    Algebra     & 600\\
    \end{tabular}
    \caption{Number of unique samples in the test sets}
    \label{tab:data_test}
\end{table}

\section{Training details}
\label{app:train-details}
\subsection{Fast Neural Rewriting System}
\label{sec:appendix-details-fastnrs}

\begin{table}[]
    \centering
    \begin{tabular}{l|r|r|r|r|r}
        & Logic & ListOps & Arithm. & Algebra & MD \\
        \hline
        Embedding size &256 & 256 & 256 & 256 & 256 \\
        Num. Enc. Layers &3 & 4 & 4 & 6 & 4 \\
        Width self-attn window &1 & 1 & 1 & 1 & 1 \\
        Learning rate &3.55e-05 & 3.65e-05 & 2.66e-05 & 4.49e-05 & 1.69e-05 \\
    \end{tabular}
    \caption{FastNRS Selector tuned hyperparameters values for each  test scenario. MD indicates the multi-domain scenario.}
    \label{tab:fastnrs-sel-hyp}
\end{table}

{\bf Selector} For all problems, we adopted a problem-dependent tokenizer whose vocabulary contains atomic values, operators and parentheses.
For example, the vocabulary for the arithmetic problem contains one token for each single- or double-digit integer, tokens for the sum, subtraction and multiplication operators and tokens for open and closed parentheses.
In preliminary experiments, we also tried using a character-level tokenizer but observed worse out-of-distribution generalization capabilities of the Selector in some domains.

In all models, we used four attention heads and a hidden state in the feed-forward layers that was four times larger than the embedding size.
We trained the models using the Adam optimizer with default parameters, a batch size of 512, a dropout probability of 10\% and a cosine annealing schedule of the learning rate with 1000 linear warm-up iterations.
We tuned the embedding size, the number of encoder layers, the width of the diagonal window applied to the self-attention matrix and the learning rate using a random search.
For all tasks, we searched hyperparameters values in the following ranges: \{128, 256, 512\} for the embedding size, [1, 9] for the number of encoder layers and [1e-6, 6e-6] for the learning rate.
All models were trained using the Adam optimizer for 5,000 iterations, apart from the multi-domain model which was trained for 7,000 iterations.
The final values chosen after tuning each hyperparameter are reported in Table \ref{tab:fastnrs-sel-hyp}.

\begin{table}[]
    \centering
    \begin{tabular}{l|r|r|r|r|r}
        & Logic & ListOps & Arithm. & Algebra & MD \\
        \hline
        Embedding size &64 & 128 & 256 & 256 & 320 \\
        Num. Enc. Layers &1 & 2 & 3 & 2 & 4 \\
        Num. Dec. Layers &1 & 2 & 3 & 2 & 4 \\
        Dropout &0.18 & 0.18 & 0.1 & 0.33 & 0.13 \\
        Learning rate &9.23e-05 & 9.59e-05 & 9e-05 & 8e-05 & 6.19e-05 \\
        Warm-up it. &1282 & 1910 & 1500 & 1500 & 1714 \\
    \end{tabular}
    \caption{FastNRS Solver tuned hyperparameters values for each test scenario. MD indicates the multi-domain scenario.}
    \label{tab:fastnrs-sol-hyp}
\end{table}

{\bf Solver} We used a simple character-level tokenizer for all problems. We tuned the hyperparameters of the Solver using a random search on the embedding size, the number of encoder and decoder layers, the dropout rate, and the learning rate.
In all models, we used four attention heads and a hidden state in the feed-forward layers that was four times larger than the embedding size.
We trained the models using the Adam optimizer with default parameters, a batch size of 512 and a cosine annealing schedule of the learning rate.
The models were trained for 10,000 iterations in the case of Logic and ListOps tasks and for 40,000 and 100,000 iterations in the case of Algebra and Arithmetic tasks, respectively.
For all tasks, we searched hyperparameters values in the following ranges: \{64, 128, 256\} for the embedding size, [1, 4] for the number of encoder and decoder layers, [0.1, 0.4] for the dropout probability, [1e-5, 1e-4] for the learning rate and [1000, 2000] for the number of warmup iterations.
The final values chosen after tuning each hyperparameter are reported in Table \ref{tab:fastnrs-sol-hyp}.

\subsubsection{Selector Depth and Width of the Self-Attention Window}
\label{sec:appendix-diagwidth}

\begin{figure}
    \centering
    \begin{subfigure}[b]{0.24\textwidth}
     \centering
         \includegraphics[width=\textwidth]{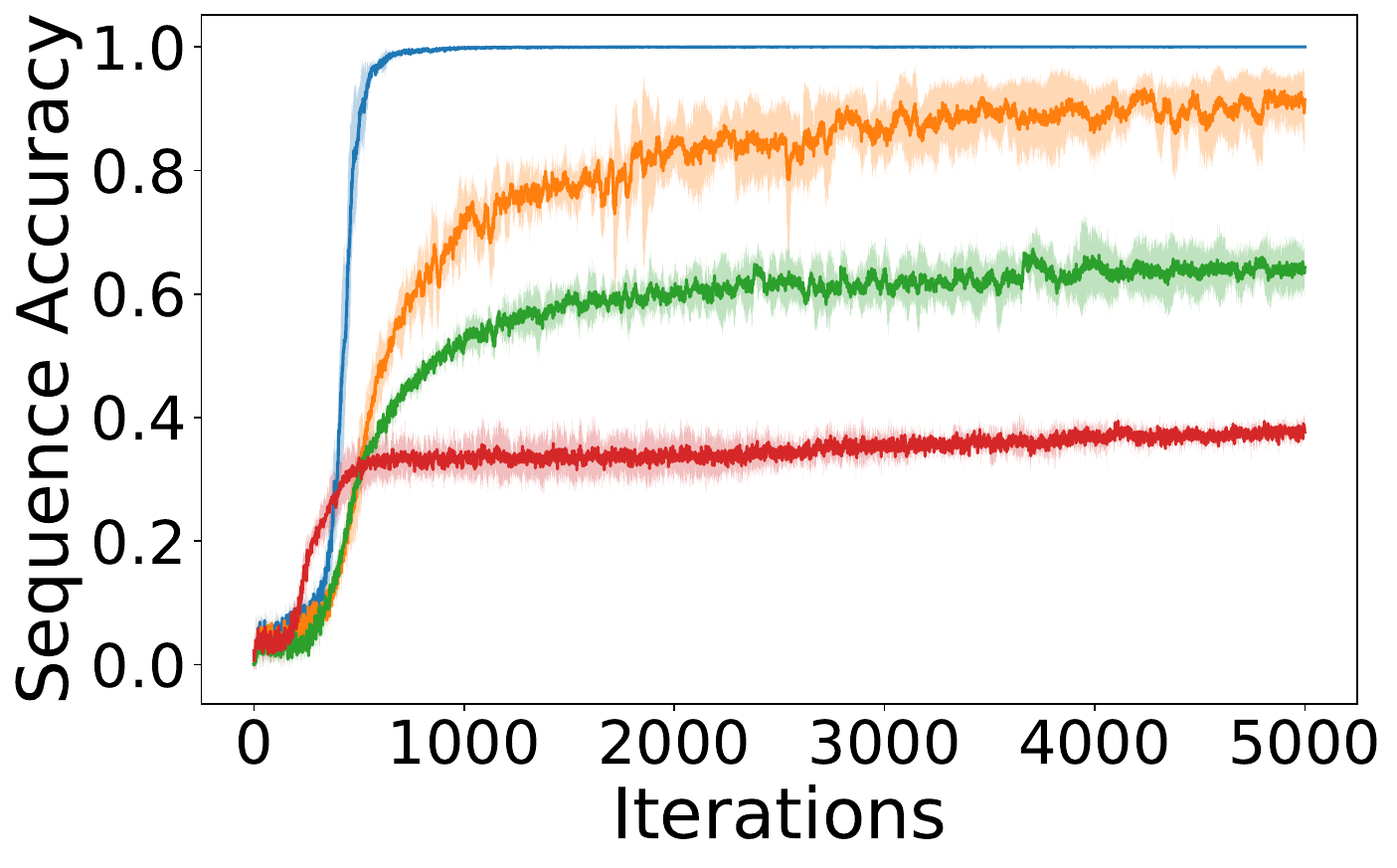}
         \caption{Logic}
         
    \end{subfigure}
    \begin{subfigure}[b]{0.225\textwidth}
     \centering
         \includegraphics[width=\textwidth]{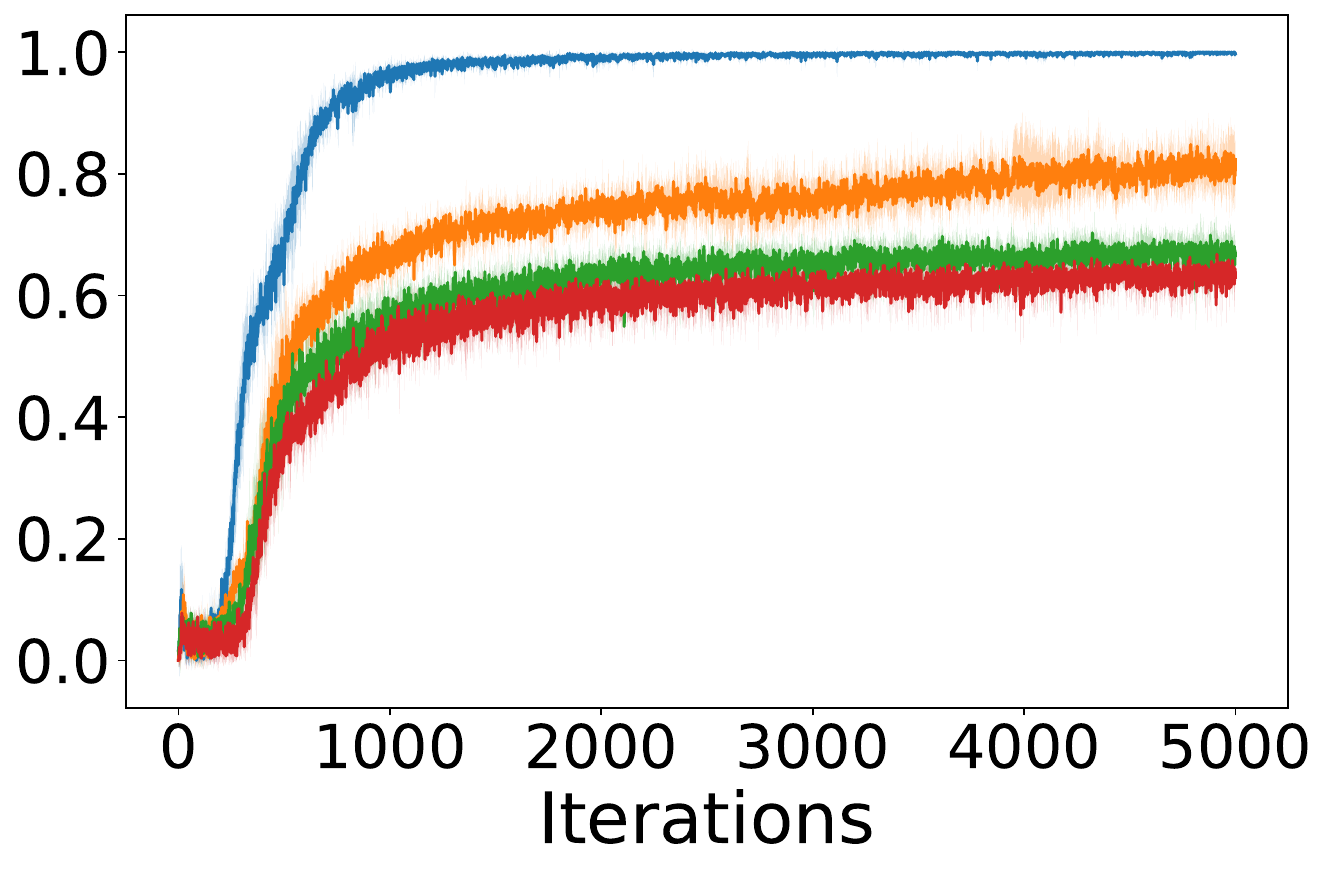}
         \caption{Listops}
         
    \end{subfigure}
    \begin{subfigure}[b]{0.225\textwidth}
     \centering
         
         \includegraphics[width=\textwidth]{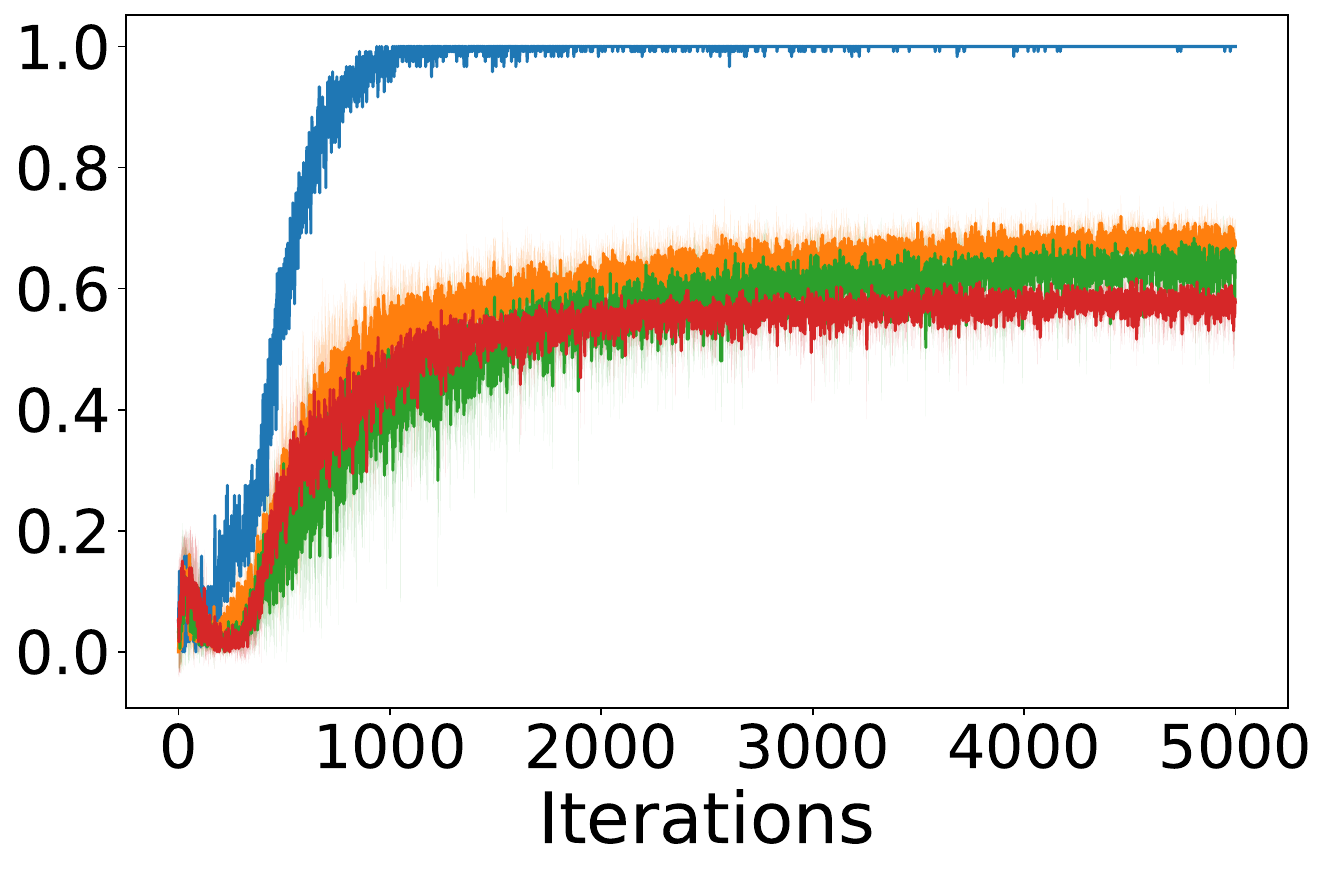}
         \caption{Arithmetic}
         
    \end{subfigure}
    \begin{subfigure}[b]{0.225\textwidth}
     \centering
         \includegraphics[width=\textwidth]{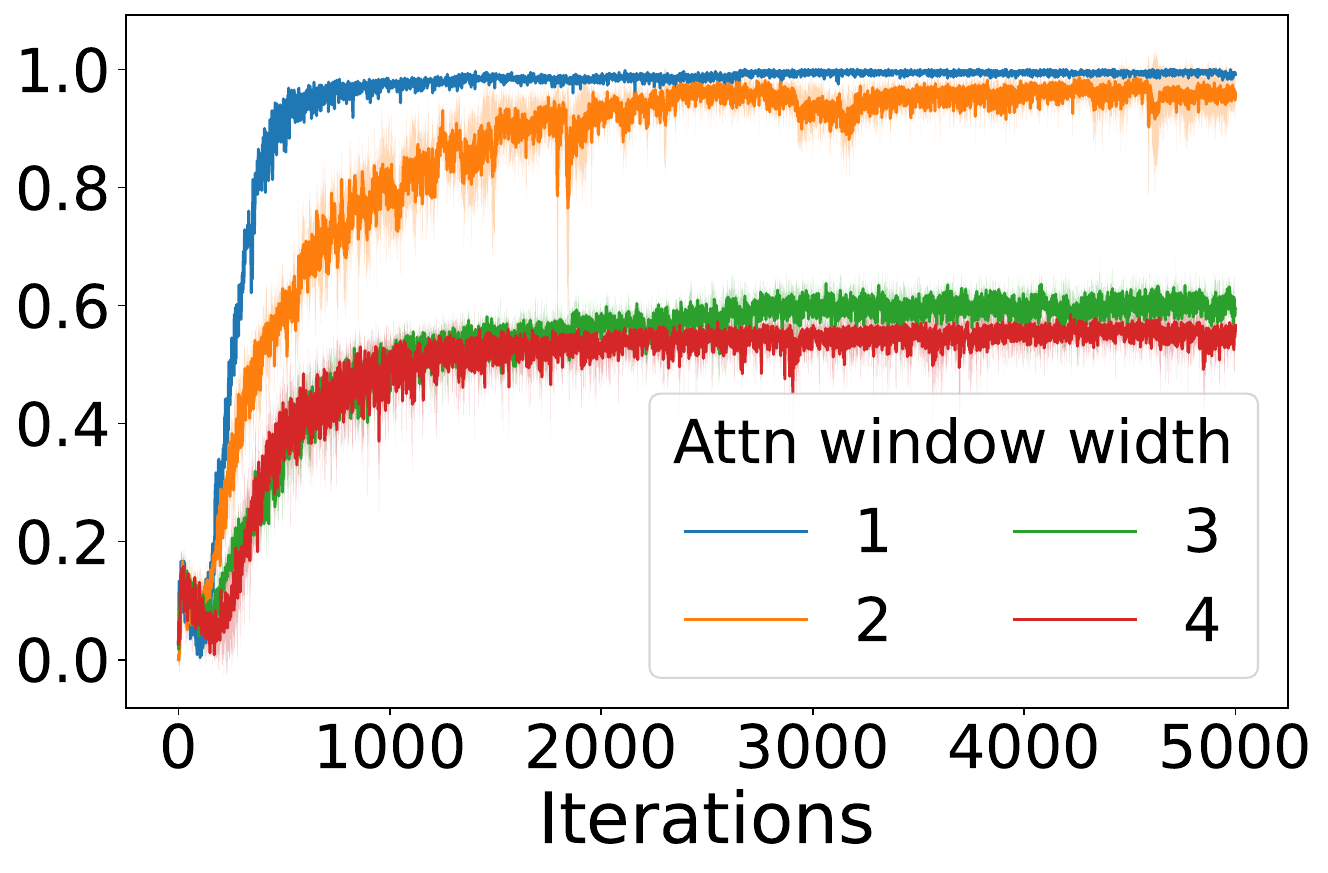}
         \caption{Algebra}
         
    \end{subfigure}
    \caption{Impact of self-attention window width on out-of-distribution sequence accuracy.}
    \label{fig:diagwith-seqacc}
\end{figure}

\begin{figure}
    \centering
    \begin{subfigure}[b]{0.24\textwidth}
     \centering
         \includegraphics[width=\textwidth]{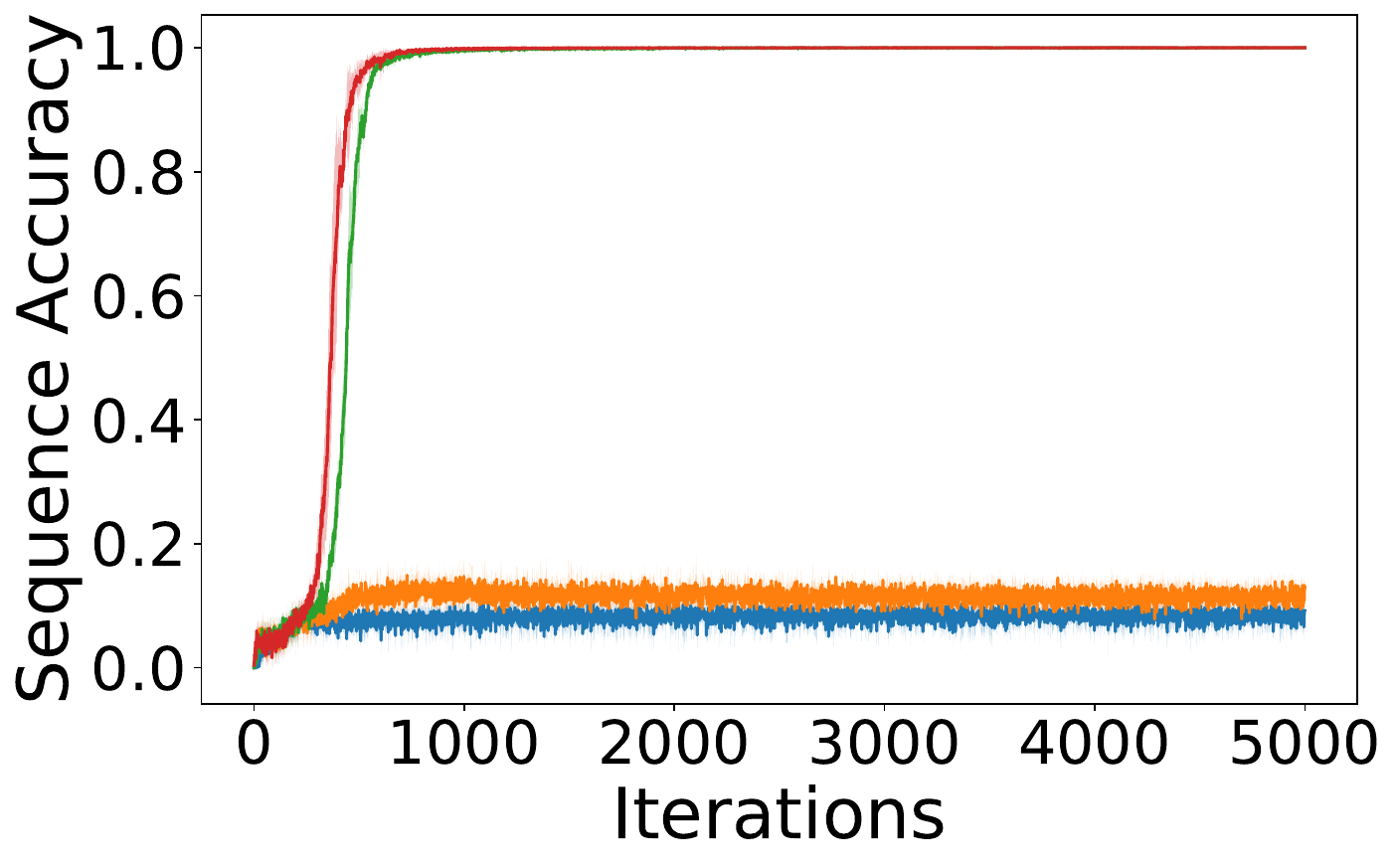}
         \caption{Logic}
         
    \end{subfigure}
    \begin{subfigure}[b]{0.225\textwidth}
     \centering
         \includegraphics[width=\textwidth]{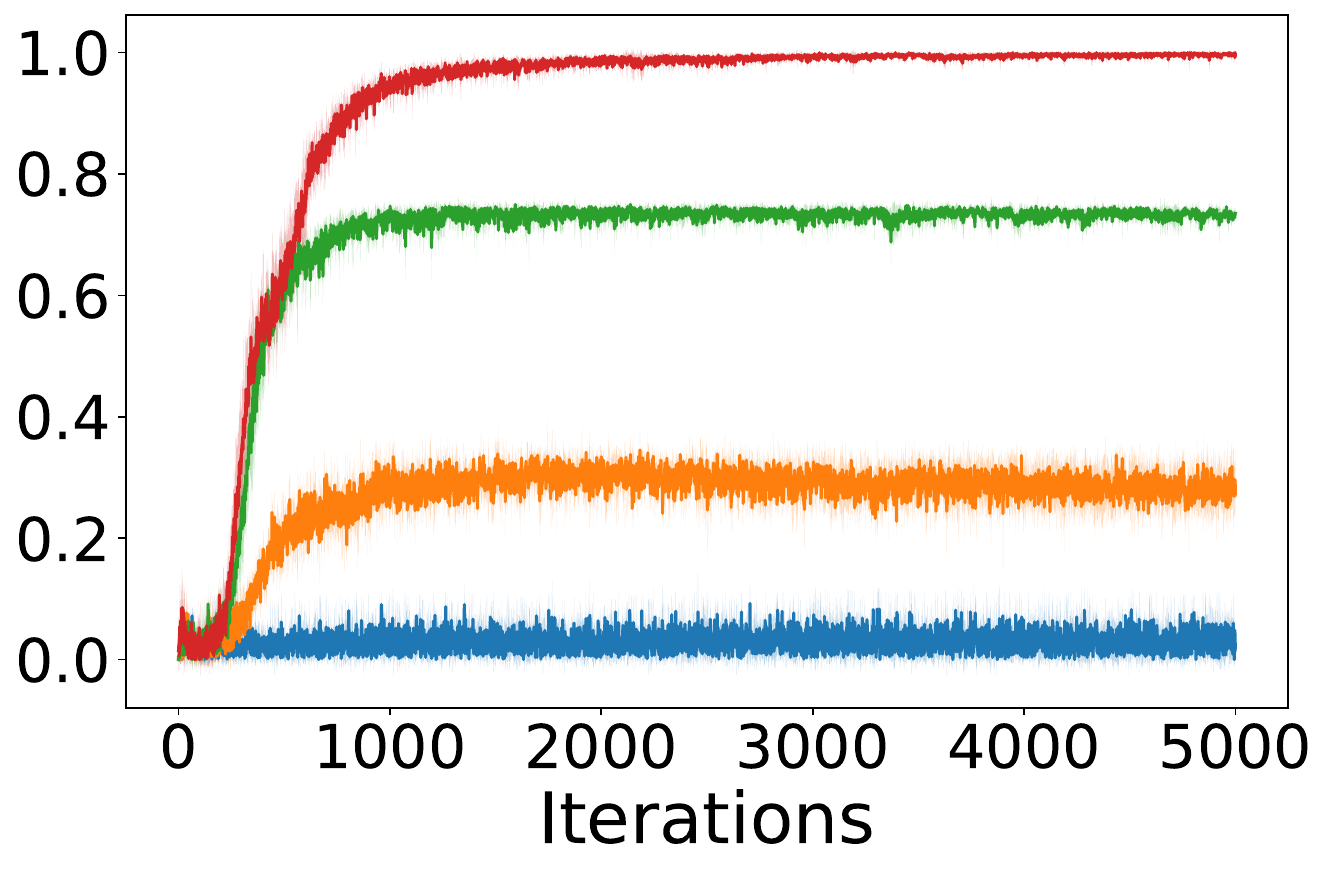}
         \caption{Listops}
         
    \end{subfigure}
    \begin{subfigure}[b]{0.225\textwidth}
     \centering
         
         \includegraphics[width=\textwidth]{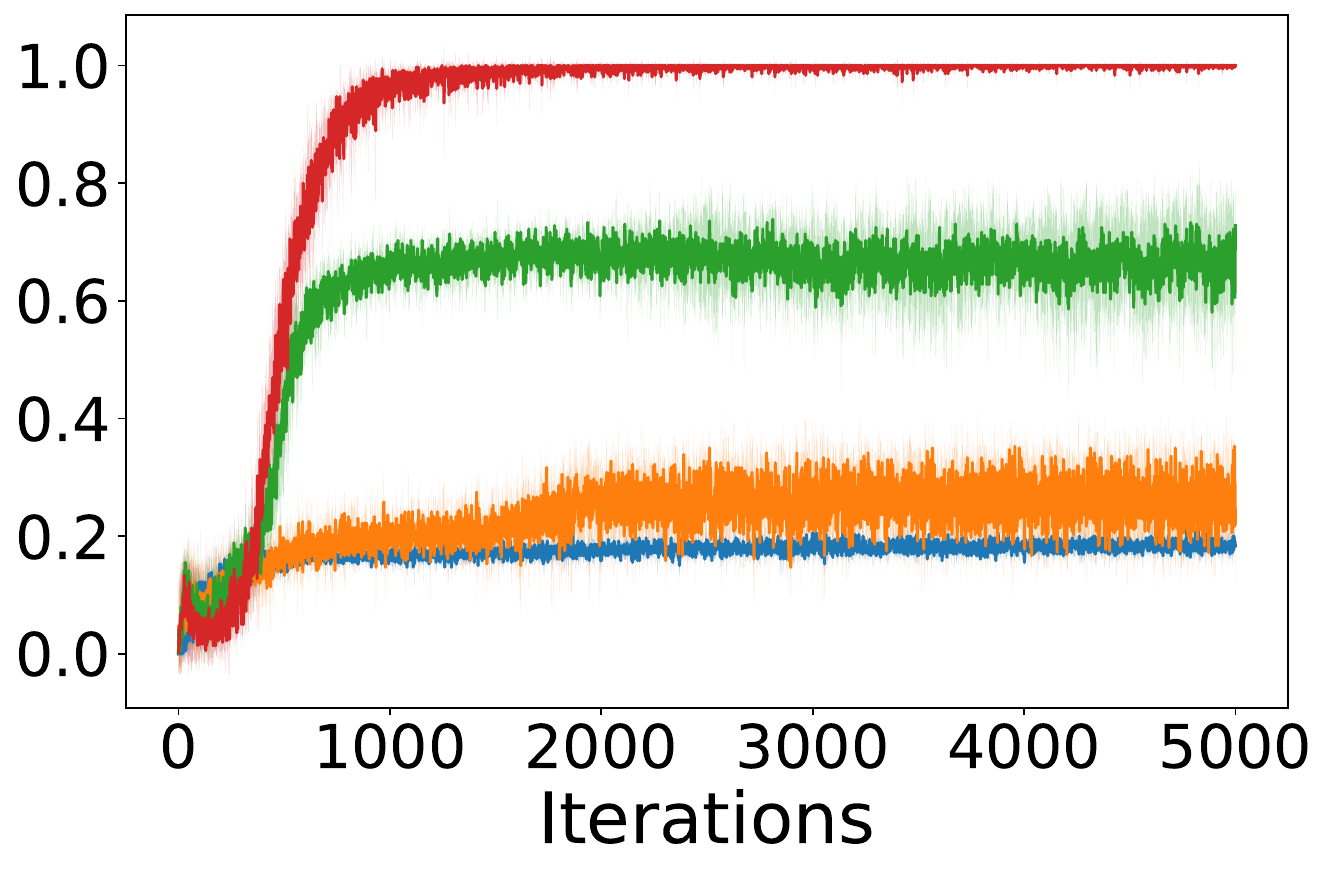}
         \caption{Arithmetic}
         
    \end{subfigure}
    \begin{subfigure}[b]{0.225\textwidth}
     \centering
         \includegraphics[width=\textwidth]{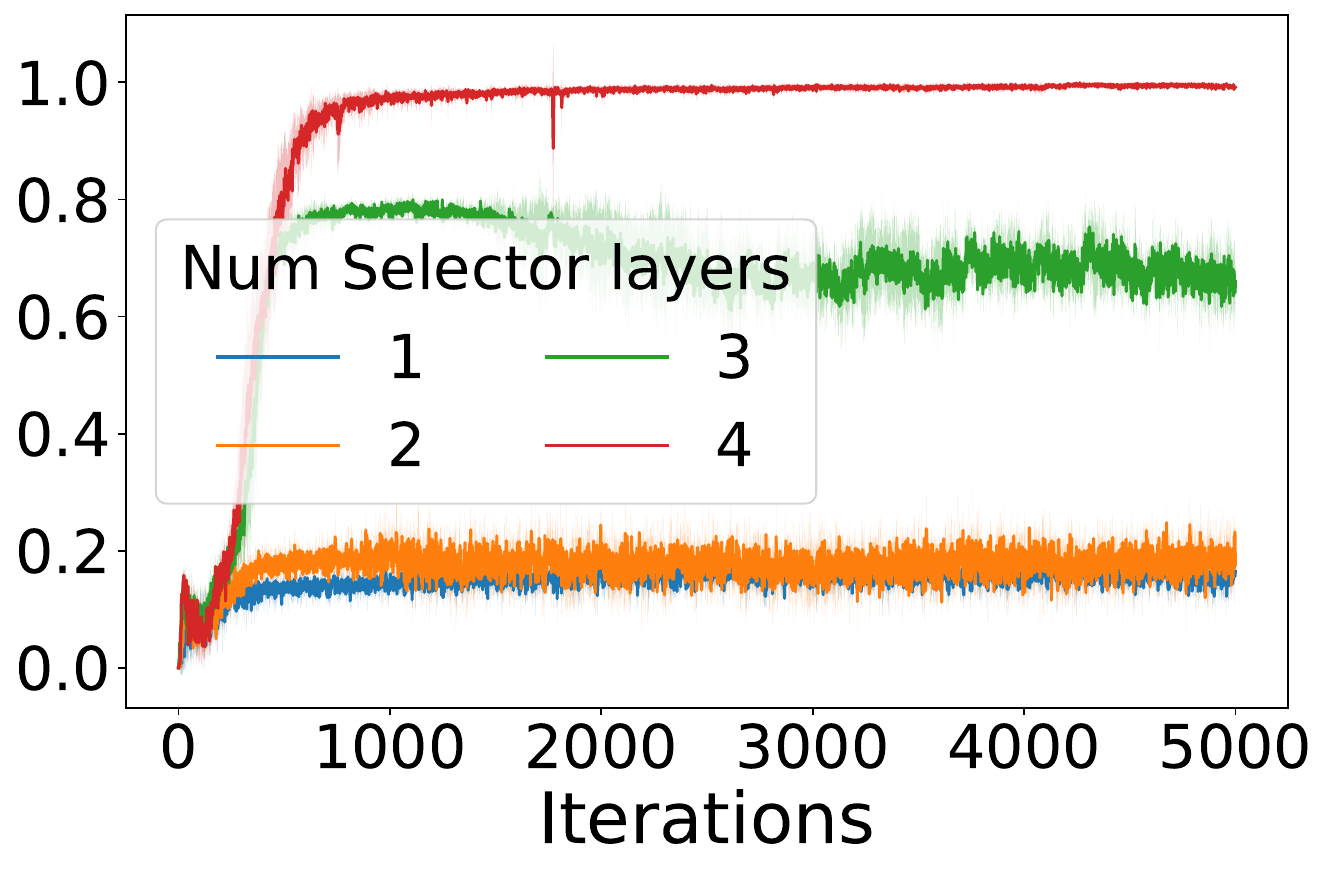}
         \caption{Algebra}
         
    \end{subfigure}
    \caption{Impact of number of layers on out-of-distribution sequence accuracy.}
    \label{fig:encdepth-seqacc}
\end{figure}

Figures \ref{fig:diagwith-seqacc} and \ref{fig:encdepth-seqacc} illustrate the impact of both the width of the self-attention diagonal window and the number of layers in the FastNRS Selector model on sequence accuracy, measured on an out-of-distribution set of samples during training. We report the mean and standard deviation of a group of three runs with different random seeds. As shown in the plots, there is a clear inverse relationship between the width of the self-attention window and model performance, consistent across all the domains we consider. The width of the self-attention window is expressed in terms of the hyperparameter $k$. Specifically, as the window width increases, the model becomes progressively less capable of generalizing on out-of-distribution samples. Peak performance, indicated by the highest accuracy, is achieved when the hyperparameter $k$ is set to 1, suggesting that a narrower focus in self-attention enhances the model's ability to generalize to out-of-distribution data.

In contrast, there is a direct relationship between the number of layers and model performance: as the number of layers increases, the model's generalization ability improves, leading to better accuracy on both in-distribution and out-of-distribution samples. 
We also notice that all the models with different window widths were equally able to fit the training set with no sign of overfitting on the in-distribution validation set.
On the contrary, models with varying numbers of layers showed similar performance across the training, in-distribution, and out-of-distribution validation sets (not shown).
Both analyses use model hyperparameters that were selected through hyperparameter tuning and are consistent with those applied in the rest of the experiments.

\subsection{Neural Rewriting System}
\label{sec:appendix-details-nrs}
{\bf Selector} As done with the FastNRS, we employ a problem-dependent tokenizer in the NRS Selector. In all models we used four attention heads and a hidden state in the feed-forward layers that was four times larger than the embedding size.
Selector models were trained for 20,000 iterations for the Logic and ListOps tasks, and 30,000 iterations for the Arithmetic and Algebra tasks.
We trained the models using the Adam optimizer with default parameters, a batch size of 512 (256 for Algebra) and a cosine annealing schedule of the learning rate with warm-up.
We tuned the embedding size, the number of encoder and decoder layers, the width of the diagonal window applied to the self-attention matrix, the dropout rate, the learning rate, the number of warm-up iterations and the value of gain parameter for initialization of the self-attention layers using a random search.
For all tasks, we searched hyperparameters values in the following ranges: \{128, 256, 512\} for the embedding size, [1, 3] for the width of the diagonal self-attention window, [2, 5] for the number of encoder and decoder layers, [0.1, 0.4] for the dropout probability, [1e-5, 6e-5] for the learning rate, [500, 3000] for the number of warm-up iterations, [0.5, 2.5] for the initialization gain parameter.
The final values chosen after tuning each hyperparameter are reported in Table \ref{tab:nrs-sel-hyp}.

\begin{table}[]
    \centering
    \begin{tabular}{l|r|r|r|r|r}
        & Logic & ListOps & Arithm. & Algebra & MD \\
        \hline
        Embedding size &256 & 256 & 256 & 256 & 256 \\
        Width self-attn window &2 & 2 & 3 & 3 & 2 \\
        Num. Enc. Layers &1 & 1 & 3 & 4 & 5 \\
        Num. Dec. Layers &2 & 2 & 2 & 2 & 2 \\
        Dropout &0.29 & 0.37 & 0.17 & 0.20 & 0.10 \\
        Learning rate &2.7e-5 & 2.65e-5 & 2.35e-5 & 5.54e-5 & 7.86e-5 \\
        Warm-up it. &1600 & 1700 & 1900 & 2900 & 1500 \\
        MHA init. gain &0.97 & 0.71 & 1.69 & 0.75 & 1.00 \\
    \end{tabular}
    \caption{NRS Selector tuned hyperparameters values for each  test scenario. MD indicates the multi-domain scenario.}
    \label{tab:nrs-sel-hyp}
\end{table}

{\bf Solver} In our experiments we used the same Solver modules both in the FastNRS and in the NRS, thus the tokenization method and hyperparameters for the NRS Solver correspond to those detailed for the FastNRS Solver in Section \ref{sec:appendix-details-fastnrs} and Table \ref{tab:fastnrs-sol-hyp}.

\subsection{Neural Data Router}
\label{sec:appendix-details-ndr}
\begin{table}[]
    \centering
    \begin{tabular}{l|r|r|r|r}
        & Logic & ListOps & Arithmetic & Algebra \\
        \hline
        Num. Enc. Layers &19 & 5 & 11 & 17 \\
        Embedding size &256 & 512 & 512 & 512 \\
        Attention heads &8 & 16 & 8 & 16 \\
        FF size &1024 & 1024 & 2048 & 2048 \\
        Learning rate &2.33e-04 & 4.13e-04 & 7.64e-04 & 9.59e-04 \\
        Dropout &0.45 & 0.09 & 0.05 & 0.40 \\
        Attention dropout &0.38 & 0.49 & 0.06 & 0.18 \\
        Weight decay &0.02 & 0.09 & 0.08 & 0.03 \\
    \end{tabular}
    \caption{NDR tuned hyperparameters values for each task.}
    \label{tab:ndr-hyp}
\end{table}
As done with the Neural Rewriting Systems, we employ a problem-dependent tokenizer at the atomic value level when training the Neural Data Router. Therefore, the size of the result window $k$ equals 3 in the case of Algebra, 2 for Arithmetic and 1 for ListOps and Logic problems.

Here we report the hyperparameters of the best Neural Data Router configuration we selected for each problem.
We searched hyperparameters values in the same ranges used in the original paper.
Models were trained using the AdamW optimizer for 5,000 iterations in the case of Logic, 30,000 iterations in the case of ListOps, and 100,000 iterations in the case of Arithmetic and Algebra.
We used a batch size of 512 for all task except Algebra, for which it was 256.
The final values chosen after tuning are reported in Table \ref{tab:ndr-hyp}.

\section{NRS Selector confidence scores}
\label{appendix-conf-scores}
\begin{figure}[ht]
    \begin{minipage}[c]{0.48\linewidth}
        \includegraphics[width=\linewidth]{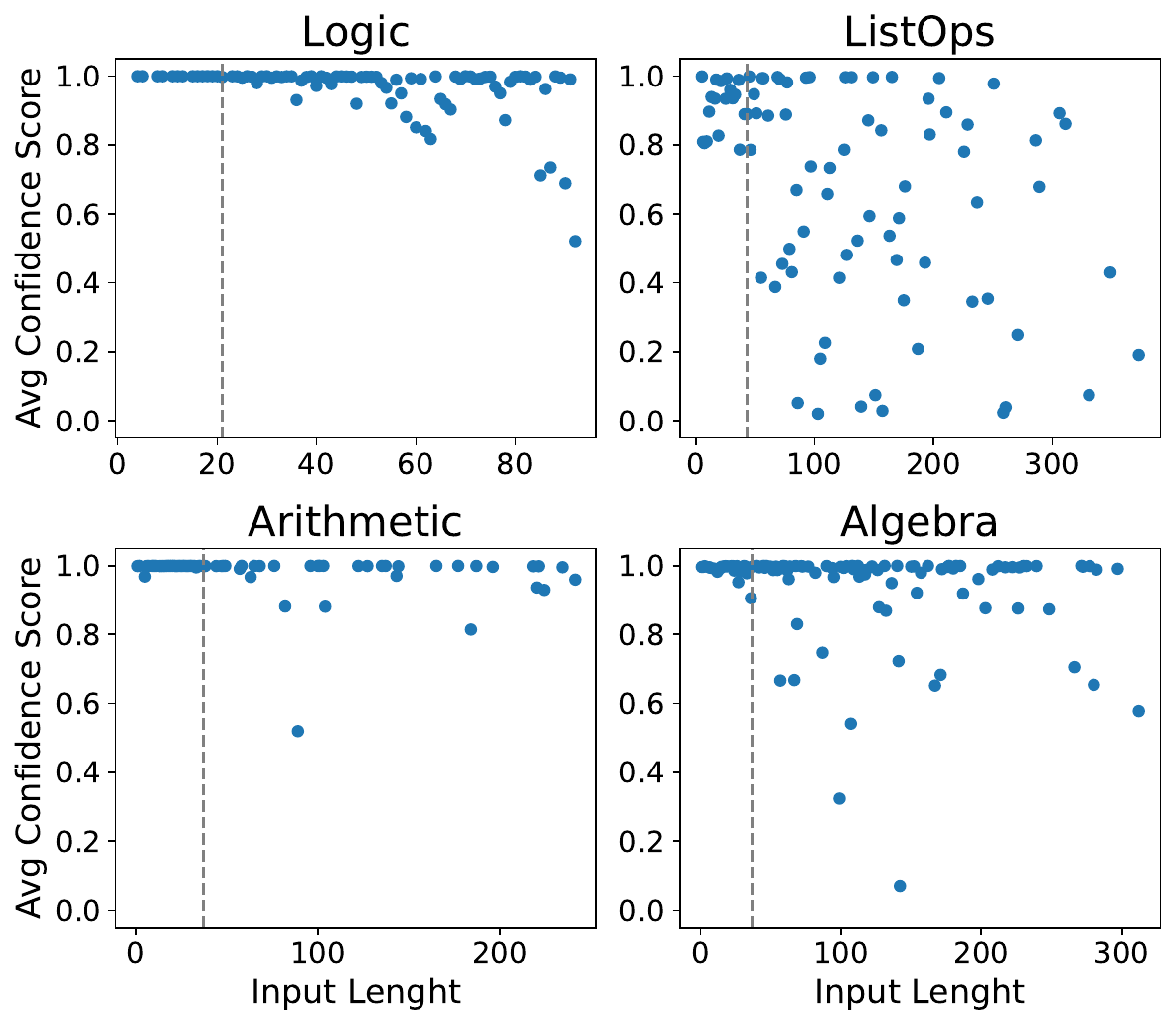}
        \caption{Average single-domain NRS Selector confidence scores by input length. The vertical line represents the maximum length of training formulas.}
        \label{fig:conf_scores_nrs}
    \end{minipage}
    \hfill
    \begin{minipage}[c]{0.48\linewidth}
        \includegraphics[width=\linewidth]{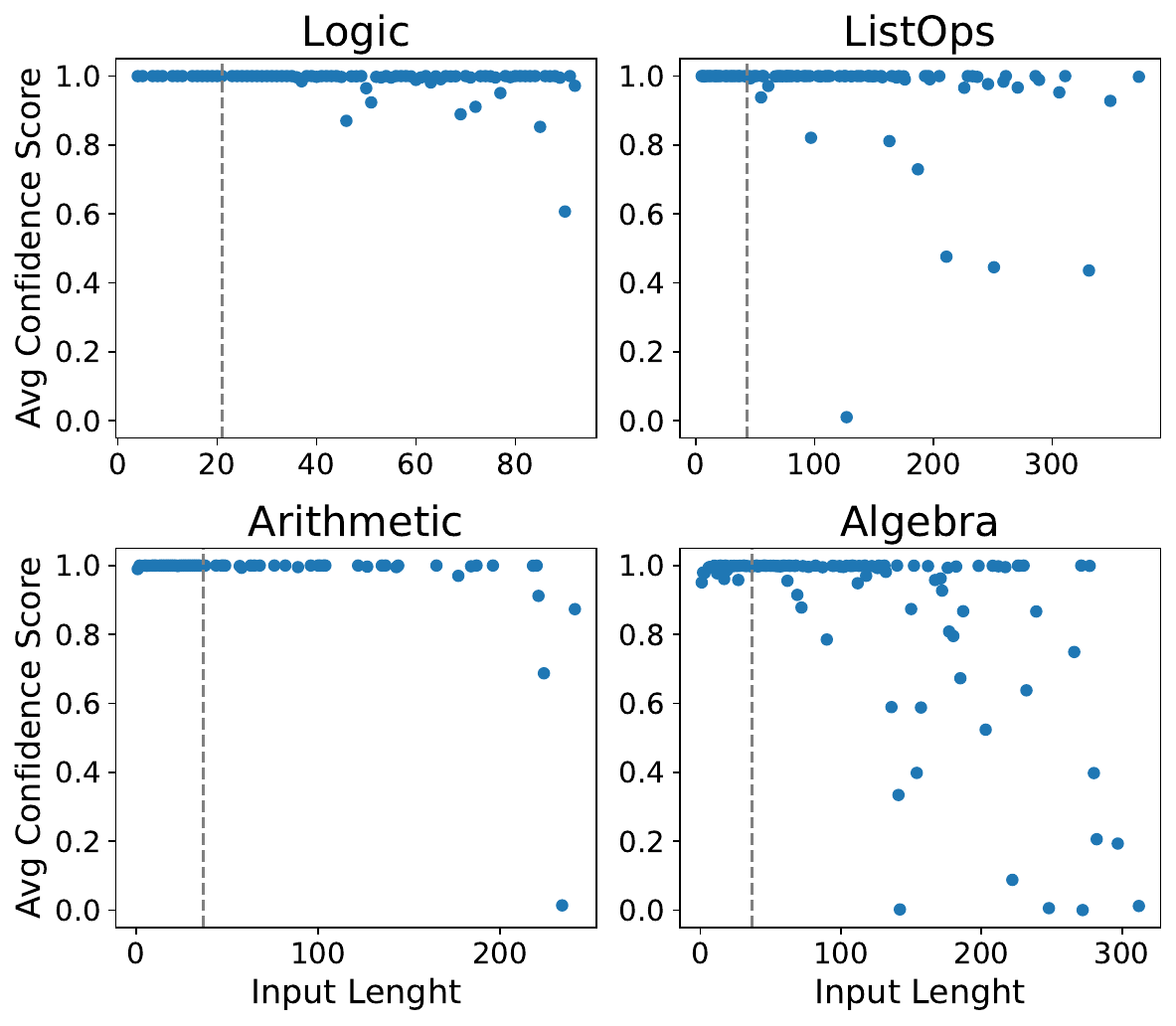}
        \caption{Average multi-domain NRS Selector confidence scores by input length. The vertical line represents the maximum length of training formulas.}
        \label{fig:conf_scores_nrs_multi}
    \end{minipage}
\end{figure}

As described in Section \ref{sec:nrs}, the Dynamic Windowing mechanism in the NRS Selector is regulated by a threshold \textit{T}, which is used to determine on which formulas the mechanism should be applied.
We select these thresholds by examining the average Selector confidence score for inputs of the same length, and choose the value corresponding to a decrease in average Selector confidence. We measure the average Selector confidence on several formulas of different lengths and nesting levels drawn from both the in-distribution and out-of-distribution validation sets. The distribution of these values is represented in Figures \ref{fig:conf_scores_nrs} and \ref{fig:conf_scores_nrs_multi}.

\section{Distribution of FastNRS Solver confidence scores}
\label{sec:appendix-solver-conf}

\begin{figure}[h]
    \centering
    \begin{subfigure}[b]{0.25\textwidth}
     \centering
         \includegraphics[width=\textwidth]{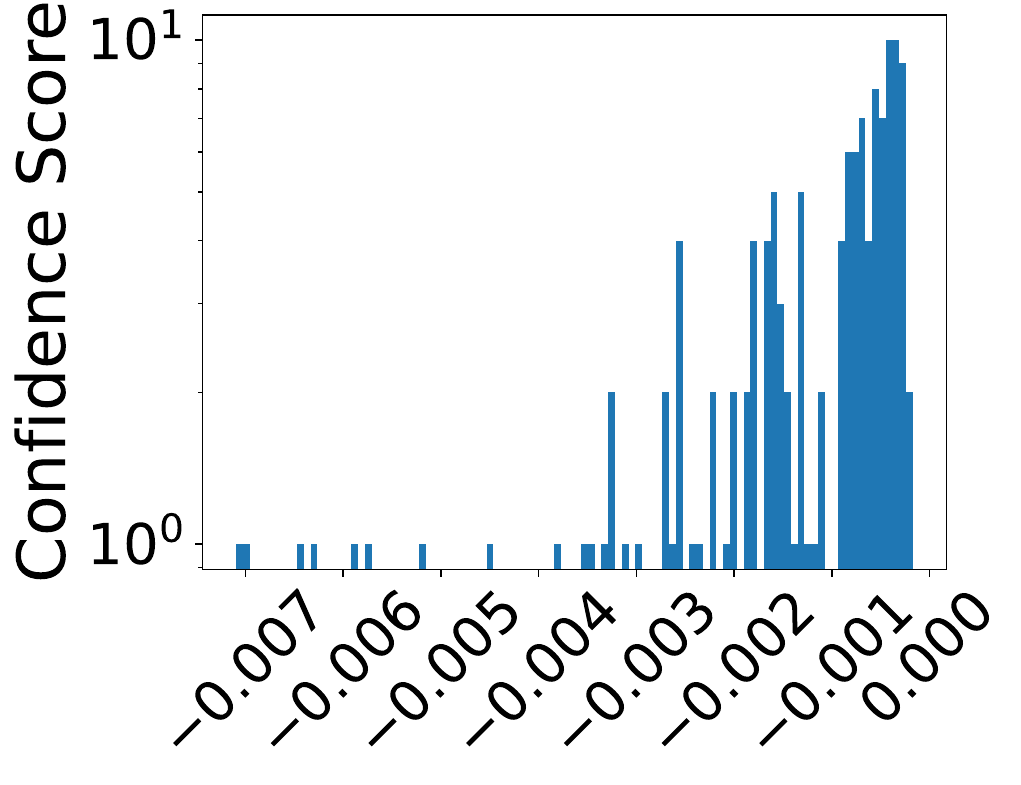}
         \caption{Logic}
         \label{fig:solver-conf-logic}
    \end{subfigure}
    \begin{subfigure}[b]{0.22\textwidth}
     \centering
         \includegraphics[width=\textwidth]{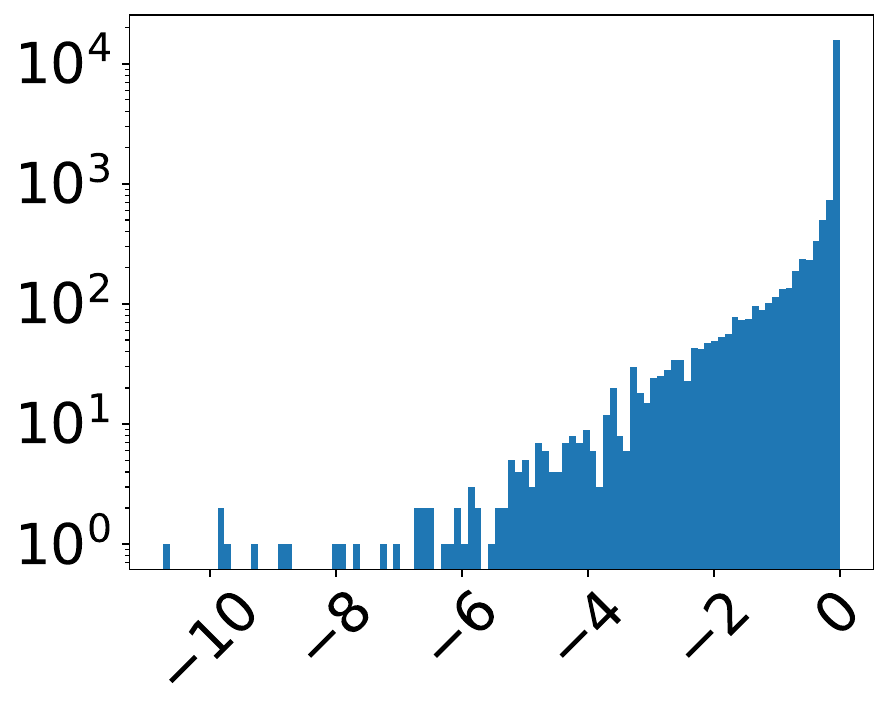}
         \vspace*{-2.6mm}\caption{Listops}
         \label{fig:solver-conf-listops}
    \end{subfigure}
    \begin{subfigure}[b]{0.22\textwidth}
     \centering
         \includegraphics[width=\textwidth]{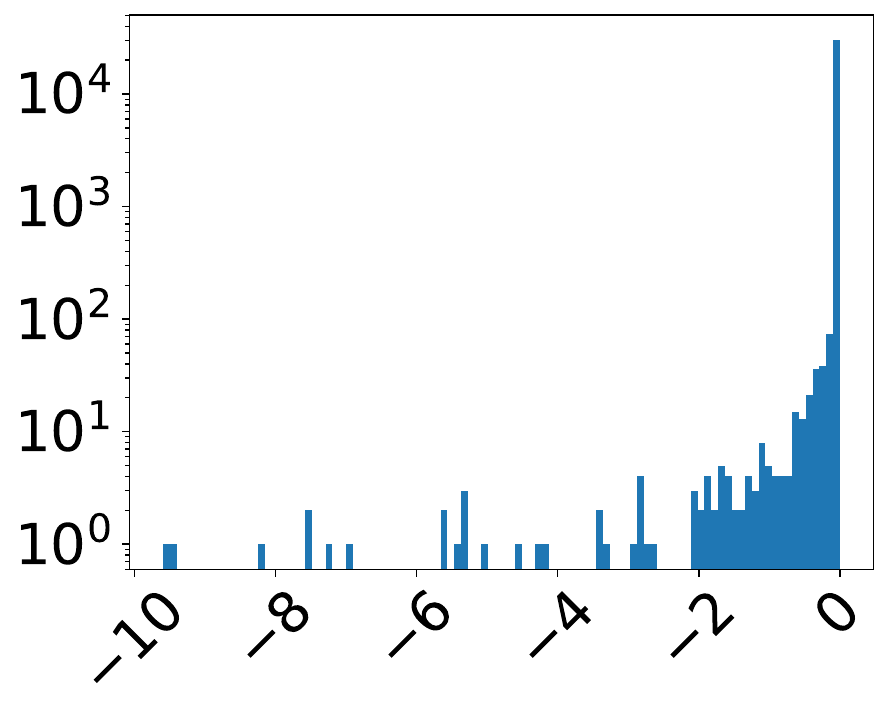}
         \vspace*{-2.6mm}\caption{Arithmetic}
         \label{fig:solver-conf-arithmetic}
    \end{subfigure}
    \begin{subfigure}[b]{0.22\textwidth}
     \centering
         \includegraphics[width=\textwidth]{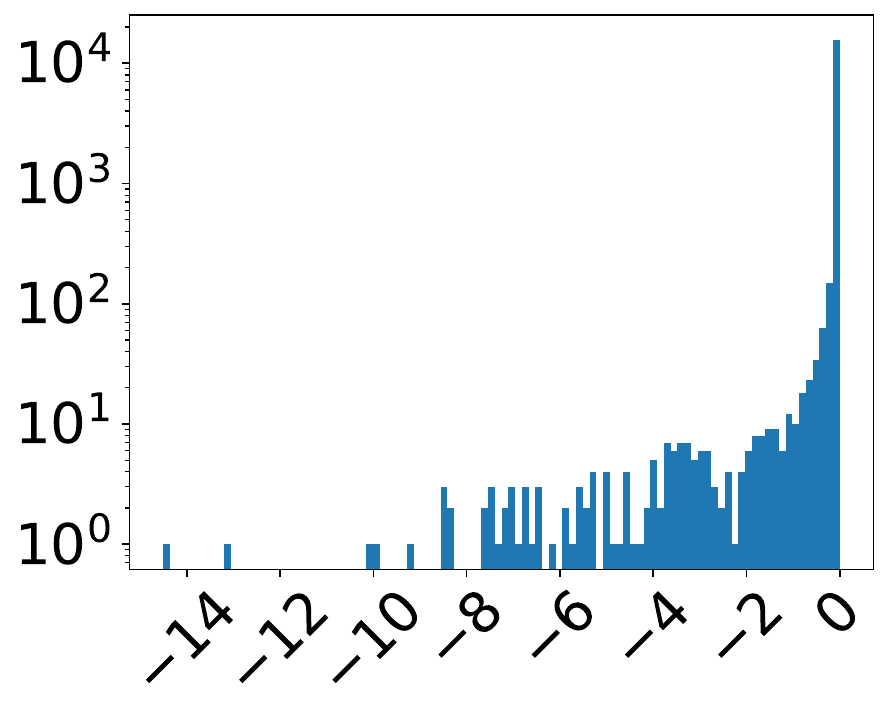}
         \vspace*{-2.6mm}\caption{Algebra}
         \label{fig:solver-conf-algebra}
    \end{subfigure}
    \caption{Distribution of Solver confidence scores on training samples (y-axis in log scale).}
    \label{fig:solver-conf}
\end{figure}

The conditional replacement of leaf formulas in the FastNRS operated by the \texttt{cond\_repl} function is regulated by a threshold on Solver confidence. For both the single- and multi-domain versions of the FastNRS, we determined these thresholds based on the distribution of these scores on training samples, as mentioned in Section \ref{sec:fast-nrs}.
Specifically, the thresholds used were -6 for ListOps, -2 for Arithmetic, -3 for Algebra, and -0.005 for Logic.
The distributions of Solver confidence scores on training samples, which informed these thresholds, are represented in Figure \ref{fig:solver-conf}. The plots can provide insight into how frequently high-confidence predictions occur.

\section{Baseline models statistics}
\label{app:baseline-stats}
In table \ref{tab:baselines_eff} we report the number of parameters and inference time statistics for the three baselines we consider in this study.
For OpenAI o1-preview, statistics were computed only on out-of-distribution data splits, on which the model was tested (namely, formulas with three or more nesting levels).
For the NDR, the runs were executed on a single NVIDIA A100 GPU, as done with our models.

\begin{table}[]
    \centering
    \begin{tabular}{l|l|r|r}
         \textbf{Architecture} & \textbf{Problem} & \textbf{\# Param.} & \textbf{Inf. time} \\
         \hline
         \hline
         o1-preview & Logic & $\sim$ 1.76 trillion & 1h54m\\
         o1-preview & ListOps & $\sim$ 1.76 trillion & 1h47m\\
         o1-preview & Arithmetic & $\sim$ 1.76 trillion & 2h01m\\
         o1-preview & Algebra & $\sim$ 1.76 trillion & 2h04m\\
         \hline
         gpt-4 & Logic & $\sim$ 1.76 trillion & 3h41m\\
         gpt-4 & ListOps & $\sim$ 1.76 trillion & 2h13m\\
         gpt-4 & Arithmetic & $\sim$ 1.76 trillion & 2h00m\\
         gpt-4 & Algebra& $\sim$ 1.76 trillion & 3h06m\\
         \hline
         NDR & Logic & 1,007,667 & 24s\\
         NDR & ListOps & 2,921,521 & 13s\\
         NDR & Arithmetic & 4,055,676 & 25s\\
         NDR & Algebra & 5,197,540 & 21s\\
         
    \end{tabular}
    \captionof{table}{Space and time efficiency statistics for the baseline models.\protect\footnotemark}
    \label{tab:baselines_eff}
\end{table}

\footnotetext{The number of parameters of GPT-4 is an independent estimate based on inference speed \cite{gptparams}. The number of parameters of o1-preview is a ballpark estimate based on the assumption that the model is a fine-tuned version of GPT-4.}

\bibliographystyle{elsarticle-harv} 
\bibliography{bibliography}

\begin{thebibliography}{53}
\expandafter\ifx\csname natexlab\endcsname\relax\def\natexlab#1{#1}\fi
\providecommand{\url}[1]{\texttt{#1}}
\providecommand{\href}[2]{#2}
\providecommand{\path}[1]{#1}
\providecommand{\DOIprefix}{doi:}
\providecommand{\ArXivprefix}{arXiv:}
\providecommand{\URLprefix}{URL: }
\providecommand{\Pubmedprefix}{pmid:}
\providecommand{\doi}[1]{\href{http://dx.doi.org/#1}{\path{#1}}}
\providecommand{\Pubmed}[1]{\href{pmid:#1}{\path{#1}}}
\providecommand{\bibinfo}[2]{#2}
\ifx\xfnm\relax \def\xfnm[#1]{\unskip,\space#1}\fi
\bibitem[{Agarwal et~al.(2021)Agarwal, Aditya and
  Goyal}]{DBLP:journals/corr/abs-2104-14095}
\bibinfo{author}{Agarwal, V.}, \bibinfo{author}{Aditya, S.},
  \bibinfo{author}{Goyal, N.}, \bibinfo{year}{2021}.
\newblock \bibinfo{title}{Analyzing the nuances of transformers' polynomial
  simplification abilities}.
\newblock \bibinfo{journal}{CoRR} \bibinfo{volume}{abs/2104.14095}.
\newblock \URLprefix \url{https://arxiv.org/abs/2104.14095},
  \href{http://arxiv.org/abs/2104.14095}{{\tt arXiv:2104.14095}}.
\bibitem[{Anil et~al.(2022)Anil, Wu, Andreassen, Lewkowycz, Misra, Ramasesh,
  Slone, Gur{-}Ari, Dyer and Neyshabur}]{DBLP:conf/nips/AnilWALMRSGDN22}
\bibinfo{author}{Anil, C.}, \bibinfo{author}{Wu, Y.},
  \bibinfo{author}{Andreassen, A.}, \bibinfo{author}{Lewkowycz, A.},
  \bibinfo{author}{Misra, V.}, \bibinfo{author}{Ramasesh, V.V.},
  \bibinfo{author}{Slone, A.}, \bibinfo{author}{Gur{-}Ari, G.},
  \bibinfo{author}{Dyer, E.}, \bibinfo{author}{Neyshabur, B.},
  \bibinfo{year}{2022}.
\newblock \bibinfo{title}{Exploring length generalization in large language
  models}, in: \bibinfo{editor}{Koyejo, S.}, \bibinfo{editor}{Mohamed, S.},
  \bibinfo{editor}{Agarwal, A.}, \bibinfo{editor}{Belgrave, D.},
  \bibinfo{editor}{Cho, K.}, \bibinfo{editor}{Oh, A.} (Eds.),
  \bibinfo{booktitle}{Advances in Neural Information Processing Systems 35:
  Annual Conference on Neural Information Processing Systems 2022, NeurIPS
  2022, New Orleans, LA, USA, November 28 - December 9, 2022}.
\bibitem[{Baader and Nipkow(1998)}]{DBLP:books/daglib/0092409}
\bibinfo{author}{Baader, F.}, \bibinfo{author}{Nipkow, T.},
  \bibinfo{year}{1998}.
\newblock \bibinfo{title}{Term rewriting and all that}.
\newblock \bibinfo{publisher}{Cambridge University Press}.
\bibitem[{Brown et~al.(2020)Brown, Mann, Ryder, Subbiah, Kaplan, Dhariwal,
  Neelakantan, Shyam, Sastry, Askell, Agarwal, Herbert{-}Voss, Krueger,
  Henighan, Child, Ramesh, Ziegler, Wu, Winter, Hesse, Chen, Sigler, Litwin,
  Gray, Chess, Clark, Berner, McCandlish, Radford, Sutskever and
  Amodei}]{DBLP:conf/nips/BrownMRSKDNSSAA20}
\bibinfo{author}{Brown, T.B.}, \bibinfo{author}{Mann, B.},
  \bibinfo{author}{Ryder, N.}, \bibinfo{author}{Subbiah, M.},
  \bibinfo{author}{Kaplan, J.}, \bibinfo{author}{Dhariwal, P.},
  \bibinfo{author}{Neelakantan, A.}, \bibinfo{author}{Shyam, P.},
  \bibinfo{author}{Sastry, G.}, \bibinfo{author}{Askell, A.},
  \bibinfo{author}{Agarwal, S.}, \bibinfo{author}{Herbert{-}Voss, A.},
  \bibinfo{author}{Krueger, G.}, \bibinfo{author}{Henighan, T.},
  \bibinfo{author}{Child, R.}, \bibinfo{author}{Ramesh, A.},
  \bibinfo{author}{Ziegler, D.M.}, \bibinfo{author}{Wu, J.},
  \bibinfo{author}{Winter, C.}, \bibinfo{author}{Hesse, C.},
  \bibinfo{author}{Chen, M.}, \bibinfo{author}{Sigler, E.},
  \bibinfo{author}{Litwin, M.}, \bibinfo{author}{Gray, S.},
  \bibinfo{author}{Chess, B.}, \bibinfo{author}{Clark, J.},
  \bibinfo{author}{Berner, C.}, \bibinfo{author}{McCandlish, S.},
  \bibinfo{author}{Radford, A.}, \bibinfo{author}{Sutskever, I.},
  \bibinfo{author}{Amodei, D.}, \bibinfo{year}{2020}.
\newblock \bibinfo{title}{Language models are few-shot learners}, in:
  \bibinfo{editor}{Larochelle, H.}, \bibinfo{editor}{Ranzato, M.},
  \bibinfo{editor}{Hadsell, R.}, \bibinfo{editor}{Balcan, M.},
  \bibinfo{editor}{Lin, H.} (Eds.), \bibinfo{booktitle}{Advances in Neural
  Information Processing Systems 33: Annual Conference on Neural Information
  Processing Systems 2020, NeurIPS 2020, December 6-12, 2020, virtual}.
\bibitem[{Cai et~al.(2018)}]{Cai2018-xx}
\bibinfo{author}{Cai, C.H.}, et~al., \bibinfo{year}{2018}.
\newblock \bibinfo{title}{Learning of human-like algebraic reasoning using deep
  feedforward neural networks}.
\newblock \bibinfo{journal}{Biol. Inspired Cogn. Arch.} \bibinfo{volume}{25},
  \bibinfo{pages}{43--50}.
\bibitem[{Caruana(1997)}]{caruana1997multitask}
\bibinfo{author}{Caruana, R.}, \bibinfo{year}{1997}.
\newblock \bibinfo{title}{Multitask learning}.
\newblock \bibinfo{journal}{Machine learning} \bibinfo{volume}{28},
  \bibinfo{pages}{41--75}.
\bibitem[{Chang et~al.(2023)Chang, Zhang, Barber, Maschinot, Lezama, Jiang,
  Yang, Murphy, Freeman, Rubinstein, Li and
  Krishnan}]{chang2023musetexttoimagegenerationmasked}
\bibinfo{author}{Chang, H.}, \bibinfo{author}{Zhang, H.},
  \bibinfo{author}{Barber, J.}, \bibinfo{author}{Maschinot, A.},
  \bibinfo{author}{Lezama, J.}, \bibinfo{author}{Jiang, L.},
  \bibinfo{author}{Yang, M.H.}, \bibinfo{author}{Murphy, K.},
  \bibinfo{author}{Freeman, W.T.}, \bibinfo{author}{Rubinstein, M.},
  \bibinfo{author}{Li, Y.}, \bibinfo{author}{Krishnan, D.},
  \bibinfo{year}{2023}.
\newblock \bibinfo{title}{Muse: Text-to-image generation via masked generative
  transformers}.
\newblock \URLprefix \url{https://arxiv.org/abs/2301.00704},
  \href{http://arxiv.org/abs/2301.00704}{{\tt arXiv:2301.00704}}.
\bibitem[{Chen et~al.(2021)Chen, Tworek, Jun, Yuan, Pinto, Kaplan, Edwards,
  Burda, Joseph, Brockman et~al.}]{chen2021evaluating}
\bibinfo{author}{Chen, M.}, \bibinfo{author}{Tworek, J.}, \bibinfo{author}{Jun,
  H.}, \bibinfo{author}{Yuan, Q.}, \bibinfo{author}{Pinto, H.P.D.O.},
  \bibinfo{author}{Kaplan, J.}, \bibinfo{author}{Edwards, H.},
  \bibinfo{author}{Burda, Y.}, \bibinfo{author}{Joseph, N.},
  \bibinfo{author}{Brockman, G.}, et~al., \bibinfo{year}{2021}.
\newblock \bibinfo{title}{Evaluating large language models trained on code}.
\newblock \bibinfo{journal}{arXiv preprint arXiv:2107.03374} .
\bibitem[{Chen and Tian(2019)}]{DBLP:conf/nips/ChenT19}
\bibinfo{author}{Chen, X.}, \bibinfo{author}{Tian, Y.}, \bibinfo{year}{2019}.
\newblock \bibinfo{title}{Learning to perform local rewriting for combinatorial
  optimization}, in: \bibinfo{editor}{Wallach, H.M.}, et~al. (Eds.),
  \bibinfo{booktitle}{Advances in Neural Information Processing Systems 32:
  Annual Conference on Neural Information Processing Systems 2019, NeurIPS
  2019, December 8-14, 2019, Vancouver, BC, Canada}, pp.
  \bibinfo{pages}{6278--6289}.
\bibitem[{Cognolato and Testolin(2022)}]{cognolato2022transformers}
\bibinfo{author}{Cognolato, S.}, \bibinfo{author}{Testolin, A.},
  \bibinfo{year}{2022}.
\newblock \bibinfo{title}{Transformers discover an elementary calculation
  system exploiting local attention and grid-like problem representation}, in:
  \bibinfo{booktitle}{2022 International Joint Conference on Neural Networks
  (IJCNN)}, \bibinfo{organization}{IEEE}. pp. \bibinfo{pages}{1--8}.
\bibitem[{Cormen and Leiserson(2022)}]{Cormen2022-px}
\bibinfo{author}{Cormen, T.H.}, \bibinfo{author}{Leiserson, C.E.},
  \bibinfo{year}{2022}.
\newblock \bibinfo{title}{Introduction to Algorithms, fourth edition}.
\newblock \bibinfo{publisher}{MIT Press}, \bibinfo{address}{London, England}.
\bibitem[{Csord{\'{a}}s et~al.(2021)Csord{\'{a}}s, Irie and
  Schmidhuber}]{DBLP:conf/emnlp/CsordasIS21}
\bibinfo{author}{Csord{\'{a}}s, R.}, \bibinfo{author}{Irie, K.},
  \bibinfo{author}{Schmidhuber, J.}, \bibinfo{year}{2021}.
\newblock \bibinfo{title}{The devil is in the detail: Simple tricks improve
  systematic generalization of transformers}, in: \bibinfo{editor}{Moens, M.},
  \bibinfo{editor}{Huang, X.}, \bibinfo{editor}{Specia, L.},
  \bibinfo{editor}{Yih, S.W.} (Eds.), \bibinfo{booktitle}{Proceedings of the
  2021 Conference on Empirical Methods in Natural Language Processing, {EMNLP}
  2021, Virtual Event / Punta Cana, Dominican Republic, 7-11 November, 2021},
  \bibinfo{publisher}{Association for Computational Linguistics}. pp.
  \bibinfo{pages}{619--634}.
\newblock \DOIprefix\doi{10.18653/V1/2021.EMNLP-MAIN.49}.
\bibitem[{Csord{\'{a}}s et~al.(2022)Csord{\'{a}}s, Irie and
  Schmidhuber}]{DBLP:conf/iclr/CsordasIS22}
\bibinfo{author}{Csord{\'{a}}s, R.}, \bibinfo{author}{Irie, K.},
  \bibinfo{author}{Schmidhuber, J.}, \bibinfo{year}{2022}.
\newblock \bibinfo{title}{The neural data router: Adaptive control flow in
  transformers improves systematic generalization}, in: \bibinfo{booktitle}{The
  Tenth International Conference on Learning Representations, {ICLR} 2022,
  Virtual Event, April 25-29, 2022}, \bibinfo{publisher}{OpenReview.net}.
\newblock \URLprefix \url{https://openreview.net/forum?id=KBQP4A\_J1K}.
\bibitem[{Davis(2024)}]{davis2024mathematics}
\bibinfo{author}{Davis, E.}, \bibinfo{year}{2024}.
\newblock \bibinfo{title}{Mathematics, word problems, common sense, and
  artificial intelligence}.
\newblock \bibinfo{journal}{Bulletin of the American Mathematical Society}
  \bibinfo{volume}{61}, \bibinfo{pages}{287--303}.
\bibitem[{Graves et~al.(2014)Graves, Wayne and
  Danihelka}]{DBLP:journals/corr/GravesWD14}
\bibinfo{author}{Graves, A.}, \bibinfo{author}{Wayne, G.},
  \bibinfo{author}{Danihelka, I.}, \bibinfo{year}{2014}.
\newblock \bibinfo{title}{Neural turing machines}.
\newblock \bibinfo{journal}{CoRR} \bibinfo{volume}{abs/1410.5401}.
\newblock \URLprefix \url{http://arxiv.org/abs/1410.5401},
  \href{http://arxiv.org/abs/1410.5401}{{\tt arXiv:1410.5401}}.
\bibitem[{Graves et~al.(2016)Graves, Wayne, Reynolds, Harley, Danihelka,
  Grabska{-}Barwinska, Colmenarejo, Grefenstette, Ramalho, Agapiou, Badia,
  Hermann, Zwols, Ostrovski, Cain, King, Summerfield, Blunsom, Kavukcuoglu and
  Hassabis}]{DBLP:journals/nature/GravesWRHDGCGRA16}
\bibinfo{author}{Graves, A.}, \bibinfo{author}{Wayne, G.},
  \bibinfo{author}{Reynolds, M.}, \bibinfo{author}{Harley, T.},
  \bibinfo{author}{Danihelka, I.}, \bibinfo{author}{Grabska{-}Barwinska, A.},
  \bibinfo{author}{Colmenarejo, S.G.}, \bibinfo{author}{Grefenstette, E.},
  \bibinfo{author}{Ramalho, T.}, \bibinfo{author}{Agapiou, J.P.},
  \bibinfo{author}{Badia, A.P.}, \bibinfo{author}{Hermann, K.M.},
  \bibinfo{author}{Zwols, Y.}, \bibinfo{author}{Ostrovski, G.},
  \bibinfo{author}{Cain, A.}, \bibinfo{author}{King, H.},
  \bibinfo{author}{Summerfield, C.}, \bibinfo{author}{Blunsom, P.},
  \bibinfo{author}{Kavukcuoglu, K.}, \bibinfo{author}{Hassabis, D.},
  \bibinfo{year}{2016}.
\newblock \bibinfo{title}{Hybrid computing using a neural network with dynamic
  external memory}.
\newblock \bibinfo{journal}{Nat.} \bibinfo{volume}{538},
  \bibinfo{pages}{471--476}.
\newblock \URLprefix \url{https://doi.org/10.1038/nature20101},
  \DOIprefix\doi{10.1038/NATURE20101}.
\bibitem[{He et~al.(2016)He, Zhang, Ren and Sun}]{7780459}
\bibinfo{author}{He, K.}, \bibinfo{author}{Zhang, X.}, \bibinfo{author}{Ren,
  S.}, \bibinfo{author}{Sun, J.}, \bibinfo{year}{2016}.
\newblock \bibinfo{title}{Deep residual learning for image recognition}, in:
  \bibinfo{booktitle}{2016 IEEE Conference on Computer Vision and Pattern
  Recognition (CVPR)}, pp. \bibinfo{pages}{770--778}.
\newblock \DOIprefix\doi{10.1109/CVPR.2016.90}.
\bibitem[{Hendrycks et~al.(2021)Hendrycks, Basart, Mu, Kadavath, Wang, Dorundo,
  Desai, Zhu, Parajuli, Guo, Song, Steinhardt and
  Gilmer}]{DBLP:conf/iccv/HendrycksBMKWDD21}
\bibinfo{author}{Hendrycks, D.}, \bibinfo{author}{Basart, S.},
  \bibinfo{author}{Mu, N.}, \bibinfo{author}{Kadavath, S.},
  \bibinfo{author}{Wang, F.}, \bibinfo{author}{Dorundo, E.},
  \bibinfo{author}{Desai, R.}, \bibinfo{author}{Zhu, T.},
  \bibinfo{author}{Parajuli, S.}, \bibinfo{author}{Guo, M.},
  \bibinfo{author}{Song, D.}, \bibinfo{author}{Steinhardt, J.},
  \bibinfo{author}{Gilmer, J.}, \bibinfo{year}{2021}.
\newblock \bibinfo{title}{The many faces of robustness: {A} critical analysis
  of out-of-distribution generalization}, in: \bibinfo{booktitle}{2021
  {IEEE/CVF} International Conference on Computer Vision, {ICCV} 2021,
  Montreal, QC, Canada, October 10-17, 2021}, \bibinfo{publisher}{{IEEE}}. pp.
  \bibinfo{pages}{8320--8329}.
\newblock \URLprefix \url{https://doi.org/10.1109/ICCV48922.2021.00823},
  \DOIprefix\doi{10.1109/ICCV48922.2021.00823}.
\bibitem[{Hinton(1990)}]{DBLP:journals/ai/Hinton90}
\bibinfo{author}{Hinton, G.E.}, \bibinfo{year}{1990}.
\newblock \bibinfo{title}{Connectionist symbol processing - preface}.
\newblock \bibinfo{journal}{Artif. Intell.} \bibinfo{volume}{46},
  \bibinfo{pages}{1--4}.
\newblock \URLprefix \url{https://doi.org/10.1016/0004-3702(90)90002-H},
  \DOIprefix\doi{10.1016/0004-3702(90)90002-H}.
\bibitem[{Hupkes et~al.(2020)Hupkes, Dankers, Mul and
  Bruni}]{DBLP:journals/jair/HupkesDMB20}
\bibinfo{author}{Hupkes, D.}, \bibinfo{author}{Dankers, V.},
  \bibinfo{author}{Mul, M.}, \bibinfo{author}{Bruni, E.}, \bibinfo{year}{2020}.
\newblock \bibinfo{title}{Compositionality decomposed: How do neural networks
  generalise?}
\newblock \bibinfo{journal}{J. Artif. Intell. Res.} \bibinfo{volume}{67},
  \bibinfo{pages}{757--795}.
\newblock \URLprefix \url{https://doi.org/10.1613/jair.1.11674},
  \DOIprefix\doi{10.1613/JAIR.1.11674}.
\bibitem[{Kahneman(2011)}]{Kahneman2011-jy}
\bibinfo{author}{Kahneman, D.}, \bibinfo{year}{2011}.
\newblock \bibinfo{title}{Thinking, Fast and Slow}.
\newblock \bibinfo{publisher}{Farrar, Straus and Giroux}.
\bibitem[{Kautz(2022)}]{Kautz2022-kx}
\bibinfo{author}{Kautz, H.A.}, \bibinfo{year}{2022}.
\newblock \bibinfo{title}{The third {AI} summer: {AAAI} robert s. engelmore
  memorial lecture}.
\newblock \bibinfo{journal}{AI Mag.} \bibinfo{volume}{43},
  \bibinfo{pages}{105--125}.
\bibitem[{Kazemnejad et~al.(2023)Kazemnejad, Padhi, Ramamurthy, Das and
  Reddy}]{DBLP:conf/nips/KazemnejadPRDR23}
\bibinfo{author}{Kazemnejad, A.}, \bibinfo{author}{Padhi, I.},
  \bibinfo{author}{Ramamurthy, K.N.}, \bibinfo{author}{Das, P.},
  \bibinfo{author}{Reddy, S.}, \bibinfo{year}{2023}.
\newblock \bibinfo{title}{The impact of positional encoding on length
  generalization in transformers}, in: \bibinfo{editor}{Oh, A.},
  \bibinfo{editor}{Naumann, T.}, \bibinfo{editor}{Globerson, A.},
  \bibinfo{editor}{Saenko, K.}, \bibinfo{editor}{Hardt, M.},
  \bibinfo{editor}{Levine, S.} (Eds.), \bibinfo{booktitle}{Advances in Neural
  Information Processing Systems 36: Annual Conference on Neural Information
  Processing Systems 2023, NeurIPS 2023, New Orleans, LA, USA, December 10 -
  16, 2023}.
\bibitem[{Kojima et~al.(2022)Kojima, Gu, Reid, Matsuo and
  Iwasawa}]{DBLP:conf/nips/KojimaGRMI22}
\bibinfo{author}{Kojima, T.}, \bibinfo{author}{Gu, S.S.},
  \bibinfo{author}{Reid, M.}, \bibinfo{author}{Matsuo, Y.},
  \bibinfo{author}{Iwasawa, Y.}, \bibinfo{year}{2022}.
\newblock \bibinfo{title}{Large language models are zero-shot reasoners}, in:
  \bibinfo{editor}{Koyejo, S.}, \bibinfo{editor}{Mohamed, S.},
  \bibinfo{editor}{Agarwal, A.}, \bibinfo{editor}{Belgrave, D.},
  \bibinfo{editor}{Cho, K.}, \bibinfo{editor}{Oh, A.} (Eds.),
  \bibinfo{booktitle}{Advances in Neural Information Processing Systems 35:
  Annual Conference on Neural Information Processing Systems 2022, NeurIPS
  2022, New Orleans, LA, USA, November 28 - December 9, 2022}.
\bibitem[{Komendantskaya.(2009)}]{icnc09}
\bibinfo{author}{Komendantskaya., E.}, \bibinfo{year}{2009}.
\newblock \bibinfo{title}{Parallel rewriting in neural networks}, in:
  \bibinfo{booktitle}{Proceedings of the International Joint Conference on
  Computational Intelligence (IJCCI 2009) - ICNC},
  \bibinfo{organization}{INSTICC}. \bibinfo{publisher}{SciTePress}. pp.
  \bibinfo{pages}{452--458}.
\newblock \DOIprefix\doi{10.5220/0002319704520458}.
\bibitem[{Lake and Baroni(2018)}]{DBLP:conf/icml/LakeB18}
\bibinfo{author}{Lake, B.M.}, \bibinfo{author}{Baroni, M.},
  \bibinfo{year}{2018}.
\newblock \bibinfo{title}{Generalization without systematicity: On the
  compositional skills of sequence-to-sequence recurrent networks}, in:
  \bibinfo{editor}{Dy, J.G.}, \bibinfo{editor}{Krause, A.} (Eds.),
  \bibinfo{booktitle}{Proceedings of the 35th International Conference on
  Machine Learning, {ICML} 2018, Stockholmsm{\"{a}}ssan, Stockholm, Sweden,
  July 10-15, 2018}, \bibinfo{publisher}{{PMLR}}. pp.
  \bibinfo{pages}{2879--2888}.
\newblock \URLprefix \url{http://proceedings.mlr.press/v80/lake18a.html}.
\bibitem[{Lample and Charton(2019)}]{Lample2019DeepLF}
\bibinfo{author}{Lample, G.}, \bibinfo{author}{Charton, F.},
  \bibinfo{year}{2019}.
\newblock \bibinfo{title}{Deep learning for symbolic mathematics}.
\newblock \bibinfo{journal}{ArXiv} \bibinfo{volume}{abs/1912.01412}.
\bibitem[{Li and McClelland(2022)}]{DBLP:journals/corr/abs-2210-00400}
\bibinfo{author}{Li, Y.}, \bibinfo{author}{McClelland, J.L.},
  \bibinfo{year}{2022}.
\newblock \bibinfo{title}{Systematic generalization and emergent structures in
  transformers trained on structured tasks}, in: \bibinfo{booktitle}{All Things
  Attention: Bridging Different Perspectives on Attention},
  \bibinfo{organization}{Annual Conference on Neural Information Processing
  Systems}.
\bibitem[{Marghetis et~al.(2016)Marghetis, Landy and
  Goldstone}]{marghetis2016mastering}
\bibinfo{author}{Marghetis, T.}, \bibinfo{author}{Landy, D.},
  \bibinfo{author}{Goldstone, R.L.}, \bibinfo{year}{2016}.
\newblock \bibinfo{title}{Mastering algebra retrains the visual system to
  perceive hierarchical structure in equations}.
\newblock \bibinfo{journal}{Cognitive research: principles and implications}
  \bibinfo{volume}{1}, \bibinfo{pages}{1--10}.
\bibitem[{Mirzadeh et~al.(2024)Mirzadeh, Alizadeh, Shahrokhi, Tuzel, Bengio and
  Farajtabar}]{DBLP:journals/corr/abs-2410-05229}
\bibinfo{author}{Mirzadeh, S.}, \bibinfo{author}{Alizadeh, K.},
  \bibinfo{author}{Shahrokhi, H.}, \bibinfo{author}{Tuzel, O.},
  \bibinfo{author}{Bengio, S.}, \bibinfo{author}{Farajtabar, M.},
  \bibinfo{year}{2024}.
\newblock \bibinfo{title}{Gsm-symbolic: Understanding the limitations of
  mathematical reasoning in large language models}.
\newblock \bibinfo{journal}{CoRR} \bibinfo{volume}{abs/2410.05229}.
\newblock \URLprefix \url{https://doi.org/10.48550/arXiv.2410.05229},
  \DOIprefix\doi{10.48550/ARXIV.2410.05229},
  \href{http://arxiv.org/abs/2410.05229}{{\tt arXiv:2410.05229}}.
\bibitem[{Nangia and Bowman(2018)}]{DBLP:conf/naacl/NangiaB18}
\bibinfo{author}{Nangia, N.}, \bibinfo{author}{Bowman, S.R.},
  \bibinfo{year}{2018}.
\newblock \bibinfo{title}{Listops: {A} diagnostic dataset for latent tree
  learning}, in: \bibinfo{editor}{Cordeiro, S.R.}, \bibinfo{editor}{Oraby, S.},
  \bibinfo{editor}{Pavalanathan, U.}, \bibinfo{editor}{Rim, K.} (Eds.),
  \bibinfo{booktitle}{Proceedings of the 2018 Conference of the North American
  Chapter of the Association for Computational Linguistics, {NAACL-HLT} 2018,
  New Orleans, Louisiana, USA, June 2-4, 2018, Student Research Workshop},
  \bibinfo{publisher}{Association for Computational Linguistics}. pp.
  \bibinfo{pages}{92--99}.
\newblock \DOIprefix\doi{10.18653/V1/N18-4013}.
\bibitem[{Newell and Simon(1956)}]{newell1956logic}
\bibinfo{author}{Newell, A.}, \bibinfo{author}{Simon, H.},
  \bibinfo{year}{1956}.
\newblock \bibinfo{title}{The logic theory machine--a complex information
  processing system}.
\newblock \bibinfo{journal}{IRE Transactions on information theory}
  \bibinfo{volume}{2}, \bibinfo{pages}{61--79}.
\bibitem[{OpenAI(2023)}]{openai2023gpt4}
\bibinfo{author}{OpenAI}, \bibinfo{year}{2023}.
\newblock \bibinfo{title}{{GPT-4} technical report}.
\newblock \href{http://arxiv.org/abs/2303.08774}{{\tt arXiv:2303.08774}}.
\bibitem[{OpenAI(2024)}]{o1report}
\bibinfo{author}{OpenAI}, \bibinfo{year}{2024}.
\newblock \bibinfo{title}{Learning to reason with {LLMs}}.
\newblock
  \bibinfo{howpublished}{\url{https://openai.com/index/learning-to-reason-with-llms/}}.
\newblock \bibinfo{note}{Accessed: 2024-09-29}.
\bibitem[{Petruzzellis et~al.(2024a)Petruzzellis, Testolin and
  Sperduti}]{petruzzellisAssessingICANN24}
\bibinfo{author}{Petruzzellis, F.}, \bibinfo{author}{Testolin, A.},
  \bibinfo{author}{Sperduti, A.}, \bibinfo{year}{2024}a.
\newblock \bibinfo{title}{Assessing the emergent symbolic reasoning abilities
  of llama large language models.}
\newblock \bibinfo{note}{To appear in Proceedings of the 33rd International
  Conference on Artificial Neural Networks (ICANN24)}.
\bibitem[{Petruzzellis et~al.(2024b)Petruzzellis, Testolin and
  Sperduti}]{lrec-coling24}
\bibinfo{author}{Petruzzellis, F.}, \bibinfo{author}{Testolin, A.},
  \bibinfo{author}{Sperduti, A.}, \bibinfo{year}{2024}b.
\newblock \bibinfo{title}{Benchmarking {GPT-4} on algorithmic problems: {A}
  systematic evaluation of prompting strategies}.
\newblock \bibinfo{journal}{Procedings of the 2024 Joint International
  Conference on Computational Linguistics, Language Resources and Evaluation,
  {LREC-COLING 2024}, Turin (Italy), May, 20-25, 2024} .
\bibitem[{Petruzzellis et~al.(2024c)Petruzzellis, Testolin and Sperduti}]{nrs}
\bibinfo{author}{Petruzzellis, F.}, \bibinfo{author}{Testolin, A.},
  \bibinfo{author}{Sperduti, A.}, \bibinfo{year}{2024}c.
\newblock \bibinfo{title}{A {Neural Rewriting System} to {Solve Algorithmic
  Problems}}.
\newblock \bibinfo{note}{To appear in Proceedings of the 27th European
  Conference on Artificial Intelligence}.
\bibitem[{Pinker and Prince(1988)}]{Pinker1988-vw}
\bibinfo{author}{Pinker, S.}, \bibinfo{author}{Prince, A.},
  \bibinfo{year}{1988}.
\newblock \bibinfo{title}{On language and connectionism: analysis of a parallel
  distributed processing model of language acquisition}.
\newblock \bibinfo{journal}{Cognition} \bibinfo{volume}{28},
  \bibinfo{pages}{73--193}.
\bibitem[{Ruiz et~al.(2021)Ruiz, Ainslie and
  Onta{\~{n}}{\'{o}}n}]{DBLP:journals/corr/abs-2110-04169}
\bibinfo{author}{Ruiz, L.}, \bibinfo{author}{Ainslie, J.},
  \bibinfo{author}{Onta{\~{n}}{\'{o}}n, S.}, \bibinfo{year}{2021}.
\newblock \bibinfo{title}{Iterative decoding for compositional generalization
  in transformers}.
\newblock \bibinfo{journal}{CoRR} \bibinfo{volume}{abs/2110.04169}.
\newblock \URLprefix \url{https://arxiv.org/abs/2110.04169},
  \href{http://arxiv.org/abs/2110.04169}{{\tt arXiv:2110.04169}}.
\bibitem[{Rumelhart et~al.(1986)Rumelhart, McClelland and
  {AU}}]{Rumelhart1986-uj}
\bibinfo{author}{Rumelhart, D.E.}, \bibinfo{author}{McClelland, J.L.},
  \bibinfo{author}{{AU}}, \bibinfo{year}{1986}.
\newblock \bibinfo{title}{Parallel distributed processing}.
\newblock \bibinfo{publisher}{The MIT Press}.
\bibitem[{Ruoss et~al.(2023)Ruoss, Del{\'{e}}tang, Genewein, Grau{-}Moya,
  Csord{\'{a}}s, Bennani, Legg and Veness}]{DBLP:conf/acl/RuossDGGCBLV23}
\bibinfo{author}{Ruoss, A.}, \bibinfo{author}{Del{\'{e}}tang, G.},
  \bibinfo{author}{Genewein, T.}, \bibinfo{author}{Grau{-}Moya, J.},
  \bibinfo{author}{Csord{\'{a}}s, R.}, \bibinfo{author}{Bennani, M.},
  \bibinfo{author}{Legg, S.}, \bibinfo{author}{Veness, J.},
  \bibinfo{year}{2023}.
\newblock \bibinfo{title}{Randomized positional encodings boost length
  generalization of transformers}, in: \bibinfo{editor}{Rogers, A.},
  \bibinfo{editor}{Boyd{-}Graber, J.L.}, \bibinfo{editor}{Okazaki, N.} (Eds.),
  \bibinfo{booktitle}{Proceedings of the 61st Annual Meeting of the Association
  for Computational Linguistics (Volume 2: Short Papers), {ACL} 2023, Toronto,
  Canada, July 9-14, 2023}, \bibinfo{publisher}{Association for Computational
  Linguistics}. pp. \bibinfo{pages}{1889--1903}.
\newblock \URLprefix \url{https://doi.org/10.18653/v1/2023.acl-short.161},
  \DOIprefix\doi{10.18653/V1/2023.ACL-SHORT.161}.
\bibitem[{Saxton et~al.(2019)Saxton, Grefenstette, Hill and
  Kohli}]{DBLP:conf/iclr/SaxtonGHK19}
\bibinfo{author}{Saxton, D.}, \bibinfo{author}{Grefenstette, E.},
  \bibinfo{author}{Hill, F.}, \bibinfo{author}{Kohli, P.},
  \bibinfo{year}{2019}.
\newblock \bibinfo{title}{Analysing mathematical reasoning abilities of neural
  models}, in: \bibinfo{booktitle}{7th International Conference on Learning
  Representations, {ICLR} 2019, New Orleans, LA, USA, May 6-9, 2019},
  \bibinfo{publisher}{OpenReview.net}.
\newblock \URLprefix \url{https://openreview.net/forum?id=H1gR5iR5FX}.
\bibitem[{Schreiner(2023)}]{gptparams}
\bibinfo{author}{Schreiner, M.}, \bibinfo{year}{2023}.
\newblock \bibinfo{title}{{GPT-4} architecture, datasets, costs and more
  leaked}.
\newblock
  \bibinfo{howpublished}{\url{https://web.archive.org/web/20230712123915/https://the-decoder.com/gpt-4-architecture-datasets-costs-and-more-leaked/}}.
\newblock \bibinfo{note}{Accessed: 2023-07-12}.
\bibitem[{Setzler et~al.(2022)Setzler, Howland and
  Phillips}]{DBLP:journals/corr/abs-2201-11766}
\bibinfo{author}{Setzler, M.}, \bibinfo{author}{Howland, S.},
  \bibinfo{author}{Phillips, L.A.}, \bibinfo{year}{2022}.
\newblock \bibinfo{title}{Recursive decoding: {A} situated cognition approach
  to compositional generation in grounded language understanding}.
\newblock \bibinfo{journal}{CoRR} \bibinfo{volume}{abs/2201.11766}.
\newblock \URLprefix \url{https://arxiv.org/abs/2201.11766},
  \href{http://arxiv.org/abs/2201.11766}{{\tt arXiv:2201.11766}}.
\bibitem[{Testolin(2024)}]{testolin2023can}
\bibinfo{author}{Testolin, A.}, \bibinfo{year}{2024}.
\newblock \bibinfo{title}{Can neural networks do arithmetic? a survey on the
  elementary numerical skills of state-of-the-art deep learning models}.
\newblock \bibinfo{journal}{Applied Sciences} .
\bibitem[{Vaswani et~al.(2017)Vaswani, Shazeer, Parmar, Uszkoreit, Jones,
  Gomez, Kaiser and Polosukhin}]{DBLP:conf/nips/VaswaniSPUJGKP17}
\bibinfo{author}{Vaswani, A.}, \bibinfo{author}{Shazeer, N.},
  \bibinfo{author}{Parmar, N.}, \bibinfo{author}{Uszkoreit, J.},
  \bibinfo{author}{Jones, L.}, \bibinfo{author}{Gomez, A.N.},
  \bibinfo{author}{Kaiser, L.}, \bibinfo{author}{Polosukhin, I.},
  \bibinfo{year}{2017}.
\newblock \bibinfo{title}{Attention is all you need}, in:
  \bibinfo{editor}{Guyon, I.}, \bibinfo{editor}{von Luxburg, U.},
  \bibinfo{editor}{Bengio, S.}, \bibinfo{editor}{Wallach, H.M.},
  \bibinfo{editor}{Fergus, R.}, \bibinfo{editor}{Vishwanathan, S.V.N.},
  \bibinfo{editor}{Garnett, R.} (Eds.), \bibinfo{booktitle}{Advances in Neural
  Information Processing Systems 30: Annual Conference on Neural Information
  Processing Systems 2017, December 4-9, 2017, Long Beach, CA, {USA}}, pp.
  \bibinfo{pages}{5998--6008}.
\bibitem[{Velickovic and Blundell(2021)}]{DBLP:journals/patterns/VelickovicB21}
\bibinfo{author}{Velickovic, P.}, \bibinfo{author}{Blundell, C.},
  \bibinfo{year}{2021}.
\newblock \bibinfo{title}{Neural algorithmic reasoning}.
\newblock \bibinfo{journal}{Patterns} \bibinfo{volume}{2},
  \bibinfo{pages}{100273}.
\newblock \URLprefix \url{https://doi.org/10.1016/j.patter.2021.100273},
  \DOIprefix\doi{10.1016/J.PATTER.2021.100273}.
\bibitem[{Velickovic et~al.(2020)Velickovic, Ying, Padovano, Hadsell and
  Blundell}]{DBLP:conf/iclr/VelickovicYPHB20}
\bibinfo{author}{Velickovic, P.}, \bibinfo{author}{Ying, R.},
  \bibinfo{author}{Padovano, M.}, \bibinfo{author}{Hadsell, R.},
  \bibinfo{author}{Blundell, C.}, \bibinfo{year}{2020}.
\newblock \bibinfo{title}{Neural execution of graph algorithms}, in:
  \bibinfo{booktitle}{8th International Conference on Learning Representations,
  {ICLR} 2020, Addis Ababa, Ethiopia, April 26-30, 2020},
  \bibinfo{publisher}{OpenReview.net}.
\newblock \URLprefix \url{https://openreview.net/forum?id=SkgKO0EtvS}.
\bibitem[{Vinyals et~al.(2015)Vinyals, Fortunato and
  Jaitly}]{DBLP:conf/nips/VinyalsFJ15}
\bibinfo{author}{Vinyals, O.}, \bibinfo{author}{Fortunato, M.},
  \bibinfo{author}{Jaitly, N.}, \bibinfo{year}{2015}.
\newblock \bibinfo{title}{Pointer networks}, in: \bibinfo{editor}{Cortes, C.},
  \bibinfo{editor}{Lawrence, N.D.}, \bibinfo{editor}{Lee, D.D.},
  \bibinfo{editor}{Sugiyama, M.}, \bibinfo{editor}{Garnett, R.} (Eds.),
  \bibinfo{booktitle}{Advances in Neural Information Processing Systems 28:
  Annual Conference on Neural Information Processing Systems 2015, December
  7-12, 2015, Montreal, Quebec, Canada}, pp. \bibinfo{pages}{2692--2700}.
\bibitem[{Wang et~al.(2023)}]{DBLP:conf/iclr/0002WSLCNCZ23}
\bibinfo{author}{Wang, X.}, et~al., \bibinfo{year}{2023}.
\newblock \bibinfo{title}{Self-consistency improves chain of thought reasoning
  in language models}, in: \bibinfo{booktitle}{The Eleventh International
  Conference on Learning Representations, {ICLR} 2023, Kigali, Rwanda, May 1-5,
  2023}.
\bibitem[{Wei et~al.(2022)Wei, Wang, Schuurmans, Bosma, Ichter, Xia, Chi, Le
  and Zhou}]{DBLP:conf/nips/Wei0SBIXCLZ22}
\bibinfo{author}{Wei, J.}, \bibinfo{author}{Wang, X.},
  \bibinfo{author}{Schuurmans, D.}, \bibinfo{author}{Bosma, M.},
  \bibinfo{author}{Ichter, B.}, \bibinfo{author}{Xia, F.},
  \bibinfo{author}{Chi, E.H.}, \bibinfo{author}{Le, Q.V.},
  \bibinfo{author}{Zhou, D.}, \bibinfo{year}{2022}.
\newblock \bibinfo{title}{Chain-of-thought prompting elicits reasoning in large
  language models}, in: \bibinfo{editor}{Koyejo, S.}, \bibinfo{editor}{Mohamed,
  S.}, \bibinfo{editor}{Agarwal, A.}, \bibinfo{editor}{Belgrave, D.},
  \bibinfo{editor}{Cho, K.}, \bibinfo{editor}{Oh, A.} (Eds.),
  \bibinfo{booktitle}{Advances in Neural Information Processing Systems 35:
  Annual Conference on Neural Information Processing Systems 2022, NeurIPS
  2022, New Orleans, LA, USA, November 28 - December 9, 2022}.
\bibitem[{Ye et~al.(2021)Ye, Xie, Cai, Li, Li and
  Wang}]{DBLP:conf/nips/YeXCLLW21}
\bibinfo{author}{Ye, H.}, \bibinfo{author}{Xie, C.}, \bibinfo{author}{Cai, T.},
  \bibinfo{author}{Li, R.}, \bibinfo{author}{Li, Z.}, \bibinfo{author}{Wang,
  L.}, \bibinfo{year}{2021}.
\newblock \bibinfo{title}{Towards a theoretical framework of
  out-of-distribution generalization}, in: \bibinfo{editor}{Ranzato, M.},
  \bibinfo{editor}{Beygelzimer, A.}, \bibinfo{editor}{Dauphin, Y.N.},
  \bibinfo{editor}{Liang, P.}, \bibinfo{editor}{Vaughan, J.W.} (Eds.),
  \bibinfo{booktitle}{Advances in Neural Information Processing Systems 34:
  Annual Conference on Neural Information Processing Systems 2021, NeurIPS
  2021, December 6-14, 2021, virtual}, pp. \bibinfo{pages}{23519--23531}.
\bibitem[{Zhou et~al.(2024)Zhou, Bradley, Littwin, Razin, Saremi, Susskind,
  Bengio and Nakkiran}]{DBLP:conf/iclr/ZhouBLRSSBN24}
\bibinfo{author}{Zhou, H.}, \bibinfo{author}{Bradley, A.},
  \bibinfo{author}{Littwin, E.}, \bibinfo{author}{Razin, N.},
  \bibinfo{author}{Saremi, O.}, \bibinfo{author}{Susskind, J.M.},
  \bibinfo{author}{Bengio, S.}, \bibinfo{author}{Nakkiran, P.},
  \bibinfo{year}{2024}.
\newblock \bibinfo{title}{What algorithms can transformers learn? {A} study in
  length generalization}, in: \bibinfo{booktitle}{The Twelfth International
  Conference on Learning Representations, {ICLR} 2024, Vienna, Austria, May
  7-11, 2024}, \bibinfo{publisher}{OpenReview.net}.
\newblock \URLprefix \url{https://openreview.net/forum?id=AssIuHnmHX}.

\end{thebibliography}





\end{document}